\pdfoutput=1
\documentclass[11pt]{article}

\usepackage{acl}
\usepackage{times}
\usepackage{latexsym}
\usepackage{inconsolata}
\usepackage{booktabs}
\usepackage{graphicx}
\usepackage{tabularx}
\usepackage{xtab}
\usepackage{afterpage}
\usepackage{placeins}
\usepackage{flafter}
\usepackage{longtable}
\usepackage[font=small,labelfont=bf]{caption}
\usepackage[caption=false]{subfig}
\usepackage{enumitem}
\usepackage{soul}

\usepackage[T1]{fontenc}
\usepackage[utf8]{inputenc}

\usepackage{microtype}

\usepackage{todonotes}

\newcommand\oursabstract{DOC}
\newcommand\oursimpl{\textsc{doc}}
\newcommand\rethree{Re$^3$}
\newcommand\rethreeimpl{\textsc{re}$^3$}
\newcommand\fudge{FUDGE}
\newcommand\oursimplonelevel{\textsc{doc-nooutline}}
\newcommand\oursimplnofudge{\textsc{doc-nocontrol}}

\newcommand\oursabstractfull{Detailed Outline Control}
\newcommand\detgen{detailed outliner}
\newcommand\detgencaps{Detailed Outliner}
\newcommand\detcon{detailed controller}
\newcommand\detconcaps{Detailed Controller}
\newcommand\vT{$\mathbf{T}$}
\newcommand\vr{$r$}
\newcommand\vc{$c$}
\newcommand\vp{$p$}

\newcommand\vP{$\mathbf{P}$}
\newcommand\rollingopt{\textsc{rolling-opt}}
\newcommand\rollinggpt{\textsc{rolling-gpt}}

\title{\oursabstract{}: Improving Long Story Coherence With \oursabstractfull{}}

\author{{\bf Kevin Yang}$^{1}$\ \ \ \ 
    {\bf Dan Klein}$^1$\ \ \ \ 
  {\bf Nanyun Peng}$^2$\ \ \ \ 
  {\bf Yuandong Tian}$^3$\\
    $^1$UC Berkeley, $^2$UCLA, $^3$Meta AI \\
     \texttt{\{yangk,klein\}@berkeley.edu,violetpeng@cs.ucla.edu,yuandong@meta.com}
     }

\begin{document}
\maketitle
\begin{abstract}% \violet{can we remove ``longer'' in the title -- it makes the paper sounds more incremental; or change it to ``long''}
We propose the \oursabstractfull{} (\oursabstract{}) framework for improving long-range plot coherence when automatically generating several-thousand-word-long stories.
% Existing methods struggle to consistently maintain the long-range plot structure, even when directed by a high-level plan. 
% \oursabstract{} improves a language model generator's ability to maintain overarching plot coherence in two complementary ways. 
\oursabstract{} consists of two complementary components: a \detgen{} and a \detcon{}. The \detgen{} creates a more detailed, hierarchically structured outline, shifting creative burden from the main drafting procedure to the planning stage. The \detcon{} ensures the more detailed outline is still respected during generation by controlling story passages to align with outline details.
% \oursabstract{} is based on two complementary ideas.
% First, \oursabstract{} uses a more detailed and hierarchically structured outline, shifting creative burden from the main drafting procedure to the planning stage.
% Second, to ensure the more detailed outline is respected during generation, we train a control scheme to generate passages that align with outline details.
% Second, to maintain faithfulness to the outline, we design a controlled generation approach which operates on natural language instructions while keeping the ability to adjust control strength. %Therefore, \oursabstract{} combines a multi-level hierarchical planning system with a controlled generation approach for drafting the main story. 
In human evaluations of automatically generated stories, \oursabstract{} substantially outperforms a strong \rethree{} baseline~\cite{yang2022re3} on plot coherence (22.5\% absolute gain), outline relevance (28.2\%), and interestingness (20.7\%). 
Humans also judged \oursabstract{} to be much more controllable in an interactive generation setting.
% TODO consider adding the character generation / usage system?
% some better motivation - not just more detailed or more concrete for the outline, but the goal should be to make it more storylike. what is the logic?
% how to conceptualize in the outline this idea that you want to shove the creativity into the planner?
\end{abstract}

\section{Introduction}

% Story generation---especially long-form story generation---is an important test bed for long-range coherence in text generation,
% but is highly challenging. Only recently have prior efforts even attempted to generate stories of comparable length to human-authored ``short stories''~\cite{yang2022re3}.
% But compared to humans, state-of-the-art story generation systems still fall short in long-range plot coherence. 

\begin{figure}[t!]
\centering
\includegraphics[width=0.98\linewidth]{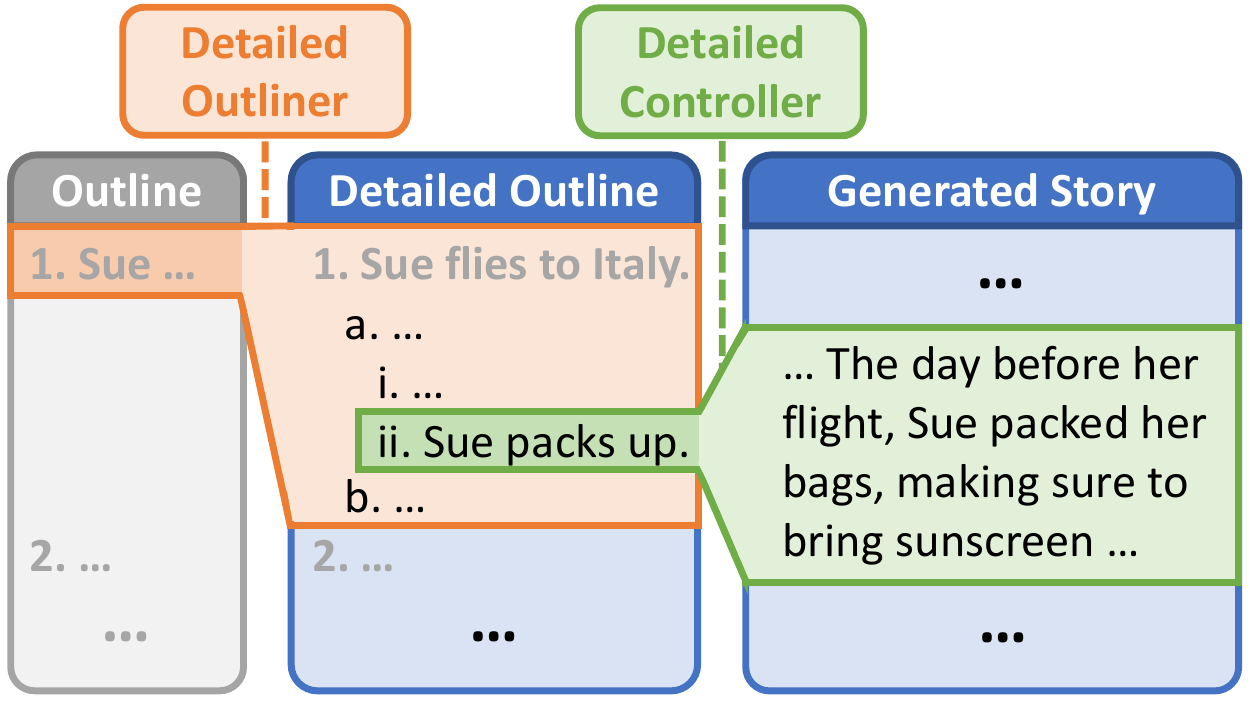}
\caption{High-level overview of \oursabstract{}. Our \detgen{} expands a brief initial outline into a more detailed outline. Our \detcon{} then maintains faithfulness to the more detailed outline when drafting the main story.}
\label{fig:overview}
\vspace{-1em}
\end{figure}

Recent advancements in natural language generation systems have fueled increased interest in \textit{long-form} text generation, in which texts may span thousands of words or more.
% greatly increased attention to long-horizon tasks, in which texts may span hundreds or even thousands of words~\cite{tay2020long,shaham2022scrolls}. 
Compared to tasks with shorter outputs, long-form generation involves meaningfully different challenges. It is nontrivial to maintain overarching coherence, or even basic relevance to an initial premise or plan. 
Even the most advanced language models to date, such as GPT4~\cite{gpt4blog}, still cite long context as a major direction for further improvement, and require structured planning to generate text longer than a few hundred words.% https://openai.com/research/gpt-4

% Maintaining overarching coherence over texts spanning hundreds or even thousands of words has long been a difficult challenge for natural language generation systems

% Previously, \rethree{} showed some approaches for improving long-range coherence and relevance, involving a planning system. However, while the planning system improves these aspects compared to simple rolling window baselines, \rethree{}'s generated stories cannot consistently maintain long-range coherence/relevance at a high level, so there's still a lot of room for improvement. In particular, in several of their example stories, parts of the plan are not well followed, or events are invented mid-generation which seem incongruent with the overarching plot. 

% Many prior works spanning diverse domains have employed hierarchical structure to facilitate longer generations.
% % One important idea to facilitate longer text generation in different domains has been the idea of hierarchical structure. 
% This hierarchical structure takes many forms, such as in the model attention~\cite{guo2021longt5}, 
% or a natural language premise or plan~\cite{fan2018hierarchical,yang2022re3}.
% The latter paradigm especially is both human-interpretable 
% and amenable to being used in conjunction with pretrained language models like GPT3~\cite{brown2020language} and OPT~\cite{zhang2022opt} via prompting. 
% But even after leveraging state-of-the-art language models, generating high-quality texts thousands of words long can be frustratingly difficult.

In this work, we focus on long-form \textit{story} generation, which is representative of the major difficulties in long text generation. Only recently have prior efforts even attempted to generate stories of comparable length to human-authored ``short stories'' (\rethree{}, \citet{yang2022re3}). Compared to humans, state-of-the-art story generation systems like \rethree{} still fall short in numerous areas: common failure modes include insufficient high-level planning resulting in local fluency amid global incoherence, or deviating from said planning even when it exists.

To bridge some of this gap, we propose the \oursabstractfull{} (\oursabstract{}) framework. While reusing the high-level planning-drafting-revision structure of \rethree{}, \oursabstract{} improves long-range plot coherence via two complementary approaches. 

First, our \textit{\detgen{}} refines a brief initial outline into a more detailed, hierarchical one (Figure \ref{fig:overview} left). %, with length scalable according to the desired
% scope of generation . 
% As motivation, we observe that a human author may also iteratively refine and expand a brief initial outline before drafting a long document. 
% Compared to improvising new plot points on the fly, the author might instead plan a coherent overarching plot at a high-level outlining stage, using the expanded outline to provide more detailed guidance during drafting. 
As motivation, a human author might also iteratively refine and expand a brief initial outline before drafting a long document, using the outline to guide a coherent plot rather than improvising plot points on the fly.
Accordingly, 
our \detgen{} employs a structured prompting procedure to create a detailed outline with length scalable according to the desired scope of generation. Individual outline items are associated with a setting and characters, and are carefully filtered for relevance and coherence in context.

Second, our \textit{\detcon{}} maintains faithfulness to our detailed outline by controlling passage generation based on corresponding outline items (Figure \ref{fig:overview} right). Because our detailed outline imposes many overlapping soft constraints, the \detcon{} must exert sufficient control strength to enforce them. The \detcon{} must also accommodate flexible natural language inputs and be computationally efficient when generating with state-of-the-art large language models. We implement the \detcon{} as an OPT-350m-based controller according to \fudge{}~\cite{yang2021fudge}, designing a contrastive training procedure that aligns summaries to passage prefixes. %which enables the controller to handle natural language instructions for the specific
% task of generating passages faithful to a given outline item. 
In particular, we construct fluent
hard negatives to encourage lengthy outputs to be not only initially on topic, but relevant throughout.

Compared to the original \rethree{}, the previous state of the art in long-form story generation, 
using \oursabstract{} achieves dramatically higher plot coherence (22.5\% absolute gain), outline relevance (28.2\%), and even interestingness (20.7\%) in pairwise human evaluations (Section \ref{sec:experiments}).
%  and following the initial premise compared to \rethree{}.
Our ablations indicate that both the detailed outliner and detailed controller are critical  (Section \ref{sec:analysis_ablations}).
We also demonstrate that \oursabstract{} can generate stories in collaboration with humans, 
interacting at a high-level planning stage rather than passage-by-passage as in many prior works~\cite{coenen2021wordcraft,lee2022coauthor}, and is overwhelmingly preferred over the original \rethree{} in this setting (Section \ref{sec:experiments_control}).\footnote{All code and models are available at \url{https://github.com/yangkevin2/doc-story-generation}.}

% in human-interactive experiments, that human find our system much more controllable via natural language outlines, 
% making it easy for non-experts to generate coherent longer stories from just a premise or outline. 

% Discuss the experiments we ran, how long our stories are, and how much better we are compared to \rethree{} and rolling window baselines. Mention human experiments if we run them. Mention interpretability/controllability

    % - [ ] general challenges in (long?) story generation
    % - [ ] some approaches to tackle long range coherence - better language models, re3
    % - [ ] problems still with coherence/relevance, still progress to be made
    %     - [ ] cite problems remaining according to what re3 says in the paper
    % - [ ] how do we propose to address these issues, our method
    %     - [ ] hierarchical planning system: we want more detailed, concrete outlines. 
    %         - [ ] intuitive concrete examples that are easier to follow
    %     - [ ] controlled generation system
    %         - [ ] instead of just prompting, you want a proper controlled generation approach where you can properly increase the control strength
    % - [ ] experimental results
    % - [ ] mention some HCI? try some human writing experiments?
\section{Related Work}

% While many previous works have explored automatic story generation, most have focused on very short stories 
Although we generate stories an order of magnitude longer compared to most prior works~\cite{wang2019t,yao2019plan,qin2019counterfactual,xu2020megatron,wang2022language}, we highlight below several works which employ related ideas.
% Most previous works on automatic story generation have focused on stories
% between a few sentences and a couple of paragraphs in length~\cite{wang2019t,yao2019plan,qin2019counterfactual,xu2020megatron,wang2022language}. 
% While we focus on much longer stories, averaging 3500 words in our experiments, we nevertheless highlight related approaches from prior work.
% In this work,
% we focus on maintaining overarching plot coherence and quality in much longer stories---3500 words on average in our main experiments. Nevertheless, some prior works have explored some similar ideas to those we employ in our work.

\medskip
\noindent\textbf{Hierarchical Generation.}
A key component of \oursabstract{} is our \detgen{}, which generates an outline hierarchically. 
%, inspired by the human outlining procedure. 
Hierarchical structure in long-form generation can be implemented as part of the model architecture itself~\cite{yang2016hierarchical,miculicich2018document,guo2021longt5}, or as natural language outlines or structured schema~\cite{fan2018hierarchical,yao2019plan,goldfarb2020content,rashkin2020plotmachines,zhao-etal-2020-bridging,narayan-etal-2021-planning,tian2022sonnet,mirowski2022co,yang2022re3}.
% \citet{fan2018hierarchical} also use a hierarchical planning scheme for generating stories, based on initially generating a premise followed by turning
% that premise into a complete story. Other works similarly use brief outlines or structured schema~\cite{yao2019plan,goldfarb2020content,rashkin2020plotmachines,tian2022sonnet}. 
% Recently, \citet{yang2022re3} use a premise as the starting point and
% construct a much more detailed plan containing a setting, characters, and brief outline. %; they then use this plan to guide the main story generation,
% and generate stories of over 2000 words in their main experiments. 
\oursabstract{}'s \detgen{} also builds a natural language outline, but can easily increase the level of detail to match the desired scope of the final story. 

\medskip
\noindent\textbf{Controlled Generation.}
A second key component of \oursabstract{} is the \detcon{}, which increases faithfulness to our detailed outline. %Prior work has proposed
% many control schemes for text generation. 
Prior works such as \citet{hu2019improved} use constrained decoding to guarantee rule-based constraints, while \citet{dathathri2019plug,krause2020gedi,yang2021fudge} propose modular control schemes based on an auxiliary model for a desired attribute. %\todo{cite any more recent controlled generation works? see what cites gedi}
However, such methods typically do not handle natural language instructions. 

In contrast, prompting~\cite{brown2020language,zhong2021adapting,sanh2021multitask,wu2022autoformalization,kojima2022large,ouyang2022training} offers a lightweight, flexible alternative. 
% Instruction-tuned models such as
% InstructGPT3~\cite{ouyang2022training} have demonstrated strong ability to follow human instructions.
However, while prompts are an effective way to \textit{provide context}, they may be insufficient for \textit{enforcing constraints} due to the limited control strength, which is not easily tunable unlike in our \detcon{}. %, unlike the control scheme we use in \oursabstract{}---although we employ prompting in tandem with our approach as well. %But prompting can be used in tandem with our approach as well, which we do. 

% In this work, we build a lightweight control scheme based on \fudge{} which has some benefits of both approaches. 
% We train a variation of \fudge{} to take \textit{natural language} descriptions for the specific task of generating story passages following the given description.
% Because the method is built on \fudge{} rather than prompting, we can increase the control strength as necessary for the complex constraints imposed by our story plan
% and detailed outline. Of course, one can use this method in tandem with prompting and reranking, as we do in our experiments.

\medskip
\noindent\textbf{Human-In-The-Loop Story Generation.} Some previous works generate longer stories with a human in the loop~\cite{goldfarb2019plan,coenen2021wordcraft,lee2022coauthor,chung2022talebrush,ippolito2022creative,mirowski2022co}. We emphasize
that \oursabstract{} is designed to generate stories without human intervention. 
Nevertheless, due to planning in natural language space, \oursabstract{} is in principle highly human-controllable. Unlike
methods which interact with the human passage by passage~\cite{coenen2021wordcraft,lee2022coauthor}, \oursabstract{} can also interact 
at a higher-level planning stage, as explored in Section \ref{sec:experiments_control}. %Finally, \citet{mirowski2022co} also use planning to generate theatre scripts, but primarily focus on human-in-the-loop generation rather than automatic generation. Like \citet{yang2022re3}, their planning operates at a fixed level of detail and their outputs are controlled only by prompting, two areas we aim to improve in \oursabstract{}. 
\section{\oursabstractfull{}}\label{sec:method}

% \todo{main figure? or just schematic overview in intro?}

We introduce the \oursabstractfull{} (\oursabstract{}) framework, aiming to improve long-range plot coherence in automatically generated long stories. %\oursabstract{} consists of two major components: 

\begin{figure}[!t]
\centering
\includegraphics[width=0.98\linewidth]{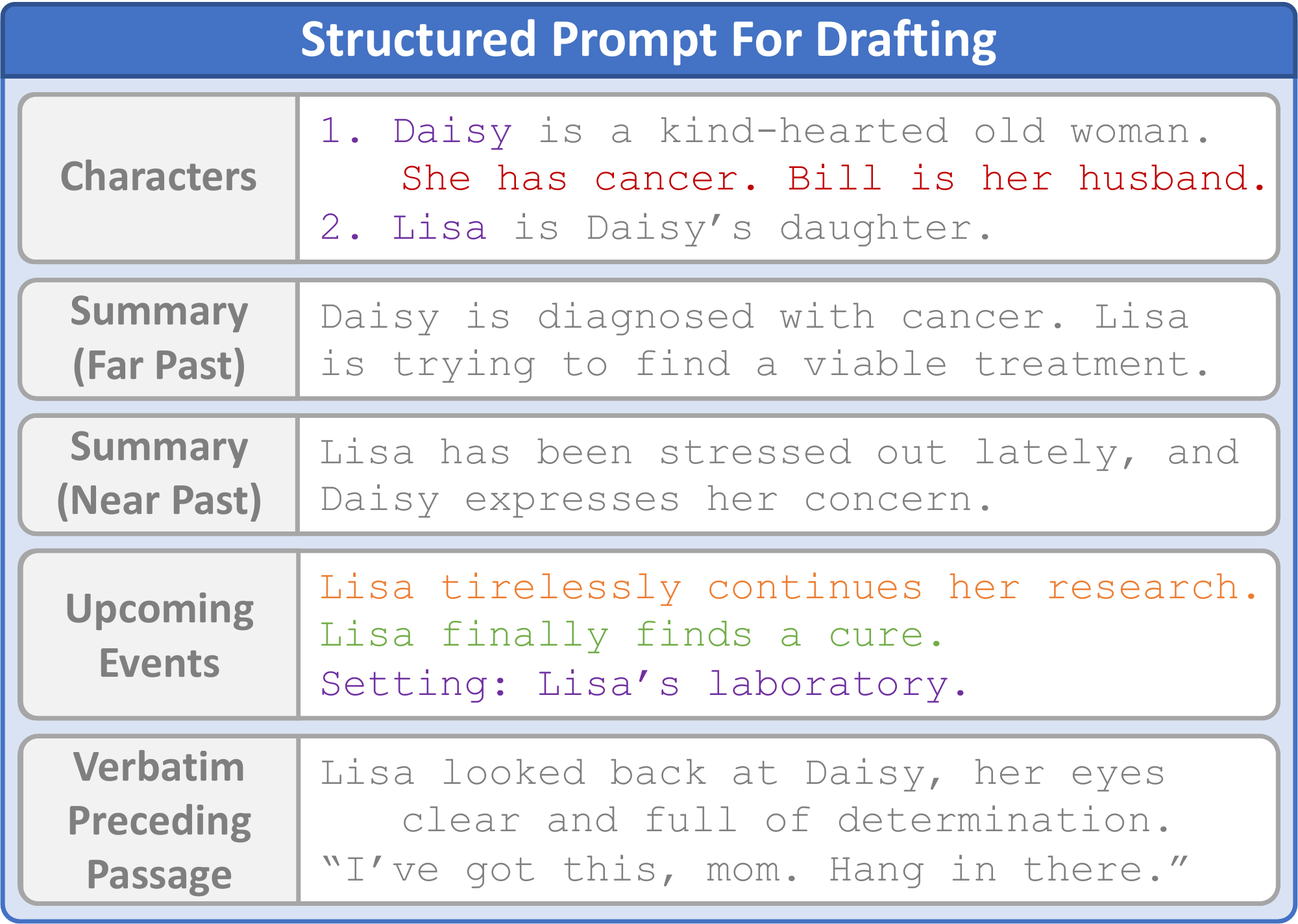}
\caption{Stylized example showing the main components of the structured prompt used to draft new story passages in \rethree{} and \oursabstract{}. Leveraging our detailed outline and \detcon{}, new elements of \oursabstract{}'s prompt include character development over time ({\color[RGB]{192,0,0}red}), more detailed events based on outline leaf nodes ({\color[RGB]{237,125,49}orange}), future context ({\color[RGB]{112,173,71}green}), and improved setting and character information ({\color[RGB]{112,48,160}purple}).}
\label{fig:drafting_prompt}
\vspace{-1em}
\end{figure}

\subsection{Background and Motivation}\label{sec:background}

A major inspiration for our work is \rethree{}~\cite{yang2022re3}, which generates plot-coherent long-form stories of over 2000 words by decomposing the writing process into planning, drafting, rewriting, and editing steps. Their high-level plan contains a setting, character inventory, and brief three-point outline (e.g., Figure \ref{fig:overview} ``Outline''). In particular, when drafting each next story passage, they inject relevant context from the high-level plan and previously generated story via structured prompting (Figure \ref{fig:drafting_prompt}). They finally rerank possible continuations using rerankers for outline relevance and passage coherence, and edit for consistency.
\oursabstract{} follows the high-level writing process and structured-prompting-based passage generation proposed by \citet{yang2022re3}, though we remove the time-consuming editing step, which they find does not significantly affect final story quality.

However, \citet{yang2022re3} note that despite greatly outperforming simple rolling-window baselines, \rethree{} still makes frequent errors in long-range coherence: some stories still contain lengthy passages which seem not to fit the surrounding context, or deviate heavily from the initial outline. \oursabstract{} aims to address these shortcomings via two major innovations: more detailed planning via our \detgen{}, and correspondingly fine-grained control during drafting via our \detcon{}. 

% However, while greatly outperforming simple rolling-window baselines, \rethree{} still often fails to maintain strong long-range plot coherence. As noted in \citet{yang2022re3}, some of \rethree{}'s stories contain lengthy passages which seem not to fit the surrounding context, and their stories frequently deviate from their initial plan. \oursabstract{} aims to address these shortcomings via two major innovations: we implement much more detailed planning via our \detgen{}, and correspondingly more detailed control during drafting via our \detcon{}. 

% we identify critical areas of potential improvement in the planning and drafting stages, motivating our \oursabstractfull{} framework.\violet{it seems to me that the following discussion makes strong assumption about readers' knowledge about DOC's strategy, while they don't know anything at this point. Maybe start with a high-level introduction to say the major innovation of DOC over RE3 is detailed planning and detailed control, with some motivation why. then expand from there.} %Our empirical results in Section \ref{sec:experiments} substantiate the efficacy of our proposed methods.

\medskip
\noindent\textbf{\detgencaps{} Motivation.} %\rethree{}'s high-level plan
% consists of a setting, list of characters, and brief three-point outline (e.g., Figure \ref{fig:overview} left). %An example outline is shown in Table \ref{tab:re3_outline_example}.\footnote{Technically this outline was generated by our code rather than \rethree{}'s, but the differences at the top-level outline are minor.
% We merely use a longer character list, and filter outline items individually
% rather than generating the full outline in one shot with rejection sampling.} 
While \rethree{}'s outlines are plausible, they are insufficiently concrete, and do not scale to longer stories. A human author would not write a novel given just a three-sentence beginning, middle, and end. Not only can a more detailed outline empirically result in improved plot coherence (Section \ref{sec:experiments}), but it can enable greater control in human interaction as well (Section \ref{sec:experiments_control}). 
% Even the very first item of the outline in Table \ref{tab:re3_outline_example} leaves many questions unanswered: what exactly happens when Jenna meets Brian?
% Do they start dating immediately--and shouldn't dating be a prerequisite for Jenna to think about marriage later?
% Given just ``Jenna Adams meets Brian Johnson and immediately feels drawn to him,'' existing models may struggle to generate a lengthy passage that maintains high-level plot coherence both internally and with the rest of the outline.

% In the absence of additional information, the story generator must weave a plot-coherent passage given just the short
% description ``Jenna Adams meets Brian Johnson and immediately feels drawn to him.'' When this passage spans a thousand words or more,
% existing language models may struggle to invent a passage following this topic which maintains high-level plot coherence both internally,
% and with the rest of the story as described in the remainder of the outline. %Indeed, \rethree{}'s generated passage for the first item of the outline
%in Table \ref{tab:re3_outline_example} leads quite poorly into the next (\todo{refer to abridged example})

Therefore, \oursabstract{} %we aim to shift creative work from drafting to planning, 
% where maintaining overarching plot coherence is easier due to lesser total text length. 
constructs a detailed outline (e.g., Figure \ref{fig:overview} ``Detailed Outline'') with depth adjustable according to the desired length of the final story. The detailed outline shifts creative burden from drafting to planning, reducing the need to improvise plot points on the fly during drafting. %Table \ref{tab:ours_outline_example} shows part of a much more detailed outline produced by \oursabstract{}.% corresponding to part of the high-level outline from Table \ref{tab:re3_outline_example}.

\medskip
\noindent\textbf{\detconcaps{} Motivation.}
% Generating each next story passage while maintaining long-range coherence and faithfulness to a detailed outline--
% containing not only lower-level event descriptions but also specified scenes and characters--
% Maintaining faithfulness to an outline during drafting is a difficult controlled generation problem.
The greater level of detail in our outline makes it much harder to stay faithful to that outline.
% is a difficult text generation problem with many simultaneous soft constraints. 
% Additionally, working with large language models such as GPT3-175B 
% greatly limits our options: the chosen control scheme must be modular as well as computationally efficient and scalable.
To work with large language models such as GPT3-175B during drafting, prior works such as \rethree{} have relied on clever prompting together with rejection sampling or reranking. 
However, prompting and reranking approaches are limited in the strength of control they can exert over the model distribution, which is especially problematic for systems like \rethree{} which rely on complex constraints and long context in a structured prompt. %(Figure \ref{fig:drafting_prompt}). 
Indeed, \citet{yang2022re3} observe that many of \rethree{}'s stories already
omit parts of even their brief three-point outline---and \oursabstract{}'s outline will impose far more detailed constraints. %, empirically requiring stronger control (Section \ref{sec:analysis_detailed_relevance}). 

Therefore, we design \oursabstract{}'s \detcon{} to more strongly enforce the complex natural language constraints set by the outline.
% to influence passage generation more strongly according to our detailed outline, while simultaneously handling the complex natural language constraints set by the outline. 
Our \detcon{}, an adaptation of \fudge{}~\cite{yang2021fudge}, will operate token-by-token throughout generation instead of relying on only an initial prompt or post-hoc rejection sampling.
% To simultaneously satisfy all the complex constraints set by our detailed outline requires a controlled generation method which supports a stronger level of control, 

\begin{figure}[!htbp]
\centering
\includegraphics[width=0.98\linewidth]{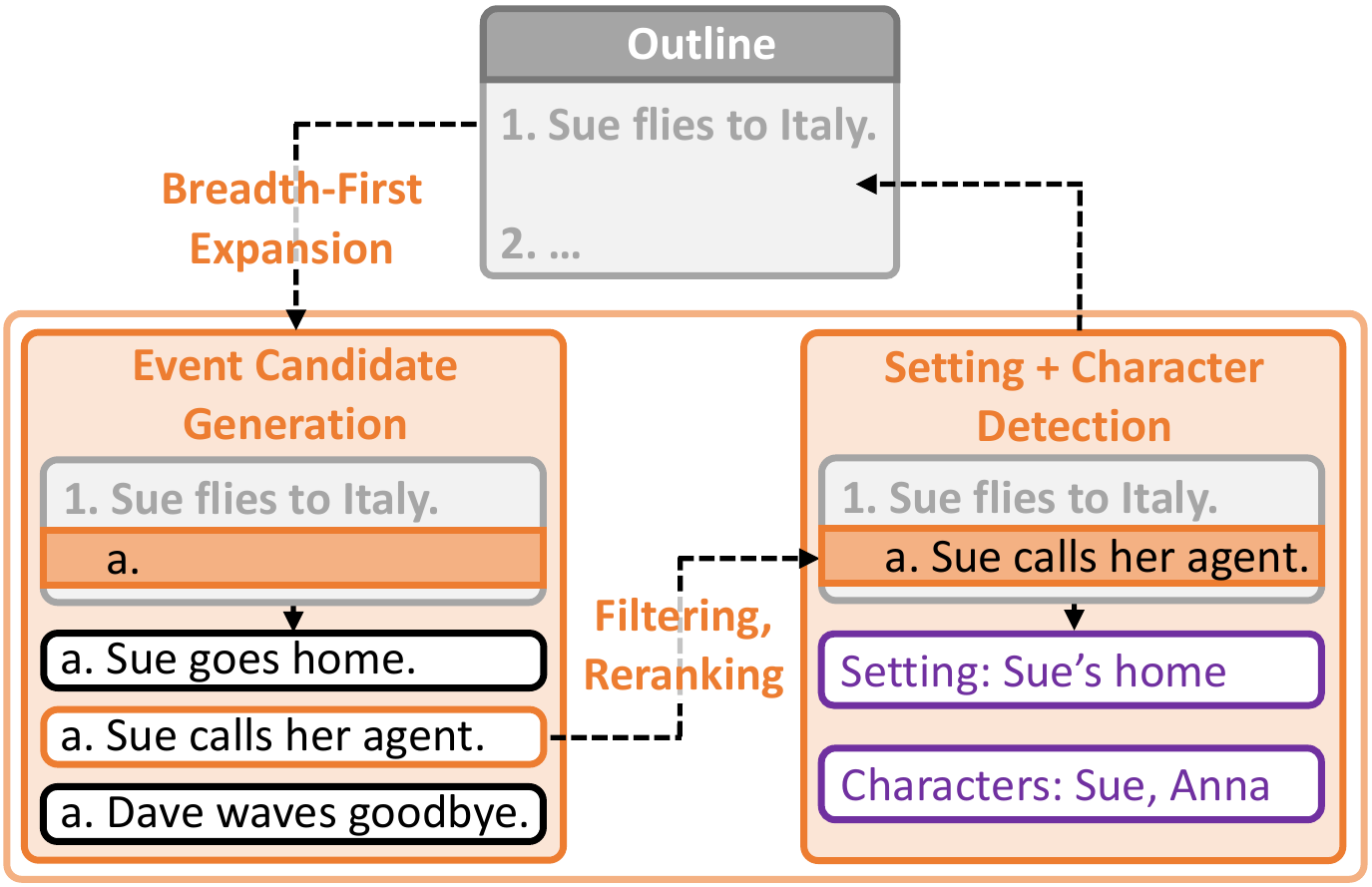}
\caption{Diagram of new entry creation in the detailed outline. Our \detgen{} recursively expands outline items in breadth-first order. To create each new entry, it proposes candidate events and selects the best via filtering and reranking, and then detects the setting and relevant characters.}
\label{fig:detgen}
\vspace{-1em}
\end{figure}

\subsection{\detgencaps{}}\label{sec:method_detailed_outlines}

% [Show examples of \rethree{} plans, e.g. something like ``they succeed after a lot of twists and turns'']
% We now describe how the \detgen{} creates a detailed outline.

% We begin with the first major goal of \oursabstractfull{}, which is to generate outlines providing much more detailed guidance to the generator at drafting time.

% \todo{make figure outline prompt example}

% \subsubsection{Hierarchical Outline Generation} 

Our \detgen{} recursively generates a hierarchical detailed outline at arbitrary granularity. Figure \ref{fig:detgen} summarizes the individual components. %A truncated example is shown in Table \ref{tab:ours_outline_example}.

% We propose a recursive, hierarchical procedure for generating a detailed outline at arbitrary granularity. A truncated example is shown in Table \ref{tab:ours_outline_example}.
% Our generation procedure will be highly scalable, enabling arbitrary levels of granularity.
% Additionally, our natural language outlines are designed to be easily editable by a human, should a human wish to provide their own input at the planning stage.  
% Table \ref{tab:ours_outline_example} shows the part of our detailed
% outline corresponding to the first item in Table \ref{tab:re3_outline_example}.

% We now describe our procedure in detail. 

\medskip
\noindent\textbf{Breadth-First Expansion.} Viewing the outline as a tree \vT{} initialized as just a root node \vr{}, we generate children in breadth-first expansion order. 
Starting from the items of the initial top-level outline (depth 1), we generate all of 
their children (depth 2), then all childrens' children (depth 3), and so forth. For each parent node \vp{}, we generate children one by one, stopping when a child \vc{}'s event description ends with the end-of-text token. We restart and resample for a given \vp{} if there are too few or too many children, although empirically this procedure almost always results in just two or three children.
We terminate outline generation after reaching a pre-specified depth. 

\subsubsection{Event Candidate Generation} %\violet{should probably motivate on why do we want to generate ``events'', there should be plenty of literature on narratives connect narrative with events. Also, it's unclear what's `Figure \ref{fig:detgen} bottom left', can you refer to the actual event that you discuss here? e.g., `sue call her agent'(?)} \kevin{is it necessary to motivate the events? re3 already uses events in the outline. i am pretty much referring to the bottom left box though}
% We now describe concretely how each outline item is generated.
To generate possible event descriptions for a new child \vc{} (Figure \ref{fig:detgen} bottom left), we use a structured prompting approach.
To maintain coherence with pre-existing nodes, the prompt contains context
from all of \vc{}'s ancestors, together with their respective children; in this way we provide relevant context whose length scales linearly with depth.
% The provided context is illustrated for a \vc{} at depth 3
% in Figure \ref{tab:outline_prompt_example}. \todo{add this back?}
Suffix context is injected via the GPT3 Insertion API using InstructGPT3-175B (\texttt{text-davinci-002}), the most advanced GPT model at the time of our experiments. See Appendix \ref{sec:appendix_example_prompts_outlineitem} for an example prompt.
% , 
% although the top-level outline at depth 1 is generated almost identically to \rethree{} due to the lack of suffix context at this stage.%\todo{note shifting depth for suffix context}

\medskip
\noindent\textbf{Filtering and Reranking.} %\violet{I feel this is part of event candidate generation. Should we promote the latter into a subsection, so the structure is clearer} 
After generating several event candidates for each \vc{}, we select the best via filtering and reranking. Specifically, we remove ill-formed candidates or those which are highly repetitive
compared to nodes not in \vc{}'s ancestors,\footnote{However, since \vc{} is a sub-event of its ancestors,
it is acceptable to repeat parts of ancestor texts. If no candidates remain after filtering,
we accept \vp{} as a leaf node which is already sufficiently concrete 
and does not require further expansion.} as determined by both word overlap and an entailment model~\cite{laurer2022less}.

For the first child of each parent, we select the remaining candidate most relevant to the parent by sentence similarity~\cite{reimers2019sentence}. %,
% and select the best candidate as the final outline item.% and
For other children, to avoid repetition and improve plot coherence, we select via an ordering model that predicts if an event occurs in the correct location relative to 
nearby context. 
% To train the ordering model, we collected 1000 brief stories (2-3 paragraphs) written by InstructGPT3-175B, which we observed typically use a high-level outline-like style---essentially, ``telling'' rather than ``showing.'' We then finetuned \texttt{roberta-large}~\cite{liu2019roberta} to predict whether a given sentence in such a story appears in the correct order by training contrastively, with negatives constructed by randomly moving the given sentence to elsewhere in the story. %\todo{cite DINO if relevant?}%
The ordering model is trained by finetuning \texttt{roberta-large}~\cite{liu2019roberta} to detect out-of-order events in short outline-like stories. See Appendix \ref{sec:appendix_order_model} for complete details on our filtering and reranking pipeline. % on \todo{describe what you finetuned this on}
% (Appendix \ref{sec:appendix_order_model}). 
% Appendix \ref{sec:appendix_trivial3level} shows examples of problematic outlines generated without filtering or reranking, demonstrating the necessity
% of doing so. 

% Thus the top-level outline is generated nearly identically to that of \rethree{}, 
% except that we generate one item at a time instead of generating a complete outline, to facilitate more detailed filtering and reranking. 

\subsubsection{Setting and Character Detection} 
% In keeping with the spirit of our more detailed outline, 
We further augment our outline by explicitly representing settings and characters for each outline item (Figure \ref{fig:detgen} bottom right), % Table \ref{tab:example_scene_chars})
thus shifting additional creative work from drafting to planning. 

Our setting and character list are obtained by prompting InstructGPT3-175B (Appendix \ref{sec:appendix_example_prompts_scenechar}). 
Characters are matched against an initial character inventory similar to that of \rethree{}, though we generate more characters since our outline is more detailed. 

% Since our outline is more detailed, we use a larger initial character inventory compared to \rethree{} following their character generation procedure.
% Given each outline item, we prompt InstructGPT3-175B for a scene and character list,
% and match the detected characters against an initial character inventory (constructed similarly to \rethree{}, though we generate characters since our outline is more detailed). %See  for example prompts. 

% In contrast, \rethree{} provides only a single scene and list of major characters for the entire story, which we observed could lead
% to consistency errors, such as when the language model ``forgets'' the current story setting after a long passage.
% The additional scene and character information in our outline can help avoid such consistency issues. 

% \begin{table}[!htbp]
% \small
% \begin{tabularx}{\linewidth}{X}
% \toprule
% \texttt{i. Daisy's cancer treatment is difficult, but with the support of her friends and family, she ultimately beats the disease. Scene: the hospital. Characters: Daisy Mayberry, Tanya Swanson, Evelyn Chambers}\\
% % \\
% % \texttt{ii. In the aftermath of her treatment, Daisy looks back on her experience and how it has changed her. Scene: her home. Characters: Daisy Mayberry}\\
% \bottomrule
% \caption{An outline item generated by \oursabstract{}, with scenes and characters added.}
% \vspace{-2em}
% \label{tab:example_scene_chars}
% \end{tabularx}
% \end{table}

\subsubsection{Drafting With Detailed Outlines}\label{sec:drafting_detailed_outlines}
%\violet{I view this as part of detail controller -- or do you have a drafting stage before drafting using detail controller?}\kevin{I think this belongs here, because it's a change to drafting that can occur only due to the existence of the detailed outline, and can occur independently of the existence of the detail controller}

After constructing our detailed outline, story drafting largely follows \rethree{}'s structured prompting procedure based on injecting context from the plan and previous story (Figure \ref{fig:drafting_prompt}; Appendix \ref{sec:appendix_example_prompts_drafting}).
% To draft the story after constructing our detailed outline, we largely follow \rethree{}'s procedure based on structured prompting with context
% selected from the plan and previous story (Figure \ref{fig:drafting_prompt}; example in Appendix \ref{sec:appendix_example_prompts_drafting}). 
However, instead of generating a fixed-length passage for each top-level outline item as in \rethree{},
we generate a \textit{variable-length} passage for each \textit{leaf} of our tree-structured outline \vT{} (Figure \ref{fig:drafting_prompt}, orange text), since different leaves may contain events at differing
levels of concreteness. %, it is ideal for story generation advance to the next item in the outline as soon as it has finished describing the previous item (and no sooner), rather
% than writing the exact same number of tokens for each. 
%  as we use the same outline relevance and text coherence rerankers as \rethree{}
% during main story generation, our early stopping criteria are simply based on these reranker scores, with the exact threshold being a hyperparameter. 
Specifically, we reuse the outline relevance and text coherence rerankers from \rethree{}'s rewriting stage to detect when drafting is done for the current outline item, implementing early stopping based on a score threshold. We also generate fewer tokens than \rethree{} before reconstructing the structured prompt, 
% (64 in \oursabstract{} compared to 256 in \rethree{}), 
for finer-grained control. 

In the prompt, we additionally highlight the current setting (Figure \ref{fig:drafting_prompt}, bottom purple text), especially changes in setting. Characters (Figure \ref{fig:drafting_prompt}, top purple text) are also retrieved from the outline. In contrast, \rethree{} selects relevant characters for each passage on the fly during drafting, and does not track setting information, which can result in unexpected changes in story setting.

% \todo{mention that the outline relevance rerankers also operate for the scene + characters?}

% \begin{table}[!htbp]
% \small
% \begin{tabularx}{\linewidth}{X}
% \toprule
% \textbf{Initial Character Description:} 
% \texttt{Daisy Mayberry is a kind-hearted woman in her early 50s who is loved by everyone in her small town.}\\
% \\
% \textbf{Later Character Description:} 
% \texttt{Daisy Mayberry is a kind-hearted woman in her early 50s who is loved by everyone in her small town. Daisy Mayberry has cancer. Elizabeth and Bill Simpson are Daisy's daughter and husband, respectively. Daisy Mayberry owns a hardware store. She has a daughter named Lisa.}\\
% \bottomrule
% \caption{An initial character description, together with a later description of the same character based on previous outline items.}
% \vspace{-2em}
% \label{tab:example_chardesc}
% \end{tabularx}
% \end{table}

\medskip
\noindent\textbf{Character Development Over Time.} %\violet{it's a little unclear when does these character description used}
% One further limitation of \rethree{}, as noted in \citet{yang2022re3}, is its inability to handle
% change over time in story characters. In fact, \rethree{}'s editing system is designed to label changes over time as contradictions, and actively remove such changes.
% As \citet{yang2022re3}
% showed that their editing system did not significantly improve metrics on their final stories, we remove the editing system in this work. 
% Instead, 
Taking advantage of our detailed outline, we explore a simple method to make \oursabstract{} aware of character development over time, which \rethree{} struggled to handle.
Concretely, we attempt to infer a new fact about each character whenever they appear in the outline (Appendix \ref{sec:appendix_example_prompts_chardesc}),
filtering out facts already entailed by a previously inferred fact from an earlier outline item. When drafting a story passage corresponding to a given outline item, 
retrieved character descriptions in the prompt context contain all facts inferred up to that outline item (Figure \ref{fig:drafting_prompt}, red text). %Table \ref{tab:example_chardesc} shows an example description.
% See Appendix \ref{sec:appendix_example_prompts} for example prompts. 

\subsection{\detconcaps{}}\label{sec:method_control}

Next, our \detcon{} enhances the generator's ability to maintain relevance to our detailed outline.
% The second major goal of \oursabstractfull{} is to , which we will do via controlled generation.
% \subsubsection{Logit Warping With Natural Language}
We implement the \detcon{} as a \fudge{}~\cite{yang2021fudge} controller to guide passage generation according to a given summary. However, we will modify the \fudge{} training procedure to improve performance on longer outputs.

\medskip
\noindent\textbf{Lightweight, Adjustable-Strength, Natural Language Control.} 
\fudge{} is a lightweight, modular control scheme that adds logits at each token of generation based on a future-aware discriminator for a desired attribute.
% that treats the base generator as a black box, merely warping the output logits at each
% token by adding log-probabilities for a desired attribute based on a future-aware discriminator. 
Control strength can be increased by multiplying the added logits, but it is nontrivial to handle natural language instructions.
% However, \fudge{} is not designed for natural language instructions. 

We adapt \fudge{} to handle natural language instructions for the specific task of guiding passage generation according to a short description. 
We collect a dataset of passage-summary pairs by prompting InstructGPT3-13B to summarize story passages from the WritingPrompts dataset~\cite{fan2018hierarchical}; these summaries can then be viewed as outline events corresponding to the original passages. We train the \fudge{} discriminator contrastively by finetuning OPT-350m to predict whether a \textit{passage prefix} matches a given summary. In particular, we construct hard negatives by matching passages with summaries from elsewhere in the same story.

The result is a computationally lightweight \detcon{} which can guide passage generation according to a short natural language description, with adjustable control strength.
% takes natural language summaries as input, and can control passage
% generation to follow that summary. 
% (though instead of using the base GPT3-175B \texttt{davinci} as in \rethree{}, we use OPT-175B~\cite{zhang2022opt} due to functionality limitations in the public GPT3 API

% \begin{figure}[t!]
% \centering
% \includegraphics[width=0.98\linewidth]{figures/fudge_harder_negatives.pdf}
% \caption{Construction of harder negative examples for training \fudge{} to maintain relevance.}
% \label{fig:fudge_harder_negatives}
% \vspace{-1em}
% \end{figure}

\medskip
\noindent\textbf{Training to \textit{Maintain} Relevance.} 
% However, this simple contrastive learning approach by itself falls short, especially when generating longer passages. 
In our training data, 
passages are either entirely correct or entirely wrong for a given summary---even for ``hard'' negatives from the same story---so the discriminator learns to predict high probabilities
for any roughly aligned passage at test time. The resulting controller allows longer passages to quickly stray off topic after starting out on topic. 
% as many passages are wont to do in a sensible story generation system. 

Thus we construct even harder training negatives. % (Figure \ref{fig:fudge_harder_negatives}). 
Given a positive passage-summary pair, we split the passage at a sentence boundary, and replace the text after the sentence boundary with text
from another passage in the same story (beginning at a sentence boundary). We thus obtain grammatically-fluent corrupted passages which
begin correctly for a given summary, but eventually stray off topic. %we then train \fudge{} on the off-topic suffix with the negative label. 
Prefixes of such passages ending after the sentence boundary can then be given the negative label during training. 
Thus our \detcon{} learns to \textit{maintain} high relevance to the input description.

Using the same methodology, we also construct ``harder positives'' by mixing negative prefixes with positive completions, improving the controller's ability to get back on track should it go astray.

\subsubsection{Drafting With Detailed Control} 
%\violet{we probably cannot handle this before the anonymity deadline, but I think this part should be made clearer by adding an overview illustration/visualization with all types of controls. For now, it feels like all control is just for relevance.}\kevin{Added italics headings on the enumeration, but all control is indeed just for relevance.}

During drafting, we illustrate the flexibility of our \detcon{} by controlling passages according to three different types of constraints imposed by our detailed outline, as follows.

% We leverage our controller to generate passages satisfying
% the many simultaneous constraints
% of our detailed outline---event description, scene, and characters---illustrating 
% the flexibility of our control scheme in the process. 

\begin{enumerate}[topsep=0pt,itemsep=-1ex,partopsep=1ex,parsep=1ex]
    \item \textit{Event.} We feed the event description (Figure \ref{fig:drafting_prompt}, orange text) verbatim to the controller.
    \item \textit{Setting.} If the setting changed from the previous outline item, we construct an input ``summary'' stating that the characters move to the new setting, using lower control strength
    compared to the event description.
    \item \textit{Character.} If a character appears who did not appear in the previous outline item, we construct an input ``summary'' stating as such, again using lower control strength.
\end{enumerate}

% \medskip
\noindent\textbf{Control Strength.} In practice, we must balance control strength: too low strength risks deviating from the constraint, while too high strength risks narrowly-focused, repetitive generations which sacrifice creativity. We aim to strike this balance dynamically during drafting by using a control strength of 0 initially for each outline item, incrementing it with each subsequent drafting step, until satisfying our early stopping criteria for moving to the next outline item and resetting back to 0.

% \medskip
% \noindent\textbf{Choice of Base Generator.} 
% During the main drafting process we use OPT-175B\cite{zhang2022opt} as served by the Alpa project\cite{zheng2022alpa}, due to the limited functionality supported in the public GPT3 API which makes our control scheme computationally impractical

% In principle, our approach is lightweight relative to the pretrained generator: 
% we only need to run the controller at each token of generation to warp logits.
% However, while our approach is technically compatible with the public GPT3 API used in \rethree{}'s main story generation procedure, 
% it is computationally impractical due to the
% limited functionality supported in the API: to continue generation after logit warping, we need to re-query the API and re-process
% the entire preceding prompt for each generated token. 
% Therefore, during main story generation we use OPT-175B\cite{zhang2022opt} as served by the Alpa project\cite{zheng2022alpa},
% which supports restarting generation from cached key values for the
% previously processed prompt.  
% In any case, when running the \rethree{} baseline with OPT-175B instead of GPT3-175B,
% \footnote{\rethree{} uses the base GPT3 rather than InstructGPT3 for the main story generation, i.e., ``davinci'' rather than ``text-davinci-002''}
% we did not observe much difference in quality in the final generated stories based on inspection.

\medskip
\noindent\textbf{Future Context in Generation.}
Context from future parts of the outline can help generated passages transition better to subsequent story events. However, including future plot points in the prompt risks premature generation of future events in the absence of proper control, which we observed when trying to include such context in \rethree{}. Our \detcon{} remedies this issue to some degree by controlling more strongly toward the current outline item. Therefore, when drafting for a given outline item, we include the next outline item as future context in the prompt (Figure \ref{fig:drafting_prompt}, green text). %, as an initial step in this direction.
\section{Evaluation}\label{sec:experiments}

\noindent\textbf{Experiment Setup.} Our setup is similar to \citet{yang2022re3}. The input is just a brief (English) premise, typically 30-60 words, sampled from InstructGPT3-175B. The output is a complete story. We do not impose further rule-based constraints, as it is unclear how to define a ``story,'' let alone a ``good'' story. Instead, quality will be judged via human-annotated metrics.

\medskip
\noindent\textbf{Metrics.} 
% Because \rethreeimpl{} uses the same top-level outline as \oursimpl{}, 
To decrease noise, we compare 1000- to 1500-word passages corresponding to the same top-level outline item, rather than complete stories. %, thus decreasing noise in evaluation.

We use three main metrics, similar to those from \citet{yang2022re3} (Appendix \ref{sec:appendix_metrics_discussion}), adapted for comparing passages instead of complete stories:

%\footnote{\citet{yang2022re3} use two additional metrics, which we omit. Their ``miscellaneous writing problems'' metric is cumbersome to annotate, requiring more careful reading to ensure accuracy, and measures an axis orthogonal to our main contributions. Their ``humanlike'' metric varies heavily by annotator population: in preliminary experiments, we found that workers on Amazon Mechanical Turk predicted 70-80\% of stories to be human-written, compared to just 30\% on Surge AI.}

\begin{enumerate}[topsep=0pt,itemsep=-1ex,partopsep=1ex,parsep=1ex]
    \item \textit{Coherent.} Percentage of passages judged plot-coherent by human annotators.
    \item \textit{Relevant.} Percentage judged faithful to the corresponding outline item.
    \item \textit{Interesting.} Percentage judged interesting.
\end{enumerate}

Annotators are shown two passages side-by-side (Appendix \ref{sec:appendix_annotation_templates_main}); for each metric we ask them to annotate which passage is better, or possibly both or neither. Thus all numbers are meaningful only relative to the method being compared against.
%(i.e., all numbers are in relation to the method being compared against).
Each pairwise comparison is labeled by three annotators. %See  for an example annotation template.

We use Surge AI 
% \footnote{\url{https://app.surgehq.ai}} 
for annotation due to observing higher-quality results compared to Amazon Mechanical Turk. We find higher agreement compared to \citet{yang2022re3} (Appendix \ref{sec:appendix_annotator_agreement}), likely due to Surge AI and our more focused annotation task.

\medskip
\noindent\textbf{Method Instantiation.} We henceforth refer to the concrete instantiation of our \oursabstract{} framework as \oursimpl{}. In particular, we set outline depth to 3 and limit the branching factor to be between 2 and 5, resulting in stories averaging roughly 3500 words in length. We limit the model context window to 1024 tokens as in \citet{yang2022re3}, so final stories are substantially longer than the visible context at any step. The base generator used during drafting is OPT-175B~\cite{zhang2022opt}, due to the practical issue of requiring deeper model access than the GPT3 API supports (specifically, \textit{efficient} token-level access to logits). See Appendix \ref{sec:appendix_gpt3_discussion} for further discussion, and Appendix \ref{sec:appendix_hyperparameters} for complete hyperparameters.
%\footnote{While our approach is technically compatible with the public GPT3 API,
% it is computationally impractical due to the
% limited functionality supported in the API: for each token, to continue generation after modifying output logits, we need to re-query the API and re-process
% the entire preceding prompt. 
% Therefore, during drafting we use OPT-175B as served by the Alpa project~\cite{zheng2022alpa},
% which supports restarting generation from cached key values for the
% previously processed prompt.  
% In any case, when running the \rethree{} baseline with OPT-175B instead of GPT3-175B,
% % \footnote{\rethree{} uses the base GPT3 rather than InstructGPT3 for the main story generation, i.e., ``davinci'' rather than ``text-davinci-002''}
% we did not observe an obvious difference in quality upon inspection.} 
% Complete hyperparameters in Appendix \ref{sec:appendix_hyperparameters}.

\medskip
\noindent\textbf{Baselines.} We run two baselines. 

\begin{enumerate}[topsep=0pt,itemsep=-1ex,partopsep=1ex,parsep=1ex]
\item \rethreeimpl{}: Our main baseline is based on \rethree{}~\cite{yang2022re3}, the only previous system we are aware of that automatically generates stories of comparable length. For fair comparison, we modify \rethree{} to also use OPT-175B during drafting.
%, and implement the same repetition penalties used during generation in \oursimpl{}. 
Hyperparameters are set to their paper values, except for the number of generation steps per outline item, which we increase slightly to match average story length with \oursimpl{}. We reuse the setting, characters, and top-level outline from \oursimpl{} for \rethreeimpl{}, as the planning differs only slightly up to here (\oursimpl{} only uses more characters, and generates the outline item-by-item instead of in one shot). %The concrete instantiation of \rethree{} is henceforth named \rethreeimpl{}. 
\item \rollingopt{}: A sanity check using OPT-175B with the same context window as \oursimpl{} and \rethreeimpl{}. The prompt contains the premise and top-level outline item (Appendix \ref{sec:appendix_dumb_baseline_prompts}), followed by a rolling window on the previously-generated story as fits in the prompt. \rollingopt{} generates the same length of text per outline item as \rethreeimpl{}. 
% \item \rollinggpt{}: The same as \rollingopt{} but using GPT3-175B (\texttt{davinci}).

%\footnote{The fairest comparisons are those based on the same OPT-175B as \oursimpl{}, but we include one GPT3-175B model for reference. \citet{yang2022re3} noted that instruction-tuned models such as \texttt{text-davinci-002} worked poorly for drafting, so we chose \texttt{davinci}. The newly released \texttt{text-davinci-003} may produce higher-quality outputs, but in preliminary experiments we struggled to generate stories of more than 600-700 words. % (Appendix \ref{sec:appendix_textdavinci003}). 
% In any case, advancements in language modeling are orthogonal to our contributions, and we are excited to explore applications of more advanced language models in future long-form story generation systems.}
\end{enumerate}

\begin{table}[htbp]
\small
\centering
\begin{tabular}{@{}lccc@{}}
\toprule
\textbf{Method}          & \textbf{Coherent} & \textbf{Relevant}& \textbf{Interesting }\\
\midrule
\rethreeimpl{}                  & 45.1     & 37.1     & 39.4       \\
\oursimpl{}                  & \textbf{67.6}       & \textbf{65.3}    & \textbf{60.1}        \\
\midrule
\rollingopt{}                  & 38.0 & 25.4&  25.4   \\
\oursimpl{}                   & \textbf{80.8}& \textbf{78.9}&  \textbf{69.5}       \\
% \midrule
% \rollinggpt{}                  & 44.1& 25.8&  42.7      \\
% \oursimpl{}               & \textbf{81.7}& \textbf{83.1}& \textbf{70.0}        \\
\bottomrule
\end{tabular}
\caption{\small Pairwise comparisons of \oursimpl{} against baselines on passages corresponding to top-level outline items from 20 stories. Bold indicates significance with $p < 0.05$. \oursimpl{} stories are rated substantially more plot-coherent, outline-relevant, and interesting compared to \rethreeimpl{} and \rollingopt{}.}
\label{tab:main_results}
\vspace{-0.5em}
\end{table}

% define new column type
\newcolumntype{Y}[1]{%
  >{\small\everypar{\hangindent=1em}\arraybackslash}p{#1}%
}

\begin{table}[!t]
\small
\begin{tabular}{@{}Y{0.95\linewidth}@{}}
\toprule
\textbf{\texttt{PREMISE:}} \texttt{A young woman is determined to never get married and live her life alone, but when she meets a man who seems perfect for her, she begins to rethink her decision.}\\
\midrule
\textbf{\texttt{GENERATED OUTLINE:}}\\
\texttt{1. Jenna Adams meets Brian Johnson and immediately feels drawn to him.}\\
\texttt{\quad \quad a. Jenna Adams meets Brian Johnson and feels an instant connection to him.}\\
\texttt{\quad \quad b. The two of them start dating and Jenna Adams begins to fall in love with Brian Johnson.}\\
\texttt{2. Jenna Adams starts to think that maybe marriage isn't so bad after all when Brian Johnson seems like the perfect man for her.}\\
\texttt{\quad \quad a. Jenna Adams starts to think that maybe marriage isn't so bad when Brian Johnson seems like the perfect man for her.}\\
\texttt{\quad \quad b. After much soul searching, Jenna Adams decides that she wants to marry Brian Johnson.}\\
\texttt{3. However, when Brian Johnson's ex-girlfriend shows up and tries to win him back, Jenna Adams realizes that marriage isn't for her after all and that it's better to be alone than with someone who doesn't truly love you.}\\
\texttt{\quad \quad a. Jenna Adams overhears a conversation between Brian Johnson and his ex-girlfriend, Teresa Campbell.}\\
\texttt{\quad \quad b. Jenna Adams confronts Brian Johnson about the conversation and Brian Johnson confesses that he still has feelings for Teresa Campbell.}\\
\texttt{\quad \quad c. Jenna Adams breaks up with Brian Johnson.}\\
\texttt{\quad \quad d. Jenna Adams decides that it's better to be alone than with someone who doesn't truly love you.}\\
\bottomrule
\caption{Example of a premise and heavily abridged \oursimpl{} outline (settings, characters, and depth-3 items omitted; see Appendix \ref{sec:appendix_story_examples}, Table \ref{tab:plan2} for complete plan). %Sub-items under the top-level outline contain plot points to guide generation.
}
\vspace{-2em}
\label{tab:main_example_outline}
\end{tabular}
\end{table}

\begin{table}[!t]
\small
\begin{tabular}{@{}Y{0.95\linewidth}@{}}
\toprule
\textbf{\texttt{GENERATED STORY:}} \\
\texttt{\textcolor{gray}{...[85 words]...}}\\
\texttt{The first time Jenna saw him she stopped short in the middle of the aisle between bookshelves and looked up at him, her heart beating faster. \textcolor{gray}{...[331 words]...}}\\
\texttt{Jenna Adams wanted their relationship to go somewhere. \textcolor{gray}{...[106 words]...}}\\
\texttt{Maybe marriage wasn’t so bad after all. \textcolor{gray}{...[419 words]...}}\\
\texttt{[Jenna:] I love you, Brian Johnson. I want to be with you forever. I want you to give me a ring and ask me to marry you. \textcolor{gray}{...[811 words]...}}\\
\texttt{[Jenna:] I still love you, but I just cannot trust your promises anymore. \textcolor{gray}{...[222 words]...}}\\
\texttt{[Jenna:] I overheard the conversations that you had with Teresa Campbell \textcolor{gray}{...[122 words]...}}\\
\texttt{[Brian:] %I love you. 
I want you in my life forever. But I am confused about how I feel towards you and Teresa Campbell. \textcolor{gray}{...[285 words]...}}\\
\texttt{Jenna Adams then threw the ring into the fire pit that was in their backyard. She left Brian Johnson standing there in shock. \textcolor{gray}{...[244 words]...}}\\
\bottomrule
\caption{A heavily abridged \oursimpl{} story generated from the outline shown in Table \ref{tab:main_example_outline} (see Appendix \ref{sec:appendix_story_examples}, Table \ref{tab:ours_ex2} for complete story). Although some issues remain, the story has a coherent overarching plot which follows the outline.}
\vspace{-2em}
\label{tab:main_example_story}
\end{tabular}
\end{table}

% \medskip
\noindent\textbf{Results.} As shown in Table \ref{tab:main_results}, \oursimpl{} passages are judged dramatically more plot-coherent and outline-relevant compared to \rethreeimpl{}, not to mention the weak \rollingopt{}. The results confirm our intuition that plot coherence and outline relevance should benefit from shifting creative work from planning to drafting, together with improved control. Perhaps surprisingly, annotators also judged \oursimpl{}'s passages to be significantly more interesting, which ablations suggest is a result of our more detailed (and more eventful) outline (Section \ref{sec:analysis_ablations}). 

% \oursimpl{} enforces a certain level of creativity and eventfulness via the more detailed outline, which may be lacking in \rethreeimpl{}; additionally, we hypothesize that readers may naturally find more plot-coherent passages to be more interesting.

Of course, qualitative inspection reveals room for improvement. While \oursimpl{} usually does not deviate heavily from the top-level outline---unlike \rethreeimpl{}, which is sometimes almost completely off-topic---\oursimpl{} often fails to follow lower-level parts of the detailed outline (Section \ref{sec:analysis_detailed_relevance}). Long-range factual consistency also remains a problem in both \oursimpl{} and \rethreeimpl{}. 
Occasional errors in the detailed outline can be particularly damaging, resulting in  cascading errors during drafting. 
% Compared to the main story, consistency errors are rarer in \oursimpl{}'s detailed outline, likely due to the shorter length; however, when errors do appear in the outline, they can result in larger cascading errors during drafting. 
%Additionally, while the outline items used for generation in \oursimpl{} are substantially more concrete compared to those used in \rethreeimpl{}, they may still be overly vague at times; for example, a human author might choose to expand them further. On rarer occasions, the outline may also be over-expanded, resulting in unnecessary details or repetition. 
Additionally, outline leaves in \oursimpl{} are often inconsistent in level of detail: some remain too vague while others seem over-expanded. Moreover, the detected settings and characters at times seem incorrect or incomplete. %Methods for more consistent outline ``concreteness'' could be an interesting avenue for future exploration. 

Table \ref{tab:main_example_story} shows a heavily abridged story written by \oursimpl{} according to the (also heavily abridged) detailed outline in Table \ref{tab:main_example_outline}. See Appendix \ref{sec:appendix_story_examples} for complete, i.i.d. examples of \oursimpl{} plans and stories.  

\subsection{Human-Interactive Story Generation}\label{sec:experiments_control}

We additionally evaluate \oursimpl{} compared to \rethreeimpl{} in an interactive setting, focusing on human controllability. Unlike prior human-in-the-loop approaches which operate passage by passage~\cite{coenen2021wordcraft,lee2022coauthor}, we explore interaction at a higher-level planning stage, though in principle \oursimpl{} can also support passage-level interaction.

\medskip
\noindent\textbf{Experiment Setup.} The human writes a story premise, from which we generate an initial plan with only a top-level (depth-1) outline. The human then edits for up to 5 minutes. The resulting intermediate plan \vP{} is used in both \oursimpl{} and \rethreeimpl{}, which subsequently diverge. For \oursimpl{}, we extend \vP{} with depth-2 and then depth-3 outline items, with up to 5 more minutes of editing after generating each depth. %followed by up to 5 more minutes of human editing, and finally we generate depth-3 outline items followed by another up to 5 minutes of editing. 
For \rethreeimpl{} the human simply edits \vP{} for up to 10 more minutes. Thus both methods are allotted 15 minutes of total editing. %(although we ask them not to change the number of outline items, to maintain alignment between story passages for evaluation). 
% Thus the final plans for both methods are edited for up to 15 minutes in total. 
We then generate stories according to the final edited plans.

\medskip
\noindent\textbf{Metrics.} %Rather than repeat the previous metrics from our main experiments, 
We asked workers to label the following metrics specific to the interactive experience. %Thus we ask the human to annotate the following metrics. 

\begin{enumerate}[topsep=0pt,itemsep=-1ex,partopsep=1ex,parsep=1ex]
    \item \textit{Intent.} Which system's passage better followed their original intent as author.
    \item \textit{Control.} Which system's workflow they felt gave them more control. 
    \item \textit{Intuition.} Which system was more helpful or intuitive to work with. 
    \item \textit{Quality.} Which system they would choose to write another story, if prioritizing quality.
\end{enumerate}

The intent metric is passage-level, while all others operate on the complete story level. Annotators label which system is better for each metric, or no preference (Appendix \ref{sec:appendix_annotation_templates_human}). %See  for the exact instructions shown to humans. 

\begin{table}[htbp]
\small
\centering
\begin{tabular}{@{}lcccc@{}}
\toprule
\textbf{Method}          & \textbf{Intent }& \textbf{Control} & \textbf{Intuition} & \textbf{Quality}\\
\midrule
\rethreeimpl{}         &  17.3        &  \phantom{0}5.0    & \phantom{0}5.0 &  15.0        \\
\oursimpl{}           &  \textbf{80.0}     &  \textbf{80.0}      & \textbf{80.0}  &   \textbf{75.0}    \\
\bottomrule
\end{tabular}
\caption{\small Pairwise comparison of \oursimpl{} vs. \rethreeimpl{} on 20 human-interactive story generation runs. Humans judged faithfulness to authorial intent, control over generation, system intuitiveness, and story quality. Numbers indicate the percentage of responses in favor of each system, with ``no preference'' responses omitted. Bolding indicates significance with $p < 0.05$. \oursimpl{} is preferred by a wide margin on all metrics.}
\label{tab:human_control_results}
\vspace{-0.5em}
\end{table}

% \medskip
\noindent\textbf{Results.}
% We hypothesized that \oursabstract{}'s breadth-first, step-by-step outline expansion procedure would be more amenable to human interaction compared to \rethree{}. 
% Indeed, 
As shown in Table \ref{tab:human_control_results}, humans overwhelmingly preferred \oursimpl{}'s interaction paradigm to \rethreeimpl{} on all four of our human-interactive metrics: at least three-fourths indicated \oursimpl{} as superior on each metric. 
In optional free-form comments (Appendix \ref{sec:appendix_interactive_comments}), reactions to overall story quality vary widely from disappointed to pleased, but clearly indicate that \oursimpl{}'s stories are more faithful to the plot outline and authors' original intentions. The results confirm that \oursimpl{}'s more detailed outline and improved control during drafting lead to humans judging \oursimpl{} as more controllable and more faithful to authorial intent. 

% \todo{ summarize results + define metrics again in table caption}
% \todo{interpret results and make judgments about \oursimpl{} vs \rethreeimpl{} systems in this human-in-the-loop context. also add optional comments to appendix}

\section{Analysis}\label{sec:analysis}

% In this section we aim to improve our understanding of different aspects of \oursimpl{} by running variations of our method. 

\subsection{Ablation Study}\label{sec:analysis_ablations}

% \medskip
\noindent\textbf{Ablated Components.} To ablate the two main components of \oursimpl{}, we modify \oursimpl{} as follows:
% respectively, the detailed outline and the controlled generation when writing the main story. Specifically, we test the following systems:

\begin{enumerate}[topsep=0pt,itemsep=-1ex,partopsep=1ex,parsep=1ex]
    \item \oursimplonelevel{}, which generates only according to the top-level outline instead of the full detailed outline, using fixed passage length per outline item (instead of early stopping) and a fixed-strength \detcon{}. %This ablation is essentially \rethreeimpl{} augmented with a fixed-strength \detcon{}.%\footnote{We also considered an ablation where we generate trivial multi-level outlines of the same depth as \oursimpl{} without doing our filtering and reranking heuristics as described in Section \ref{sec:method_detailed_outlines}. However, the plans were obviously problematic so we didn't continue with a formal comparison (Appendix \ref{sec:appendix_trivial3level}).} 
    % To maintain consistent length, we generate a fixed number of 256-token passages per top-level outline item and remove the early stopping criteria, similar to \rethreeimpl{}. Due to removing early stopping, we fix the \fudge{} control strength throughout instead of incrementing over time for a given outline item. 
    \item \oursimplnofudge{}, which is identical to \oursimpl{} except the \detcon{} is turned off. 
\end{enumerate}

We reuse the same coherence, relevance, and interestingness metrics from Table \ref{tab:main_results}.

\begin{table}[htbp]
\small
\centering
\begin{tabular}{@{}lccc@{}}
\toprule
\textbf{Method}          & \textbf{Coherent} & \textbf{Relevant}& \textbf{Interesting }\\
\midrule
\oursimplonelevel{}               & 61.8     & 41.2    & 57.8         \\
\oursimpl{}                  & 73.5       & \textbf{64.7}   &  66.7         \\
\midrule
\oursimplnofudge{}               & 62.7     & 52.0     & 58.8        \\
\oursimpl{}                  & 70.6       & \textbf{73.5}   & 50.0         \\
\bottomrule
\end{tabular}
\caption{\small Pairwise comparisons of \oursimpl{} vs. ablations without the \detgen{} and \detcon{}, respectively, on passages from 10 stories. Bold indicates significance with $p < 0.05$. Although the results on plot-coherence and interestingness are inconclusive, both the \detgen{} and \detcon{} are important for outline relevance.}
\label{tab:ablations}
\vspace{-0.5em}
\end{table}

\medskip
\noindent\textbf{Results.} As shown in Table \ref{tab:ablations}, compared to both ablations, \oursimpl{} maintains significantly higher relevance to top-level outline items. Thus both the \detgen{} and \detcon{} meaningfully contribute to our method's ability to follow the high-level outline. Although the gaps in plot coherence and interestingness are not statistically significant, the ablations suggest that \oursimpl{}'s gain in interestingness compared to prior work is mainly due to the more detailed outline; if anything, the \detcon{} may slightly hurt interestingness. Indeed---perhaps unsurprisingly---we observe qualitatively that further increasing control strength yields increasingly narrowly-focused, repetitive outputs at the expense of creativity.

% \subsection{Passage-Level Beam Search}

% \todo{describe the passage-level beam search, + style reranker if we use it}

% \begin{table}[htbp]
% \small
% \centering
% \begin{tabular}{@{}lccc@{}}
% \toprule
% \textbf{Method}          & \textbf{Interesting }& \textbf{Coherent} & \textbf{Relevant}\\
% \midrule
% \oursimpl{}           &       &      &            \\
% \oursimplbeamtwo{}         &           &    &            \\
% \bottomrule
% \end{tabular}
% \caption{\small \todo{results based on 10 stories}}
% \label{tab:beam2}
% \vspace{-0.5em}
% \end{table}

% \todo{describe results}

\subsection{Detailed Relevance Evaluation}\label{sec:analysis_detailed_relevance}

We now examine \oursimpl{}'s faithfulness to the outline at the leaves instead of at the top level. For each leaf-node outline item, we ask one annotator whether the event specified in the leaf occurred in either the corresponding passage or in the immediately preceding and following passages (Appendix \ref{sec:appendix_annotation_templates_detailed_relevance}). We do the same for \oursimplnofudge{}. %, though we do not compare pairwise since we view this task as relatively more objective due to the shorter passages.% \todo{and beam 2?}. 

% We view this task as relatively more objective due to the shorter passages and more concrete events, so we label each method individually instead of doing pairwise comparison, and only collect one label per example.

% Because these passages are shorter compared to our main experiments, we view this annotation as relatively more objective compared to our main experiment metrics, so we label each method by itself (rather than doing pairwise comparison between methods) and only collect one label per outline-passage pair. %See Appendix \ref{sec:appendix_annotation_templates} for an example annotation task. 

\begin{table}[htbp]
\small
\centering
\begin{tabular}{@{}lccc@{}}
\toprule
\textbf{Method}    & \textbf{Detailed-Relevant}\\
\midrule
\oursimplnofudge{}         &   37.8       \\
\oursimpl{}           &     \textbf{58.5}     \\
% \oursimplbeamtwo{} & \\
\bottomrule
\end{tabular}
\caption{\small Percentage of short passages that are faithful to corresponding outline leaf nodes, ablating the \detcon{}. Bold indicates significance with $p < 0.05$. The \detcon{} greatly improves relevance to leaf nodes.}% \todo{describe results, including for beam 2; significance/bolding}}
\label{tab:detailed_relevance}
\vspace{-0.5em}
\end{table}

% \todo{detailed outline items relevance table for 10 stories, \oursimpl{} and \oursimplnofudge{}.}
% \todo{include beam size 2 later?}

\medskip
\noindent\textbf{Results.} Table \ref{tab:detailed_relevance} confirms that the \detcon{} substantially improves \oursimpl{}'s ability to follow low-level outline details during drafting. 

However, the overall numbers remain low, pointing to two issues. First, the outline leaf itself may be problematic: it may be unexpected in context, or overly vague. Second, the \detcon{} may be unable to sufficiently steer the generation without further increasing control strength (which may sacrifice fluency). 
% In practice, we believe both issues contribute to the low rate of following the detailed outline items. 
Thus, while \oursimpl{} is substantially more faithful to the outline compared to baselines, a good deal of headroom remains.% for further improving lower-level outline control.
% Thus, while \oursimpl{} has substantially improved ability to follow the top-level outline compared to the previous \rethree{} system, there remains a good deal of headroom for further improving lower-level outline control. 

% \todo{analyze beam size 2 performance? }
% \todo{mention differences in length between ours, nofudge}

\section{Discussion}

We have presented the \oursabstract{} framework for improving long-range coherence in long-form story generation. \oursabstract{} uses a \detgen{} to shift creative work from drafting to planning, and employs a \detcon{} to maintain faithfulness to the detailed outline during drafting.
Compared to the prior state-of-the-art, \rethree{}, \oursabstract{} dramatically improves the plot-coherence,
outline relevance, and even interestingness of generated stories according to human annotators. 
Nevertheless, there remain many interesting future directions.

\medskip
\noindent\textbf{Other Text Domains.} We have focused on creative stories in this work, but we believe many of our high-level ideas could be applicable to other long-form text generation settings, such as Wikipedia articles or movie scripts. Generation in such settings could potentially benefit from detailed planning via an outline, combined with additional control to maintain faithfulness to the initial plan. Of course, many of our specific prompts would require substantial modification to adapt to a new domain.

\medskip
\noindent\textbf{Improved Human Interaction.} In Section \ref{sec:experiments_control} we experimented with \oursimpl{} in a human-interactive setting, enabling the human to interact with \oursimpl{} at a high-level planning stage, in contrast to previous works which operated at the drafting level~\cite{coenen2021wordcraft,lee2022coauthor}. We are excited to continue exploring novel forms of human interaction that become possible as automated generation capabilities continue to improve. % however there's a lot more room to improve here as systems get better and we can explore new forms of interaction that make most sense for the author

\medskip
\noindent\textbf{Scaling to Longer Texts.} While our stories (exceeding 3500 words on average) are lengthy by neural text generation standards, they remain relatively short by human authors' standards. We hope to eventually develop systems which can scale to full-length novels. We believe \oursabstract{}
makes an important contribution toward this ambitious goal by generating outlines with granularity scalable to story length, 
% by providing a recursive mechanism to generate hierarchical outlines with granularity commensurate with the desired scope of the final generated story, 
while also providing better control mechanisms to maintain faithfulness to the outline during drafting. However, there remain major barriers to high-quality longer generations, two of which we describe below.

\medskip
\noindent\textbf{Evaluation.} 
% Compared to generation tasks with shorter outputs, such as translation or summarization, 
% strong automated metrics for long-form generation which correlate well to human judgments are relatively lacking. 
While some recent works have suggested metrics for longer generations~\cite{castricato2021towards,matiana2021cut}, there is currently no substitute
for human judgments for our metrics in this work, due to the sheer length of evaluated passages
and complexity of our metrics. For example, it is unclear how one might automatically measure overarching plot coherence,
or especially interestingness. However, automated metrics for relevance may be more tractable,
especially as applied to our more fine-grained experiments on low-level outline items with shorter passages (Section \ref{sec:analysis_detailed_relevance}).
% , where the task is relatively objective:
% the annotator simply needs to indicate whether the given event unambiguously occurred in the passage. 
To facilitate such efforts, we have open-sourced all annotations
collected during our experiments in our public GitHub repository, in hopes that they prove useful for developing improved metrics for long-form generation. 

\medskip
\noindent\textbf{Long-Range Consistency.} A second major problem is internal consistency over long passages,
of which one major component is factual consistency. While more detailed outlines may help somewhat in this respect, we have largely not focused on factual consistency in this work. \oursimpl{}'s stories 
occasionally contain glaring errors, e.g., inconsistent names or genders, and errors sometimes occur even 
during outlining, leading to cascading errors during drafting. Moreover, we have not yet mentioned 
non-factual aspects of long-range consistency besides overarching plot coherence.
Such aspects include maintaining consistent story pacing, or literary devices such as foreshadowing, 
which are themselves interesting directions for exploration. %We believe focused efforts toward improving long-range consistency of different kinds would greatly benefit future long-form generation methods.

\newpage

% ofir idea of simplifying the prompts - it's a bit complex and a lot of different prompts, might make it difficult to transfer to other domains in current form for example

\section*{Limitations}

As with previous work on long-form text generation, it is difficult to evaluate the quality of our story outputs without resorting to expensive human annotations. Although we have ablated the main components of \oursabstract{}, the difficulty of evaluation limits us from running more detailed ablations on sub-components, which might help us to better streamline the framework which currently contains many different interacting pieces. 

% Additionally, our system is highly specialized for story generation in English. While we believe our high-level ideas---detailed outlining and detailed control---would be applicable and useful to other long-form text generation domains as well, some low-level details of \oursabstract{}---such as the setting and character detection in detailed outlines---may be more specific to the story generation domain, and may not transfer as well to other tasks. Similarly, many of our prompts are carefully engineered in English, and it may be cumbersome to adapt \oursabstract{} to work in other languages. 

Additionally, our system is highly specialized for story generation in English. While we believe our high-level ideas---detailed outlining and detailed control---are broadly applicable, adaptation to different text domains or languages would require substantial prompt modification.
\section*{Ethical Considerations}

Strong automated systems for natural language generation have the potential for harm, for instance by generating toxic
or untruthful text. In this work, we focus on creative stories, limiting the potential
for abuse. Although we have not explicitly attempted to decrease
the likelihood of harmful text in this work, \oursabstract{} is built to be modular with respect to the base
language models we depend on, so advancements in those systems can in principle be transferred to \oursabstract{} as well. 
Additionally, 
controlled generation schemes can be used to reduce output toxicity, similar to how we used \fudge{} in this work to control for outline relevance. %In fact, controlling against toxicity may be simpler if it does not require handling natural language instructions. 

\oursabstract{} is currently designed only for English; transferring to other languages would require adapting our prompts. Performance might suffer in lower-resource languages, as we depend heavily on large pretrained language models which may perform worse
on such languages.

\section*{Acknowledgments}

We thank the Berkeley NLP group, our colleagues at Meta AI, and our anonymous reviewers for their helpful discussions and feedback. 
This work was
supported by Berkeley AI Research, Meta AI, Open Philanthropy, DARPA 
under the SemaFor program (HR00112020054), the Machine
Common Sense (MCS) program under Cooperative
Agreement N66001-19-2-4032, and the NSF through
a fellowship to the first author. The content does
not necessarily reflect the position or the policy
of the government, and no official endorsement
should be inferred.

\bibliography{anthology,custom}
\bibliographystyle{acl_natbib}

\newpage

\appendix

% \onecolumn

\section{Filtering and Reranking Details}\label{sec:appendix_order_model}

For filtering candidate outline events, we enforce that outline events should be declarative sentences, have proper capitalization at the beginning, contain no uncommon punctuation symbols (e.g., ``<''), not be overly repetitive compared to pre-existing events in the outline (other than the current event's direct ancestors) based on edit distance and the entailment model of \citet{laurer2022less}, and be between 3 and 50 tokens long.

Sentence similarity for reranking uses the model provided at \url{https://huggingface.co/sentence-transformers/all-mpnet-base-v2}.

To train the ordering model, we collected a dataset of 1000 very brief stories of two to three paragraphs written by InstructGPT3-175B (\texttt{text-davinci-002}), as we observed the stories produced by InstructGPT3-175B are conveniently written in a high-level outline-like style---essentially, ``telling'' rather than ``showing.'' We trained a model based on \texttt{roberta-large}~\cite{liu2019roberta} that predicts whether a given sentence in such a story appears in the correct order by training contrastively, with negatives constructed by randomly moving the given sentence to elsewhere in the story. %\todo{cite DINO if relevant?}

\section{Example Structured Prompts}\label{sec:appendix_example_prompts}

We show some real examples of structured prompts used in our \detgen{} and during drafting.

\subsection{Event Descriptions}\label{sec:appendix_example_prompts_outlineitem}

Table \ref{tab:prompt_outline_item} shows a prompt for generating one outline item's event description near the end of generation at depth 3. 

\begin{table*}[!htbp]
\small
\begin{tabularx}{\linewidth}{X}
\toprule
% \textbf{Example Outline Item Generation Prompt}\\
% \midrule
\textbf{Prefix:}\\
\texttt{Premise: After the loss of her father, Shannon is determined to follow in his footsteps and become a successful journalist. However, when she lands her first major assignment, she quickly discovers that the ugly reality of life in the city is far different from the dream she imagined. With the help of her new friend, a street-wise teenager, Shannon comes to understand the harsh realities of life in the inner city and learns that sometimes the truth is much more than just a story.}\\
\\
\texttt{Setting: The story is set in the inner city of a large metropolitan area.}\\
\\
\texttt{Characters:}\\
\\
\texttt{Shannon Doyle is a young woman in her early twenties.}\\
\texttt{Gary Saunders is a teenage boy who lives in the inner city.}\\
\texttt{Mike Doyle is Shannon's father and a successful journalist.}\\
\texttt{Lena Saunders is Gary's mother and a local business owner.}\\
\texttt{Eddie Saunders is Gary's older brother and a gang member.}\\
\texttt{Dexter Brown is a local drug dealer.}\\
\texttt{News Director is Shannon's boss at the television station.}\\
\texttt{Jamal Walker is a teenage boy who is a member of Eddie's gang.}\\
\texttt{Ernesto Jimenez is a police detective who is investigating a string of murders in the inner city.}\\
\texttt{Luis Chavez is a reporter who works with Shannon at the television station.}\\
\\
\texttt{Outline:}\\
\\
\texttt{1. Shannon's father, Mike, dies unexpectedly, leaving her determined to follow in his footsteps and become a successful journalist.}\\
\\
\texttt{\quad \quad a. Shannon's father, Mike, dies unexpectedly.}\\
\\
\texttt{\quad \quad b. Shannon decides to follow in her father's footsteps and become a successful journalist.}\\
\\
\texttt{2. Shannon lands her first major assignment, a feature on the inner city, but quickly discovers that the ugly reality of life in the city is far different from the dream she imagined.}\\
\\
\texttt{\quad \quad a. Shannon lands her first major assignment, a feature on the inner city.}\\
\\
\texttt{\quad \quad \quad \quad List the main events that occur under this heading, starting from the beginning.}\\
\\
\texttt{\quad \quad \quad \quad i.}\\
\\
\textbf{Suffix:}\\
\texttt{\quad \quad \quad \quad ii. Shannon quickly discovers that the ugly reality of life in the city is far different from the dream she imagined.}\\
\\
\texttt{\quad \quad c. With the help of her new friend, Gary, Shannon comes to understand the harsh realities of life in the inner city and learns that sometimes the truth is much more than just a story.}\\
\\
\texttt{\quad \quad \quad \quad i. Shannon meets Gary.}\\
\\
\texttt{\quad \quad \quad \quad ii. Gary teaches Shannon about the inner city.}\\
\\
\texttt{\quad \quad \quad \quad iii. Shannon learns that the truth is much more than just a story.}\\
\bottomrule
\caption{Example prompt showing the exact prefix and suffix for generating a depth 3 outline item. Note that the suffix is shifted in depth for prompting purposes only so that it begins at the same depth as the current outline item that we are generating (i.e., the suffix shown here corresponds to 2b, 3, 3a-c in the completed outline in Table \ref{tab:plan1}). We observed this depth-shifting to improve coherence, though this may cease to be necessary with improved language models in the future. The prefix and suffix together include all previously generated ancestor nodes of the current outline item, together with those ancestors' respective children, thus providing relevant context while also maintaining scalability to higher depth.}
\label{tab:prompt_outline_item}
\end{tabularx}
\end{table*}

% \FloatBarrier

\subsection{Setting and Character Detection}\label{sec:appendix_example_prompts_scenechar}

\noindent\textbf{Setting.} For implementation convenience in practice, since other parts of the detailed outline do not depend on the setting, the setting is generated for each leaf node in depth-first order after the rest of the outline is complete. The prompt for generating a setting for a given outline item is similar to that used for the event, but also includes previously generated settings. An example prompt is shown in Table \ref{tab:example_scene_prompt}.

\begin{table*}[!htbp]
\small
\begin{tabularx}{\linewidth}{X}
\toprule
\textbf{Prefix:}\\
\texttt{Sherry had the perfect life--three healthy children, a loving wife, and a job to support them; until she discovers what was happening right in front of her. Sherry's wife has been cheating on her with her brother ever since they've been together and she's been too blind to see it. A bitter divorce ensues and Sherry is left to raise her children on her own. Broken and heartbroken, Sherry swears off love entirely...until she meets someone who makes her question everything she thought she knew.}\\
\\
\texttt{The story is set in the present day, in a small town in the United States.}\\
\\
\texttt{Sherry Jackson is a middle-aged woman who is struggling to get over her divorce.}\\
\\
\texttt{Melissa Jackson is Sherry's ex-wife who cheated on her with her own brother.}\\
\\
\texttt{Brad Jackson is Sherry's ex-husband's brother and her former lover.}\\
\\
\texttt{Lena Edwards is a woman who Sherry meets after her divorce who helps her to heal and move on.}\\
\\
\texttt{Abigail Jackson is one of Sherry's three children.}\\
\\
\texttt{Caleb Jackson is one of Sherry's three children.}\\
\\
\texttt{Sophia Jackson is one of Sherry's three children.}\\
\\
\texttt{Luke Edwards is Lena's son who befriends Sherry's children.}\\
\\
\texttt{Steven Warner is Sherry's boss who she starts dating after her divorce.}\\
\\
\texttt{Outline:}\\
\\
\texttt{Sherry's life falls apart when her wife cheats on her with her brother and she gets divorced.}\\
\\
\texttt{\quad \quad a. Sherry's wife cheats on her with her brother.}\\
\\
\texttt{\quad \quad \quad \quad i. Sherry's wife cheats on her with her brother. This scene is located in}\\
\\
\textbf{Suffix:}\\
\texttt{\quad \quad \quad \quad ii. Sherry finds out about the affair.}\\
\\
\texttt{\quad \quad \quad \quad iii. Sherry confronts her wife about the affair.}\\
\\
\texttt{\quad \quad b. Sherry gets divorced.}\\
\\
\texttt{\quad \quad \quad \quad i. Sherry and her wife get divorced.}\\
\\
\texttt{\quad \quad \quad \quad ii. Sherry gets custody of her three children.}\\
\\
\texttt{\quad \quad \quad \quad iii. Sherry's ex-wife moves away with her brother.}\\
\\
\texttt{Lena helps Sherry to heal and move on from her divorce.}\\
\\
\texttt{\quad \quad a. Lena helps Sherry to heal from her divorce.}\\
\\
\texttt{\quad \quad b. Lena and Sherry become friends.}\\
\\
\texttt{Sherry starts dating her boss, Steven Warner.}\\
\\
\texttt{\quad \quad a. Sherry starts dating her boss.}\\
\\
\texttt{\quad \quad b. Steven and Sherry get married.}\\
\bottomrule
\caption{Example prompt for detecting setting for a given outline item, after the non-setting parts of the detailed outline are complete.}
\label{tab:example_scene_prompt}
\end{tabularx}
\end{table*}

\FloatBarrier

\medskip
\noindent\textbf{Character.} Character detection, operating in tandem with the event generation procedure for each outline item, is more involved. After generating the event for a given outline item, we first prompt for a list of possibly unnamed characters (Table \ref{tab:prompt_unnamed_characters}), allowing the model to continue generating the list if the most recently generated name contained the next number in the list (i.e., if the model generates ``Shannon 2. ...'' for the prompt in Table \ref{tab:prompt_unnamed_characters}, we save ``Shannon'' as the first detected character, and take the presence of the string ``2.'' as an indication that we should continue detecting more characters). 

\begin{table*}[!htbp]
\small
\begin{tabularx}{\linewidth}{X}
\toprule
\texttt{Shannon decides to follow in her father's footsteps and become a successful journalist.}\\
\\
\texttt{List all characters mentioned in this sentence.}\\
\\
\texttt{1.}\\
\bottomrule
\caption{Initial prompt for detecting (possibly unnamed) characters in an outline item.}
\label{tab:prompt_unnamed_characters}
\end{tabularx}
\end{table*}

Characters mentioned by name are directly matched against our character inventory based on word overlap.

For remaining unnamed character strings, we first detect if they refer to a single character or a group of characters. For example, if we want to match ``her father'' in the outline item shown in Table \ref{tab:prompt_unnamed_characters}, we would first detect whether this string refers to a single character or group using the prompt shown in Table \ref{tab:prompt_single_group}, followed by checking whether the token `` single'' or `` group'' has higher next-token probability.

\begin{table*}[!htbp]
\small
\begin{tabularx}{\linewidth}{X}
\toprule
\texttt{Shannon decides to follow in her father's footsteps and become a successful journalist.}\\
\\
\texttt{In this passage, is her father a single character or a group of characters?}\\
\\
\texttt{her father is a}\\
\\
\texttt{1.}\\
\bottomrule
\caption{Prompt for detecting whether an unnamed character string (``her father'') refers to a single character or group of characters.}
\label{tab:prompt_single_group}
\end{tabularx}
\end{table*}

\begin{table*}[!htbp]
\small
\begin{tabularx}{\linewidth}{X}
\toprule
\texttt{Full Name: Calvin Klein Calvin Klein is a well-known fashion designer.}\\
\\
\texttt{Full Name: Rachel Wu Rachel Wu is a journalist who covers Fashion Week for a popular fashion magazine.}\\
\\
\texttt{Full Name: Mia Zhang Mia Zhang is a supermodel who wears Angie's dress during Fashion Week.}\\
\\
\texttt{Full Name: Lily Li Lily Li is Angie's mother.}\\
\\
\texttt{Full Name: Andrew Wang Andrew Wang is Angie's father.}\\
\\
\texttt{Full Name: Viktor Kaminsky Viktor Kaminsky is a Russian oligarch who is interested in purchasing the design house where Angie works.}\\
\\
\texttt{Full Name: Dmitri Gregorovich Dmitri Gregorovich is Viktor Kaminsky's right-hand man. He is in a top design house.}\\
\\
\texttt{Full Name: Owen Shaw Owen Shaw is Angie's boss at the design house where she interned.}\\
\\
\texttt{Full Name: Angie Wang Angie Wang is a twenty-two year old Chinese-American woman. Angie Wang is a designer. She is an intern. Angie works at a design house. She is a best friend and roommate of Jen Chen.}\\
\\
\texttt{Full Name: Jen Chen Jen Chen is Angie's best friend and roommate.}\\
\\
\texttt{----------------------------------}\\
\\
\texttt{The characters in the following context include: Angie Wang, Dmitri Gregorovich.}\\
\\
\texttt{Previous context: Angieinterns at a top design house for a year. Angie interns at a top design house for a year.}\\
\\
\texttt{Current passage: She meets her best friend and roommate, Jen Chen.}\\
\\
\texttt{best friend's full name:}\\
\bottomrule
\caption{Prompt for determining the character name corresponding to a character string (``best friend'') which has been predicted to correspond to a single character.}
\label{tab:prompt_single}
\end{tabularx}
\end{table*}

If the character is a single character, we then provide our character inventory as context together with some previous outline nodes (if they exist) to resolve potential coreferences, as shown in Table \ref{tab:prompt_single}, followed by parsing the output for a name that matches our character inventory. The characters in the inventory are given in reverse order of predicted relevance based on their descriptions' similarities compared to the context, according to a sentence similarity model~\cite{reimers2019sentence}. Note when we provide the character inventory, we leverage the descriptions from our updated character descriptions over time, to improve matching; an example can be seen under the description of Angie Wang in Table \ref{tab:prompt_single}. For strings which represent groups of characters, the prompt is nearly identical, except we allow the model to generate up to two characters one at a time in a list, similar to how we generated multiple unnamed character strings initially. (While it may be desirable to generate more than two characters for the group in some cases, we observed that the model would frequently hallucinate additional characters instead of stopping appropriately if we did not enforce a maximum of two characters.)

% \todo{generate a new plan with this logging}

We allow a maximum of 5 characters to be detected per outline item. 

% \FloatBarrier

\subsection{Character Development Over Time}\label{sec:appendix_example_prompts_chardesc}

Whenever we detect that a character appears in a given outline item, we attempt to update the character's description with a new string which will appear whenever we query for the character again while processing any later outline item (but not for earlier outline items). 

The new description is generated based on the new outline item and the preexisting character description as shown in the prefix and suffix respectively of the example prompt in Table \ref{tab:prompt_add_description}. The newly generated description is added to the description only if it is not already entailed by a preexisting description; additionally, if the new description entails a preexisting description, then the preexisting description will be removed whenever the new description is used (i.e., at the current outline item or later). 

\begin{table*}[!htbp]
\small
\begin{tabularx}{\linewidth}{X}
\toprule
\textbf{Prefix:}\\
\texttt{Angie's design hits the runway at New York Fashion Week.}\\
\\
\texttt{This context tells us the following about Angie Wang:}\\
\\
\texttt{1.}\\\\
\textbf{Suffix:}
\\
\texttt{Additionally, we know from elsewhere that Angie Wang is a twenty-two year old Chinese-American woman. Angie Wang is a designer. She is an intern. Angie works at a design house. She is a best friend and roommate of Jen Chen. She is designing clothes.}\\

\bottomrule
\caption{Prompt for adding more information to the description of a character.}
\label{tab:prompt_add_description}
\end{tabularx}
\end{table*}

% An example of character descriptions mid-outline can also be seen in the suffix to the prompt in Table \ref{tab:prompt_add_description}

\FloatBarrier

\subsection{Example Prompt During Drafting}\label{sec:appendix_example_prompts_drafting}

Finally, in Table \ref{tab:drafting_prompt} we show an example of a prompt for generating the next story passage during drafting.

\begin{table*}[!htbp]
\small
\begin{tabularx}{\linewidth}{X}
\toprule
\texttt{Premise: The townspeople of Mayberry rally around Daisy and help her through her treatment. Daisy's treatment is difficult and the townspeople continue to support her.}\\
\\
\texttt{This book was authored by a well-known novelist, and received glowing reviews from critics, who praised the interesting dialogue and interactions between characters.}\\
\\
\texttt{Relevant Context:}\\
\\
\texttt{Daisy Mayberry is a kind-hearted woman in her early 50s who is loved by everyone in her small town. Daisy Mayberry has cancer. Elizabeth and Bill Simpson are Daisy's daughter and husband, respectively. Daisy Mayberry owns a hardware store. She has a daughter named Lisa.}\\
\\
\texttt{Charles Grayson is Andrea's husband and the town's financial advisor.}\\
\\
\texttt{Previous story summary: Daisy Mayberry receives a diagnosis of cancer and her family and friends come together to support her. Daisy's daughter, Lisa, becomes her primary caregiver and works tirelessly to find a treatment that will save her mother's life. Daisy begins her treatment and the townspeople rally around her. The townspeople help Daisy with her treatment and offer their support. Daisy's treatment is difficult and the townspeople offer their support.}\\
\\
\texttt{Events immediately prior to the upcoming passage: Lisa has been through a lot recently, and it has not been easy for her. Daisy is her mother and knows how to take care of her, even when Lisa is not feeling well. Daisy asks Lisa if she can stay with her tonight so that they can talk about what happened in Lisa's office. There has been a lot of activity going on around them, and it seems as though everyone is busy.}\\
\\
\texttt{The characters currently in the scene are Lisa Chambers, Daisy Mayberry, Evelyn Chambers, Maggie Simpson.}\\
\\
\texttt{In the upcoming passage, Daisy's treatment is difficult and the townspeople offer their support. The townspeople help Daisy through her treatment and she ultimately beats cancer. Daisy ultimately beats cancer and her story inspires hope in her community.}\\
\\
\texttt{This part of the story initially takes place in the hospital. The characters then move to Daisy's home.}\\
\\
\texttt{Full text below:}\\
\\
\texttt{--------------------------------}\\
\\
\texttt{However, the commotion abruptly ended as they entered the hallway and walked by without a glance at them.}\\
\\
\texttt{In the quiet of the hallway, they made their way down to Lisa’s office and took seats across from each other on her sofa. “Okay, tell me everything you know about this trial,” Lisa said as she picked up her laptop and began turning it on.}\\
\\
\texttt{Daisy sighed heavily as she sat back in the love seat and pulled her feet up underneath her. She pulled both hands through her hair in frustration, and then started talking. “I really don’t know much about it except that the doctor said it is an experimental treatment for people with the particular type of lung cancer I have. He told me that he was sending me to Memorial Hospital in St. Louis for an evaluation before I could be enrolled in the trial. He said he had been contacted by a research committee at the hospital and that they would meet me and evaluate me. I’m supposed to leave tomorrow at noon,” she said as she leaned back and covered her eyes with her hand.}\\
\\
\texttt{Lisa sat behind her desk and folded her hands in front of her.}\\
\bottomrule
\caption{Prompt for story passage, partway through drafting. ``Premise'' includes context from the ancestors of the current leaf. ``Relevant Context'' includes information about characters predicted to appear in the following passage, with inferred facts up to the current point in time. ``Previous story summary'' is a far-past summary containing prior outline items, with previous sections collapsed into lower-depth items where possible. ``Events immediately prior to the upcoming passage'' is a near-past summary of several preceding paragraphs. ``Characters currently in the scene'' are characters from the previous passage. ``In the upcoming passage'' describes the previous, current, and subsequent outline items for context, although the \detcon{} will only apply to the current outline item (``The townspeople help Daisy through her treatment and she ultimately beats cancer''). Finally, there is a setting description, including description of a change in setting if applicable, followed by the immediately preceding story passage reproduced verbatim.}
\label{tab:drafting_prompt}
\end{tabularx}
\end{table*}

\FloatBarrier

\section{Additional Metrics Discussion}\label{sec:appendix_metrics_discussion}

\citet{yang2022re3} use two additional metrics, which we omit in our experiments. Their ``miscellaneous writing problems'' metric (jarring narration/style, inconsistency, confusing writing, grammatical disfluency, repetitiveness) measures an axis orthogonal to our main contributions, and we did not expect much change in \oursimpl{} compared to the original \rethreeimpl{} (Table \ref{tab:misc_writing_problems}). Their ``humanlike'' metric varies heavily by annotator population: in preliminary experiments, we found that workers on Amazon Mechanical Turk predicted 70-80\% of stories to be human-written, compared to just 30\% on Surge AI. Therefore, we focus on the coherence, relevance, and interestingness metrics in the main text, modified to operate on passages instead of complete stories to reduce noise. 

\begin{table}[htbp]
\small
\centering
\begin{tabular}{@{}lc@{}}
\toprule
\textbf{Method}          & \textbf{Misc. Writing Problems$\downarrow$}\\
\midrule
\rethreeimpl{}                  & 1.17      \\
\oursimpl{}                  & 1.00    \\
\bottomrule
\end{tabular}
\caption{\small Average number of writing problems as defined by \citet{yang2022re3} indicated by annotators in 20 stories from our main experiments (fewer is better). \oursimpl{} performs equal or better compared to \rethreeimpl{} on this metric, although we didn't expect much difference since these writing problems measure a direction orthogonal to our main contributions.}
\label{tab:misc_writing_problems}
\vspace{-0.5em}
\end{table}

\FloatBarrier

\section{GPT3 vs. OPT Base Generator}\label{sec:appendix_gpt3_discussion}

Technically, our approach is compatible with the public GPT3 API,
but it is computationally impractical due to the
limited functionality supported in the API: for each token, to continue generation after modifying output logits, we need to re-query the API and re-process
the entire preceding prompt. 
Therefore, during drafting we use OPT-175B as served by the Alpa project~\cite{zheng2022alpa},
which supports restarting generation from cached key values for the
previously processed prompt; this caching is the only additional feature we need. As language models continue to improve, it may become possible to use smaller models for better computational efficiency as well, such as LLAMA~\cite{touvron2023llama}.

Although OPT has been observed to perform somewhat worse than GPT3 on many tasks~\cite{iyer2022opt}, as a story passage generator in our experiments we found OPT to write similar-quality outputs upon manual inspection. A formal comparison using \rollinggpt{}, an identical baseline to \rollingopt{} except using GPT3 instead of OPT, reveals that both remain dramatically worse compared to \oursimpl{} (Table \ref{tab:gpt3_vs_opt}). If anything, perhaps \rollinggpt{} is only a little more interesting compared to \rollingopt{}. 

\begin{table}[htbp]
\small
\centering
\begin{tabular}{@{}lccc@{}}
\toprule
\textbf{Method}          & \textbf{Coherent} & \textbf{Relevant}& \textbf{Interesting }\\
\midrule
\rethreeimpl{}                  & 45.1     & 37.1     & 39.4       \\
\oursimpl{}                  & \textbf{67.6}       & \textbf{65.3}    & \textbf{60.1}        \\
\midrule
\rollingopt{}                  & 38.0 & 25.4&  25.4   \\
\oursimpl{}                   & \textbf{80.8}& \textbf{78.9}&  \textbf{69.5}       \\
\midrule
\rollinggpt{}                  & 44.1& 25.8&  42.7      \\
\oursimpl{}               & \textbf{81.7}& \textbf{83.1}& \textbf{70.0}        \\
\bottomrule
\end{tabular}
\caption{\small A version of Table \ref{tab:main_results} which additionally includes the \rollinggpt{} baseline. Bold indicates significance with $p < 0.05$.}
\label{tab:gpt3_vs_opt}
\vspace{-0.5em}
\end{table}

We note that our setup uses \textit{substantially} longer prompts and also fairly long outputs compared to tasks used in common benchmark suites, i.e., our task could be considered ``out of domain'' in some sense relative to common NLP benchmarks. In particular, as observed previously in \citet{yang2022re3}, instruction-tuned models such as InstructGPT (\texttt{text-davinci-002}) may actually perform \textit{worse} than the non-instruction-tuned models (\texttt{davinci}) as story passage generators, simply because they are tuned for a different distribution (i.e., common human interactions) compared to what we require for story generation. We also tested the newly released \texttt{text-davinci-003}, which we found could produce higher-quality outputs. However, in preliminary experiments we struggled to generate stories of more than 600-700 words, and observed a tendency to revert back to a higher-level ``summary-like'' style appropriate for much shorter stories compared to what we aim for in this work. GPT-4 seemed to bring further improvement, but not qualitatively so. Structured planning approaches are still necessary to generate longer text on the range of thousands of words, such as in \citet{atlantis} which generates a relatively simple novel using GPT-4 with some minimal human guidance. In any case, advancements in language modeling are orthogonal to our contributions, and we are excited to explore applications of more advanced language models in future long-form story generation systems. % (Appendix \ref{sec:appendix_textdavinci003}). 
% In any case, advancements in language modeling are orthogonal to our contributions, and we are excited to explore applications of more advanced language models in future long-form story generation systems.

% hen running the \rethree{} baseline with OPT-175B instead of GPT3-175B,
% % \footnote{\rethree{} uses the base GPT3 rather than InstructGPT3 for the main story generation, i.e., ``davinci'' rather than ``text-davinci-002''}
% we did not observe an obvious difference in quality upon inspection.

\FloatBarrier

\section{\oursimpl{} Additional Implementation Details and Hyperparameters}\label{sec:appendix_hyperparameters}

\noindent\textbf{Length and Early Stopping.} For length, we allow the outline to have a maximum depth of 3. We allow generating at most 8 consecutive 64-token passages per outline item, i.e., the maximum number of generated tokens per outline item is 512. Whenever we generate a 64-token passage, we truncate the last incomplete paragraph if we are fewer than 10 tokens into the start of a new paragraph.

For early stopping we move to the next outline item if the combined log-probability scores of the relevance and coherence rerankers exceed -0.5 and the scores do not improve further. That is, if at any step we see that the previous passage had combined relevance and coherence log-probabilities exceeding -0.5 according to our rerankers, and the current passage does not further improve the score, we stop at the end of the previous passage and move on to the next outline item. We additionally skip the current passage and directly move on to the next outline item in the rare case where all candidate passage extensions are problematic according to simple heuristics (e.g., highly repetitive). 

When reranking story passages at any given step, we generate 8 candidates at a time. 

\medskip
\noindent\textbf{\detgencaps{}.} We attempt to generate up to 10 characters for our initial inventory of characters before drafting the outline, though we do not always achieve the full 10 due to \rethreeimpl{}'s filtering heuristics for valid names. After detailed outline generation we remove characters which were not detected to appear anywhere in the outline. 
We generate 10 possible event candidates for each outline node when filtering and reranking. When generating children for each parent node, we restart and resample if there are fewer than 2 or more than 5 children.

\medskip
\noindent\textbf{\detconcaps{}.} For control strength of the event description, we increment the \fudge{} control strength by 3 for each passage generation substep within a single outline item, starting at 0 and capped at 10. Control strength for new settings (i.e., changed setting from previous outline item) is set to 0.5 times the control for the event description, and 0.2 times for new characters (i.e., characters that did not appear in the previous outline item). \fudge{} considers the top 100 tokens according to the base generator, so we are approximately running top-$k$ sampling with $k=100$. 

\medskip
\noindent\textbf{Base Generator.} When using OPT-175B, we use a frequency penalty of 1. Unlike in the GPT3 API, the penalty additionally includes the full prompt. The reason to do so is because there is significant scaffolding text in the prompt and we find that including the prompt in the penalty decreases repetitiveness in generation; additionally, we observe that OPT-175B is often more repetitive with smaller penalties. However, also unlike in the GPT3 API, our penalty decays exponentially at a rate of 0.98 per token, in order to avoid e.g., overly penalizing stopwords during longer generations. 

The temperature for the OPT generator is set to 0.8 while generating the main story. The temperature for InstructGPT3 is set to 1.2 when generating both initial character names and detailed outline events in order to increase diversity; we additionally increment the temperature by 0.1 each time for up to two more attempts when outline expansion fails for a given parent node during detailed outlining.

The same OPT-175B hyperparameters are used in the \rethreeimpl{} and \rollingopt{} baseline implementations where applicable.

\FloatBarrier

\section{Prompts For \rollingopt{} and \rollinggpt{}}\label{sec:appendix_dumb_baseline_prompts}

\rollingopt{} and \rollinggpt{} use the same prompts. For the very first 256-token passage of generation, an example prompt is shown in Table \ref{tab:rolling_prompt_beginning}. Subsequent prompts follow the pattern in Table \ref{tab:rolling_prompt_later}. 

\begin{table*}[!htbp]
\small
\begin{tabularx}{\linewidth}{X}
\toprule
\texttt{Premise: After the loss of her father, Shannon is determined to follow in his footsteps and become a successful journalist. However, when she lands her first major assignment, she quickly discovers that the ugly reality of life in the city is far different from the dream she imagined. With the help of her new friend, a street-wise teenager, Shannon comes to understand the harsh realities of life in the inner city and learns that sometimes the truth is much more than just a story.}\\
\\
\texttt{Current Story Outline: Shannon's father, Mike, dies unexpectedly, leaving her determined to follow in his footsteps and become a successful journalist.}\\
\\
\texttt{Write a story according to this premise, starting with the current outline.}\\
\\
\texttt{-----------------------------}\\
\\
\texttt{Chapter 1}\\
\bottomrule
\caption{Example prompt for first passage of generation for \rollingopt{} and \rollinggpt{}.}
\label{tab:rolling_prompt_beginning}
\end{tabularx}
\end{table*}

\begin{table*}[!htbp]
\small
\begin{tabularx}{\linewidth}{X}
\toprule
\texttt{Premise: After the loss of her father, Shannon is determined to follow in his footsteps and become a successful journalist. However, when she lands her first major assignment, she quickly discovers that the ugly reality of life in the city is far different from the dream she imagined. With the help of her new friend, a street-wise teenager, Shannon comes to understand the harsh realities of life in the inner city and learns that sometimes the truth is much more than just a story.}\\
\\
\texttt{Current Story Outline: With the help of her new friend, Gary, Shannon comes to understand the harsh realities of life in the inner city and learns that sometimes the truth is much more than just a story.}\\
\\
\texttt{Write a story according to this premise, continuing from the current outline.}\\
\\
\texttt{-----------------------------}\\
\\
\texttt{that I think he may not have disappeared of his own accord." She wasn't sure if that was how it would sound or not but it was what came naturally at the moment so Shannon decided not to worry about it!}\\
\\
\texttt{"I see," the woman said slowly after a long pause. Clearly no one had called in two years telling them they thought their loved one didn't just up and disappear...that must have been something they weren't used to hearing. "I'm sorry to hear that you think your father may have been a victim of foul play."}\\
\\
\texttt{"Thank you for understanding; however, I do have a reason for believing this," Shannon explained, hoping that her voice didn't sound too shaky. She was sure no one had called in two years to say they thought their loved one hadn't disappeared at all! "For one thing, he was working on an important story about the inner city and the police force."}\\
\\
\texttt{"Really?" the woman asked with a confused look in her voice. Shannon nodded, unable to speak because she knew no one would believe her if she tried to tell them that someone had called just like this two years ago! But she was going to tell this woman everything and then see if they would help her figure out what happened...or at least try to find Mike's killer before she figured it out herself! "I'm sorry but it sounds like you think your father's disappearance may be related to his work...and I'm sorry but I can't help you there," she told Shannon apologetically. "If he disappeared under suspicious circumstances then you can report it to the department and we'll investigate again but we only investigate if foul play is suspected," she continued. "Otherwise the case is considered closed."}\\
\\
\texttt{"I don't understand," Shannon explained slowly. "Did you not hear me earlier? I called to report something suspicious."}\\
\\
\texttt{"Oh this isn't about what happened to your father," the woman said, shaking her head as if Shannon were being silly. "I can tell you that from what I've read in the files, there was nothing suspicious about his disappearance and no evidence of foul play...it wasn't a murder or anything like that."}\\
\\
\texttt{"I don't understand," Shannon repeated slowly. "I'm not the one who called...this is exactly why I wanted to call!" She pressed her lips together again, trying to figure out how she had messed up; she was sure no one had told her Michael's case had been officially closed! Sure, he hadn't been reported missing because it was believed he had taken off on his own...but that didn't mean he wasn't a victim! It just meant he didn't have any friends or family who would care enough to report him missing in the first place! And there hadn't been any way for anyone else to find out what happened until Shannon started looking for answers on her own two years later!}\\
\\
\texttt{"Look, all I can do is tell}\\
\bottomrule
\caption{Example prompt for later passage of generation for \rollingopt{} and \rollinggpt{}.}
\label{tab:rolling_prompt_later}
\end{tabularx}
\end{table*}

\FloatBarrier

\section{Experiment Costs}

Over the course of this work, we estimate that we spent \$3000-\$4000 on GPT3 API costs and roughly \$4000 on Surge AI annotation costs, including both development/preliminary experiments and final experiment costs. We estimate that we used about 2000 GPU hours on 80GB NVIDIA A100 GPUs for all experiments, in addition to a smaller number of GPU hours on smaller GPUs during earlier experiments. 

\oursimpl{} takes two to three times longer to generate stories compared to \rethreeimpl{} (which is in turn slower than the GPT3-175B-based version from \citet{yang2022re3}; we assume the public GPT3-175B API is heavily optimized for performance). The slowdown seems to be largely due to our \fudge{} implementation which requires token-level caching and restarting in OPT-175B served by Alpa, which we did not heavily optimize. In principle it should be possible to make \oursimpl{} only marginally slower than \rethreeimpl{} or the original implementation from \citet{yang2022re3}.

\FloatBarrier

\section{Average Story Lengths}\label{sec:appendix_story_length}

We show the average lengths of stories for different methods. The lengths of stories from our main comparisons in Table \ref{tab:main_results} are shown in Table \ref{tab:length_ours_rethree}, while the ablations from Table \ref{tab:ablations} are shown in Table \ref{tab:length_ablations}. Besides \oursimplnofudge{} in the ablations which has somewhat longer average length (because the early stopping heuristic triggers less frequently, due to weaker relevance), different methods have fairly similar average lengths. 

\begin{table}[htbp]
\small
\centering
\begin{tabular}{@{}lccc@{}}
\toprule
\textbf{Method}    & \textbf{Average Story Word Count}\\
\midrule
\rethreeimpl{} & 3810 \\
\rollingopt{} & 3437 \\
\rollinggpt{} & 3831 \\
\oursimpl{}           &   3875      \\
% \oursimplbeamtwo{} & \\
\bottomrule
\end{tabular}
\caption{\small Average word counts of 20 stories per method in our main comparisons in Table \ref{tab:main_results}.}% \todo{describe results, including for beam 2; significance/bolding}}
\label{tab:length_ours_rethree}
\vspace{-0.5em}
\end{table}

\begin{table}[htbp]
\small
\centering
\begin{tabular}{@{}lccc@{}}
\toprule
\textbf{Method}    & \textbf{Average Story Word Count}\\
\midrule
\oursimplonelevel{} & 3547 \\
\oursimplnofudge{}         &   4190      \\
\oursimpl{}           &   3527       \\
% \oursimplbeamtwo{} & \\
\bottomrule
\end{tabular}
\caption{\small Average word counts of 10 stories per method in our ablations in Table \ref{tab:ablations}.}% \todo{describe results, including for beam 2; significance/bolding}}
\label{tab:length_ablations}
\vspace{-0.5em}
\end{table}

\FloatBarrier

\section{Annotator Agreement}\label{sec:appendix_annotator_agreement}

In Table \ref{tab:annotator_agreement}, we show Fleiss' kappa for annotation agreement for our main comparisons in Table \ref{tab:main_results}. Although the annotator agreement remains fairly low due to the subjective nature of the metrics, our agreement is clearly better compared to \citet{yang2022re3}, who observed Fleiss' kappa values largely below 0.1 or even negative in some cases. 

\begin{table*}[htbp]
\small
\centering
\begin{tabular}{@{}lccc@{}}
\toprule
\textbf{Comparison}    & \textbf{Coherent Agreement} & \textbf{Relevant Agreement} & \textbf{Interesting Agreement}\\
\midrule
\rethreeimpl{} vs \oursimpl{} & 0.19 & 0.24 & 0.15 \\
\rollingopt{} vs \oursimpl{}       &   0.22 & 0.33 & 0.35     \\
\rollinggpt{} vs \oursimpl{}   & 0.21      &   0.42 & 0.20      \\
\bottomrule
\end{tabular}
\caption{\small Fleiss' kappa for different metrics from our experiments in Table \ref{tab:main_results} comparing \oursimpl{} to \rethreeimpl{}, \rollingopt{}, and \rollinggpt{}.}
\label{tab:annotator_agreement}
\vspace{-0.5em}
\end{table*}

% \begin{table}[htbp]
% \small
% \centering
% \begin{tabular}{@{}lccc@{}}
% \toprule
% \textbf{Method}    & \textbf{Average Story Length}\\
% \midrule
% \oursimplonelevel{} & 3547 \\
% \oursimplnofudge{}         &   4190      \\
% \oursimpl{}           &   3527       \\
% % \oursimplbeamtwo{} & \\
% \bottomrule
% \end{tabular}
% \caption{\small Average lengths of 10 stories per method in our ablations.}% \todo{describe results, including for beam 2; significance/bolding}}
% \label{tab:length_ablations}
% \vspace{-0.5em}
% \end{table}

\FloatBarrier

\newpage

\section{Optional Free-Form Comments From Human-Interactive Experiment}\label{sec:appendix_interactive_comments}

In Table \ref{tab:optional_comments} we show all of the optional comments written by annotators following our human-interactive experiment (Section \ref{sec:experiments_control}), omitting empty comments. \rethreeimpl{} is System A and \oursimpl{} is System B. Perceptions of overall story quality vary, but annotators clearly prefer \oursimpl{} for controllability. The complete plans and stories from this experiment are available at \url{https://github.com/yangkevin2/doc-story-generation}.

\onecolumn

\begin{small}
\begin{longtable}{p{0.9\textwidth}}
\toprule
\texttt{The AI does a quite commendable job with my original three-sentence premise. There are mistakes here and there that a (good) human writer would not make - multiple paragraphs beginning the exact same way was the most glaring in one section. But I'm pleased. Hope there will be more experiments like this - thank you.}\\
\midrule
\texttt{Both stories made me want to read them. But the style of the output of System B was a lot closer to what I had in mind originally.}\\
\midrule
\texttt{I mean, the result is FAR from what I was looking for. I could imagine a system having a template to fill out for various platpoints, characters, timelines, etc. I like the idea of having some base story ideas and scenes being generated, but very little of the outline seemed to be followed or integrated into the story. It was a real hodgepodge. I understand you might need to go through some iterations but I would rather have less writing that is more on topic and outline than something that confused the people, city, location, base material in general so much. The story only hints at fragments of the story I envisioned. A fun exercise, albeit also frustrating. I did prefer the results os System B in all cases except the first, where it mixed up my imagination country Liberius with Liberia.}\\
\midrule
\texttt{Both of my stories are pretty nonsensical and aren't cohesive. While I feel like System B kept things a bit closer to the outline described, I think System A contradicted itself a little less than B and potentially told a better story.}\\
\midrule
\texttt{Quick takeaways:}\\
\texttt{1. The ability to align time is a mess. For example, in story one the children have just moved out, sooner than expected. Travel down through the story and "Nadine was unsure if her daughter would even want to see her, or talk to her again after allthese years.". Very confusing. This happened throughout both versions, in various forms and in abundance.}\\
\texttt{2. Characters descriptions in the story did not match those presented in the outline. This was a major issue regarding storyline and clarity in both versions of the story. Ex. Lillian is her best friend, Nadine just finished publishing her book, yet in version 2 of the story she is introduced for the first time in Nadine's life.}\\
\midrule
\texttt{System A seemed to go more astray and get involved in plot points not directly related to the overall plot.}\\
\midrule
\texttt{The difference between the two systems was pretty big. System A didn't seem to stick to important plot points at all (having a deceased character come back without explanation, a "missing father" arch, made up teachers, wrong location, etc.) While system B had a very blunt approach to the story somewhere between the border of comical/offensive (which was not the point of the story). That said, B did stick to the plot points in there entirety and made a lot more sense than A.}\\
\midrule
\texttt{To start, having an AI write a story from the prompts we gave is impressive to me, and both of them came out as cohesive stories.  But, neither of them really hit exactly what I was looking for with my prompts and they had a few flaws.  System B seemed to get stuck in a “loop” sometimes with the dialog, like when they were talking about who was faster. It got repetitive really quickly and took me out of the story. It also focused a lot on an iPod for some reason, which also pulled me out of it. The writing and story telling in System A was more enjoyable and easier to read, but the storyline of System B seemed more in line with what I was thinking, so it was hard to chose between the 2 of them.  If I were using this system, I would be very happy with either result, as they are both great rough drafts of the story.}\\
\midrule
\texttt{I didn't feel like with either system that I had very much control, and it seemed like the final passages derived didn't match the outlines very well and were not particularly coherent.  There were a lot of repeated moments and portions that literally were impossible or simply didn't make any sense in the context of the story at all.}\\
\midrule
\texttt{I think the more detailed outline in System B really helped shape the story into more of what I was envisioning. Both passages had some inconsistencies where the quality would seem lacking, but passage A was worse in that way. For example, a major one in passage A is that it describes how Daniel and his wife have no children, but the character listing in the outline shows them having two daughters. Passage A, however, did have a more exciting story overall with more details and dialogue. In a way, it read as a more traditional fictional story, but it was inconsistent with the outline. My preference would still be for System B for the level of detail I was able to control and how it stayed truer to the outline.}\\
\midrule
\texttt{I don’t know what system a was trained on, but it definitely had issues. Beyond knowing what content is appropriate or relevant it had a lot of nonsequiturs and contradictory facts about the characters. B was much much higher quality.}\\
\midrule
\texttt{it seems like the more detail that can be provided, the better the story would be—without the sublevels of detail in System A, my story seemed a lot less cohesive/sensible. And when writing a story I definitely want to control as much detail as possible/not make it so general that I'm leaving a big part of the plot up to chance, so I liked System B because of that.}\\
\midrule
\texttt{It was interesting to me that System A generated more lengthy passages despite having a less complex outline to go by...System A's story was maybe more suspenseful/interesting but sometimes didn't make sense and ignored my outline, so System B definitely fit my vision better in almost every situation. That being said, had I just been evaluating these two stories on their sheer entertainment value without realizing what my outline and intentions were, I may have found it to be more entertaining (though it does seem slightly more all over the place than the more focused story from System B).}\\
\bottomrule
\caption{Optional comments written by annotators following our human-interactive experiment (Section \ref{sec:experiments_control}). While judgments of overall story quality are mixed, with some being disappointed and others pleased, they overwhelmingly describe \oursimpl{} (System B) as more faithful to the plot and their original authorial intent.}
\label{tab:optional_comments}
\end{longtable}
\end{small}

\twocolumn

% \FloatBarrier

% \newpage

\section{Annotation Task Details}\label{sec:appendix_annotation_templates}

Surge AI describes their platform's worker population as ``highly skilled and educated native speakers''; we did not apply further filters. Our data collection was determined exempt from an ethics review board. 

Below we show annotation templates shown to Surge AI workers for our various experiments.

\subsection{Main Experiment Annotation Template}\label{sec:appendix_annotation_templates_main}

\begin{figure*}[htp]

\subfloat{
  \includegraphics[clip,width=\textwidth]{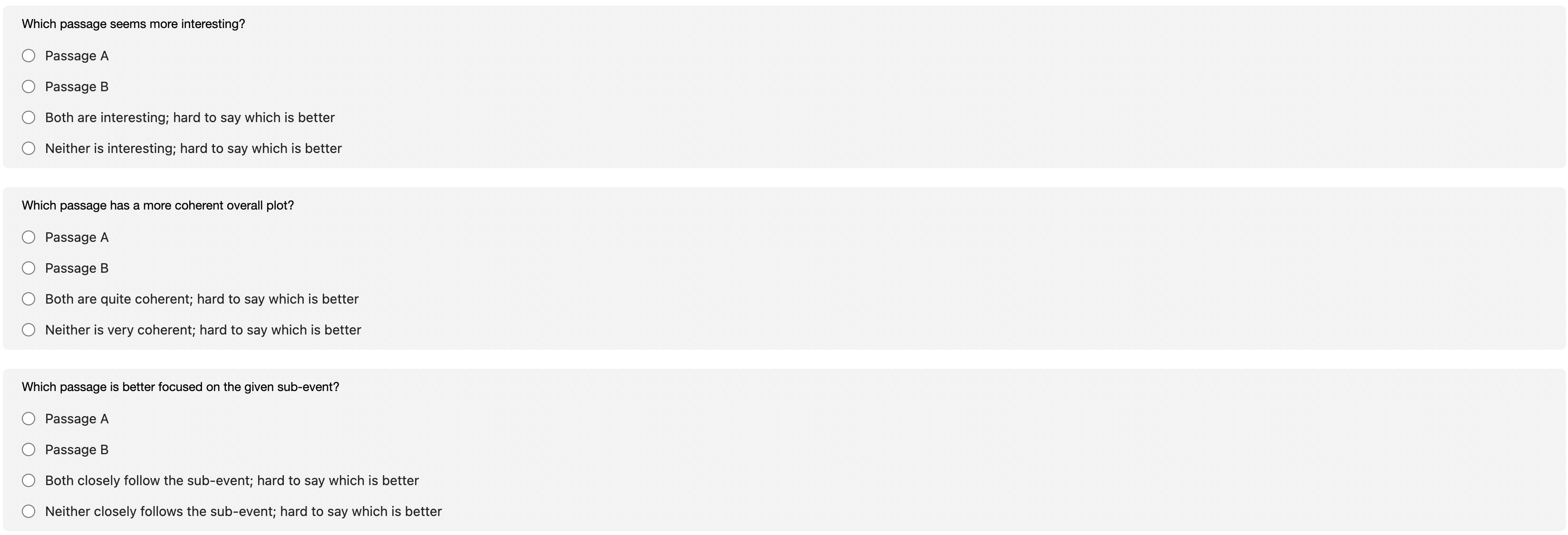}
}

\subfloat{
  \includegraphics[clip,width=\textwidth]{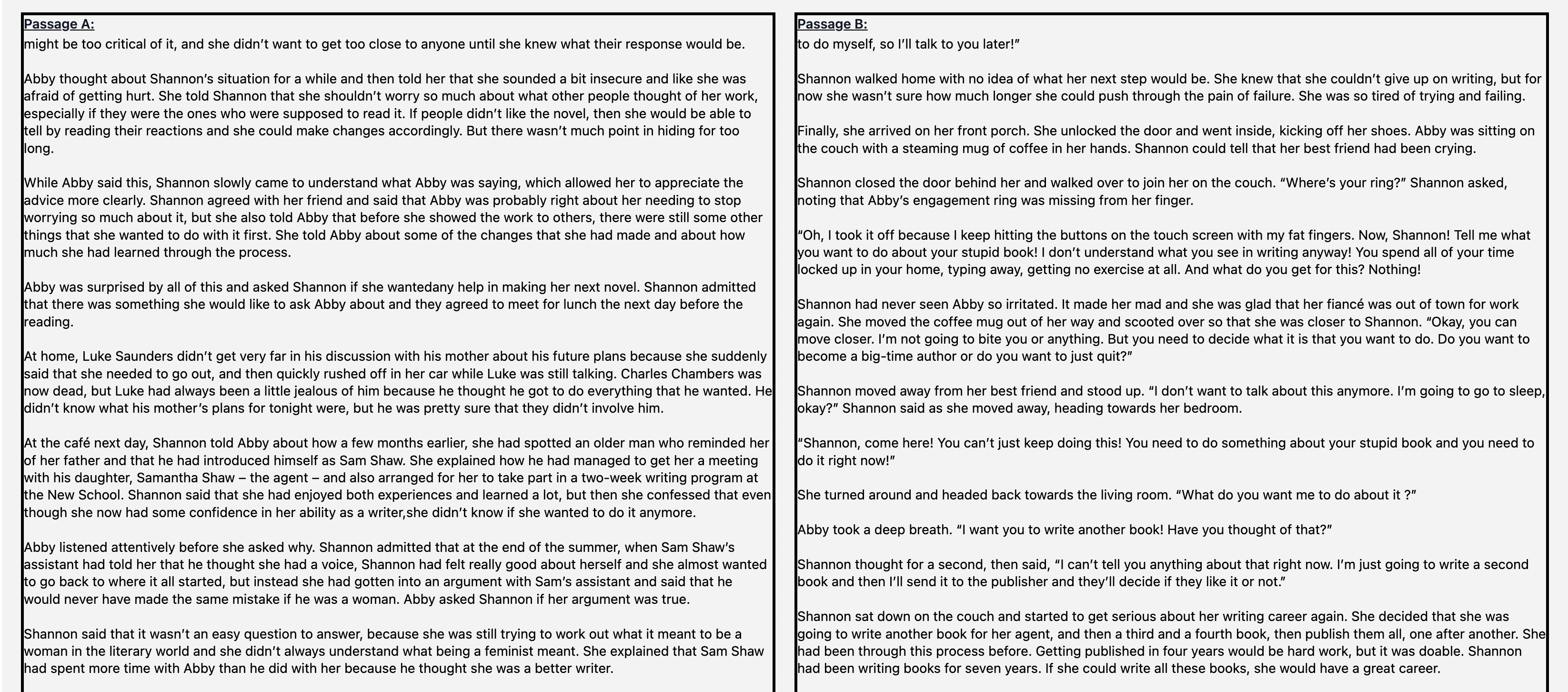}
}

\subfloat{
  \includegraphics[clip,width=\textwidth]{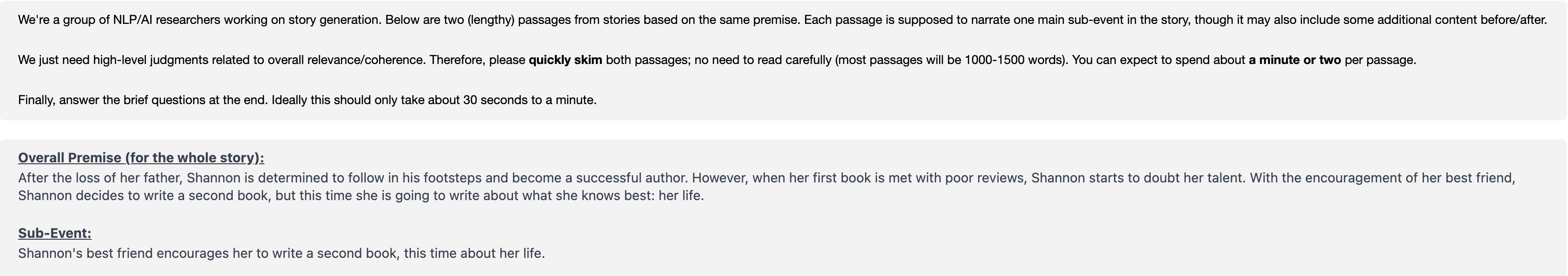}
}

\caption{Surge AI annotation example for main comparisons in Table \ref{tab:main_results}. The stories are truncated here for brevity.}
\label{fig:annotation_template_main}
\end{figure*}

Figure \ref{fig:annotation_template_main} shows an example of our annotation template for our main comparisons from Table \ref{tab:main_results}. We paid workers \$1.20 per annotation, aiming to pay roughly \$20 per hour based on our time estimates of average task length.

% \FloatBarrier

\subsection{Human Interactive Experiment Annotation Template}\label{sec:appendix_annotation_templates_human}

\begin{figure*}[htp]
\includegraphics[clip,width=\textwidth]{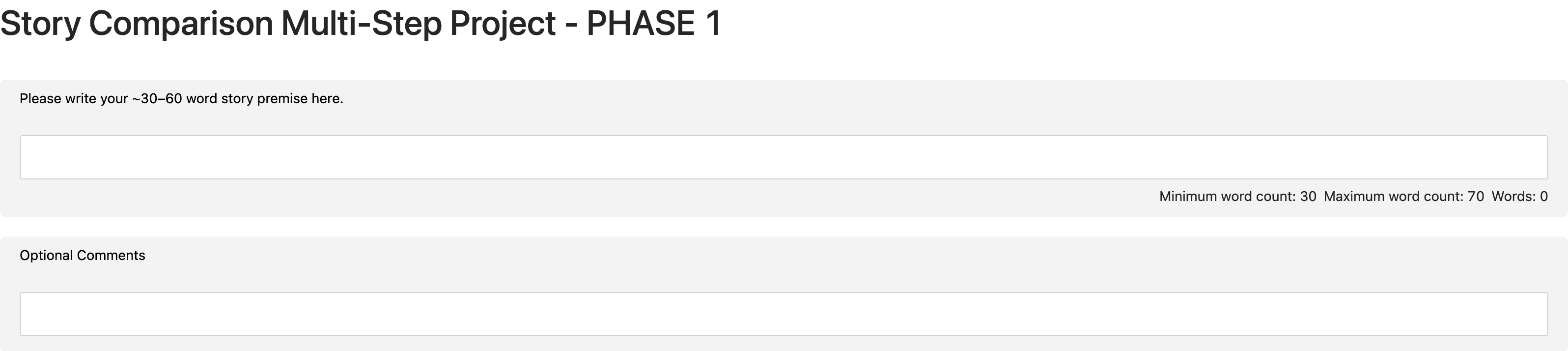}
\caption{Surge AI annotation example for human interactive experiment, Phase 1.}
\label{fig:annotation_template_control_phase1}
\end{figure*}

\begin{figure*}[htp]
\subfloat{
  \includegraphics[clip,width=\textwidth]{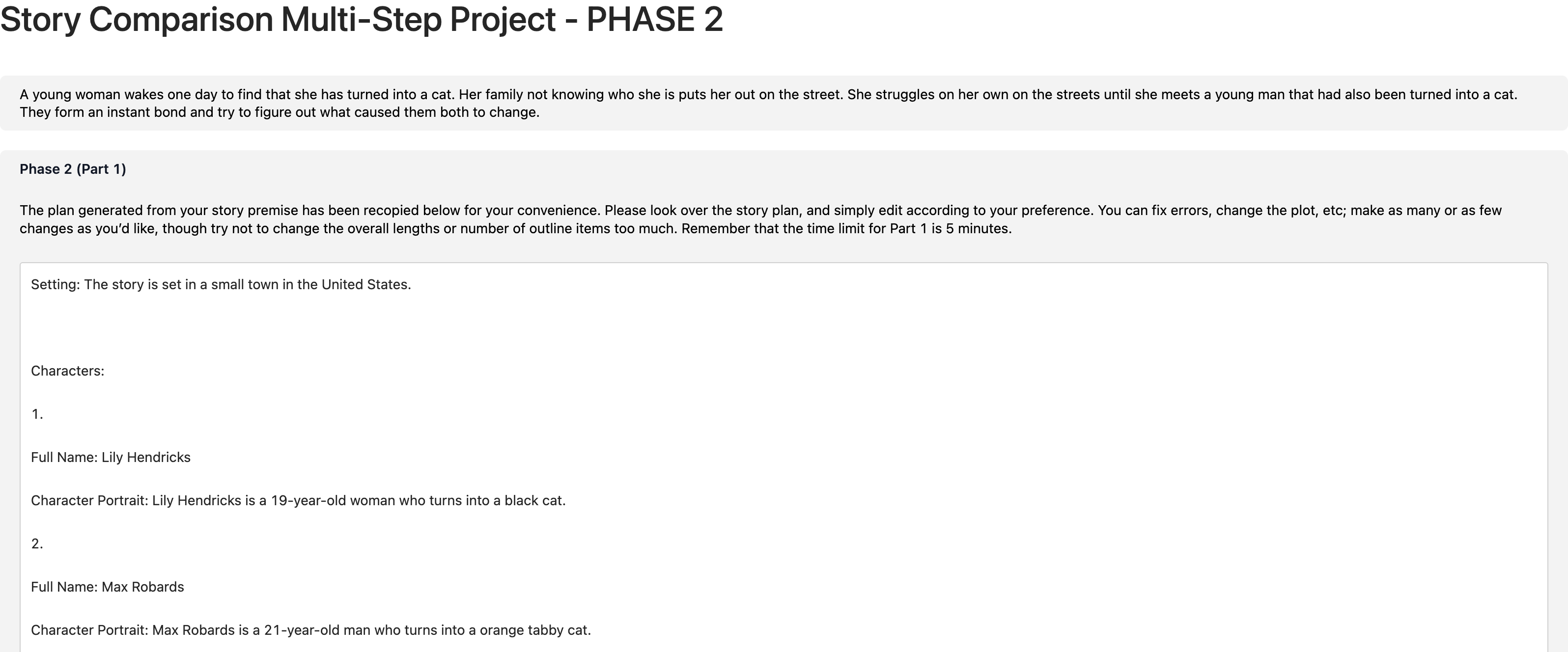}
}

\subfloat{
  \includegraphics[clip,width=\textwidth]{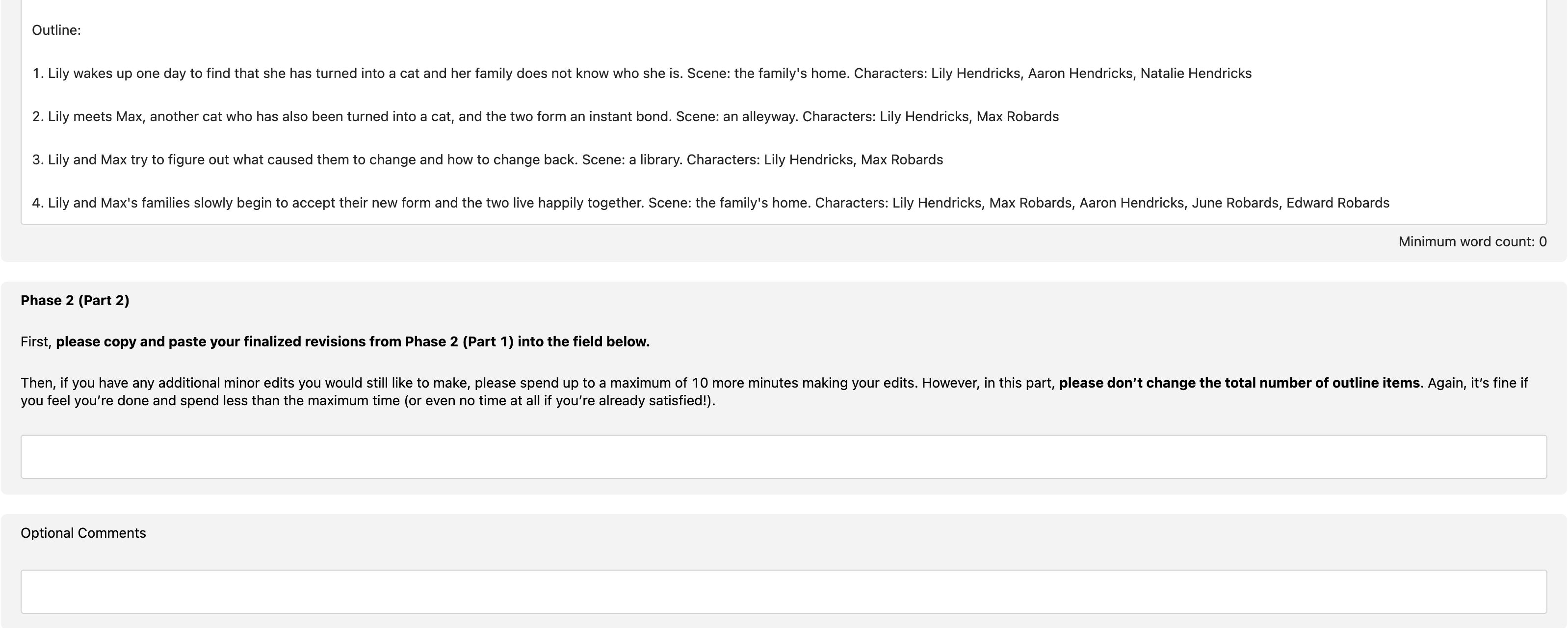}
}
\caption{Surge AI annotation example for human interactive experiment, Phase 2. Plans are abridged. The output of Phase 2 Part 2 is the final plan for \rethreeimpl{} (System A).}
\label{fig:annotation_template_control_phase2}
\end{figure*}

\begin{figure*}[htp]
\subfloat{
  \includegraphics[clip,width=\textwidth]{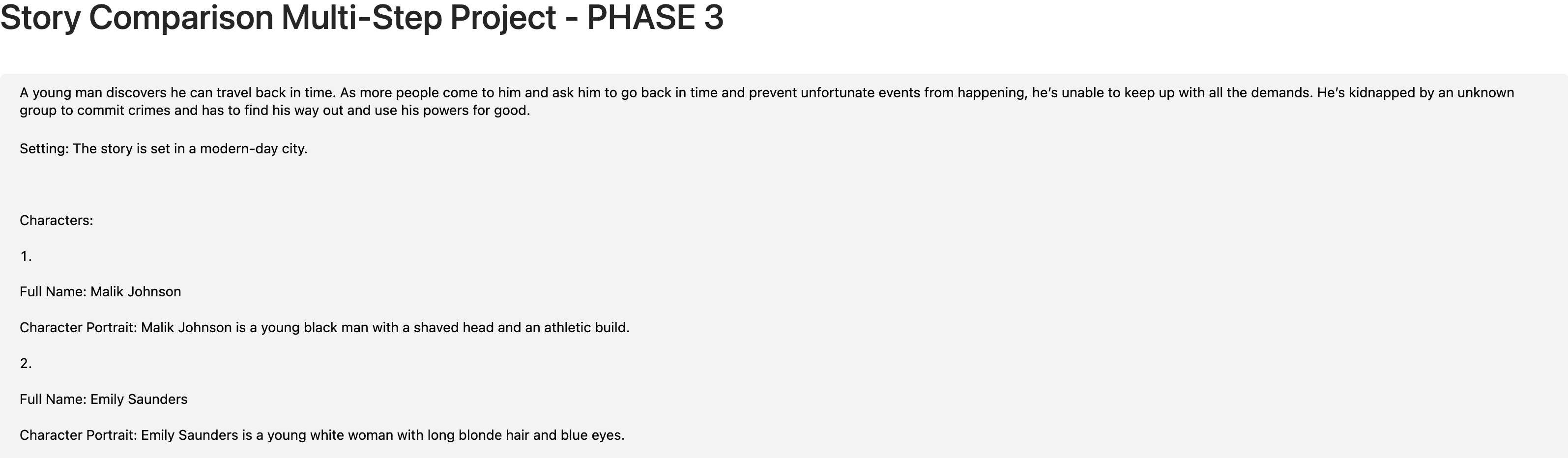}
}

\subfloat{
  \includegraphics[clip,width=\textwidth]{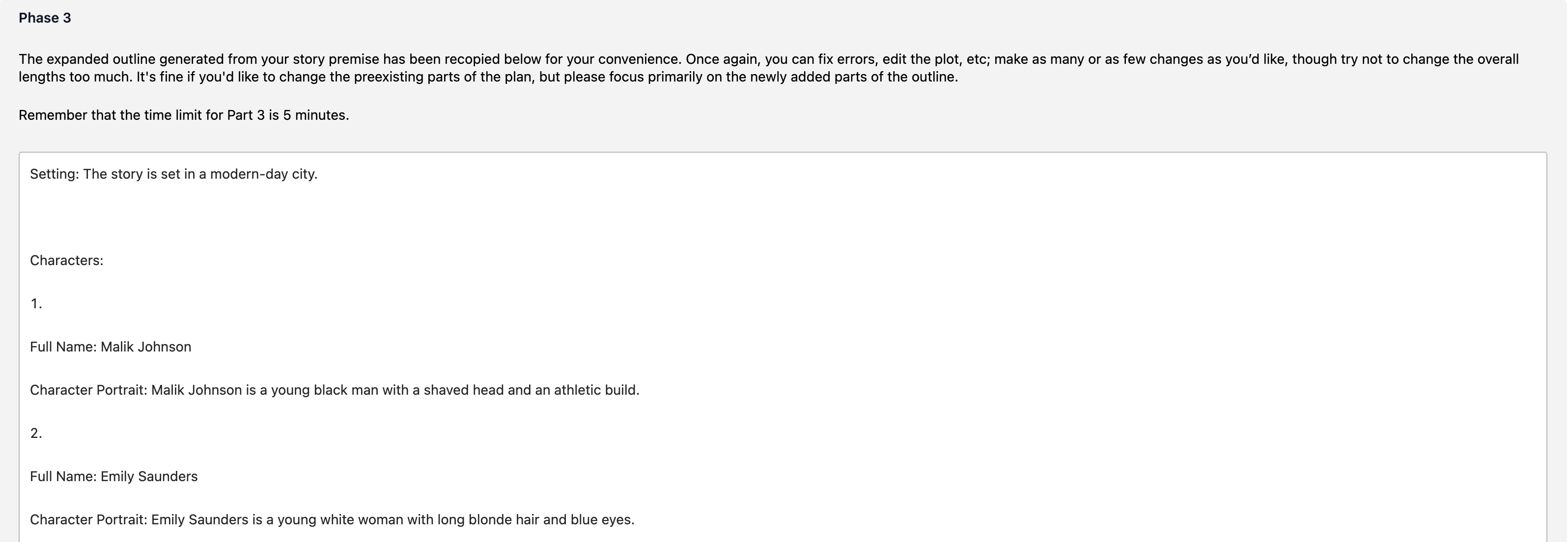}
}

\subfloat{
  \includegraphics[clip,width=\textwidth]{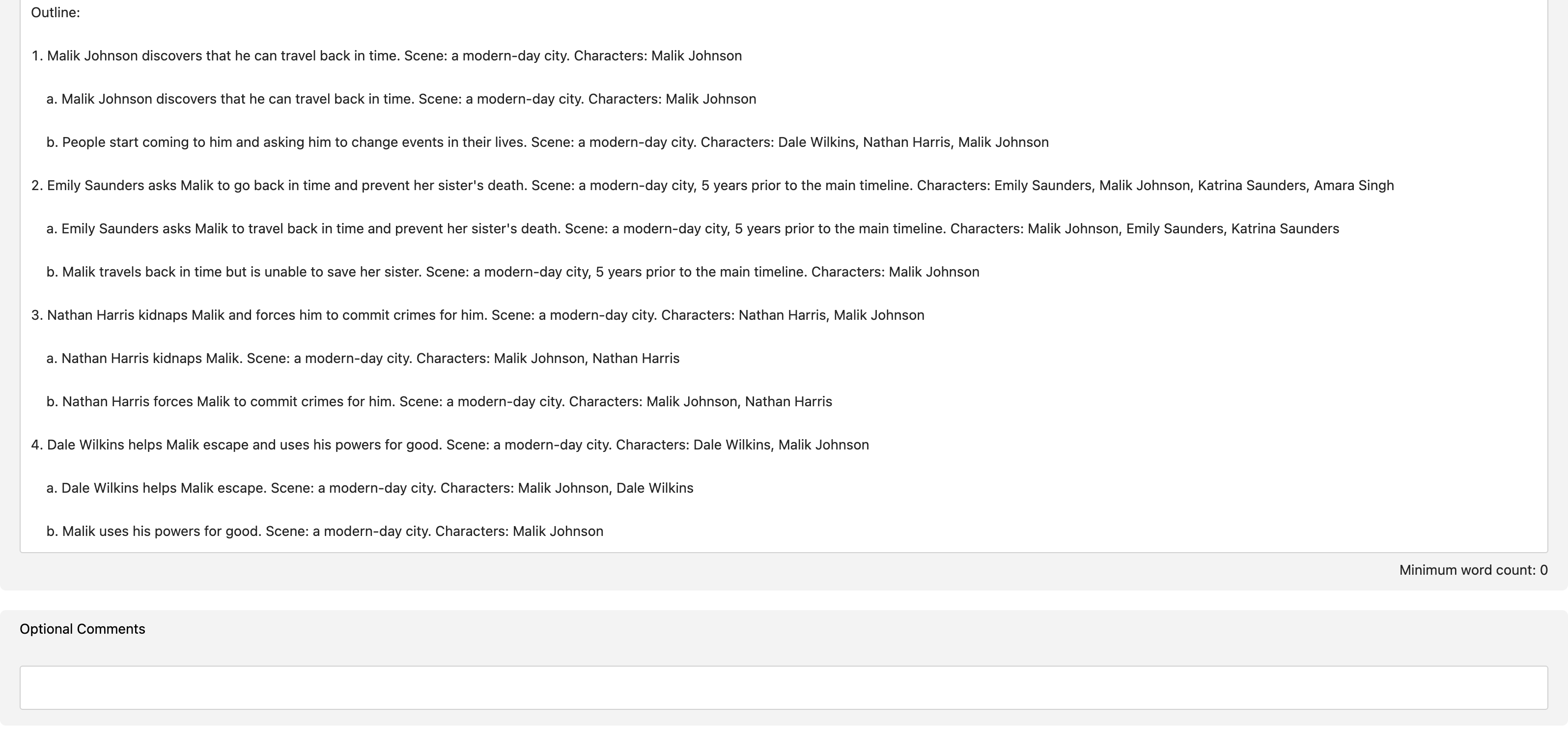}
}
\caption{Surge AI annotation example for human interactive experiment, Phase 3. Plans are abridged.}
\label{fig:annotation_template_control_phase3}
\end{figure*}

\begin{figure*}[htp]
\subfloat{
  \includegraphics[clip,width=\textwidth]{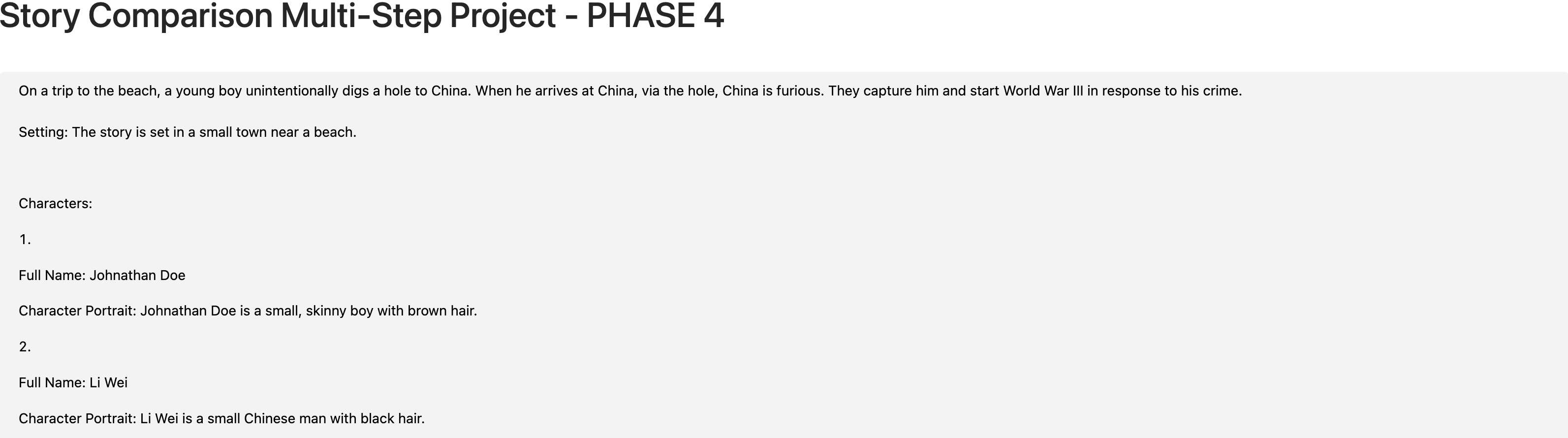}
}

\subfloat{
  \includegraphics[clip,width=\textwidth]{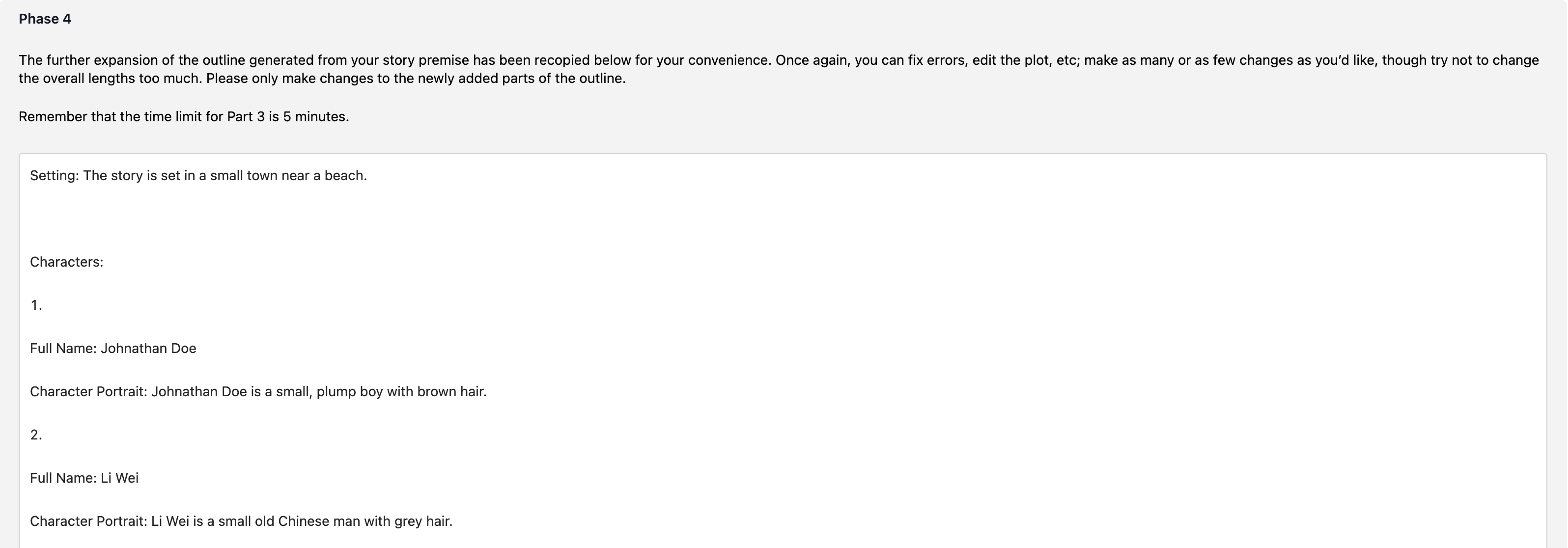}
}

\subfloat{
  \includegraphics[clip,width=\textwidth]{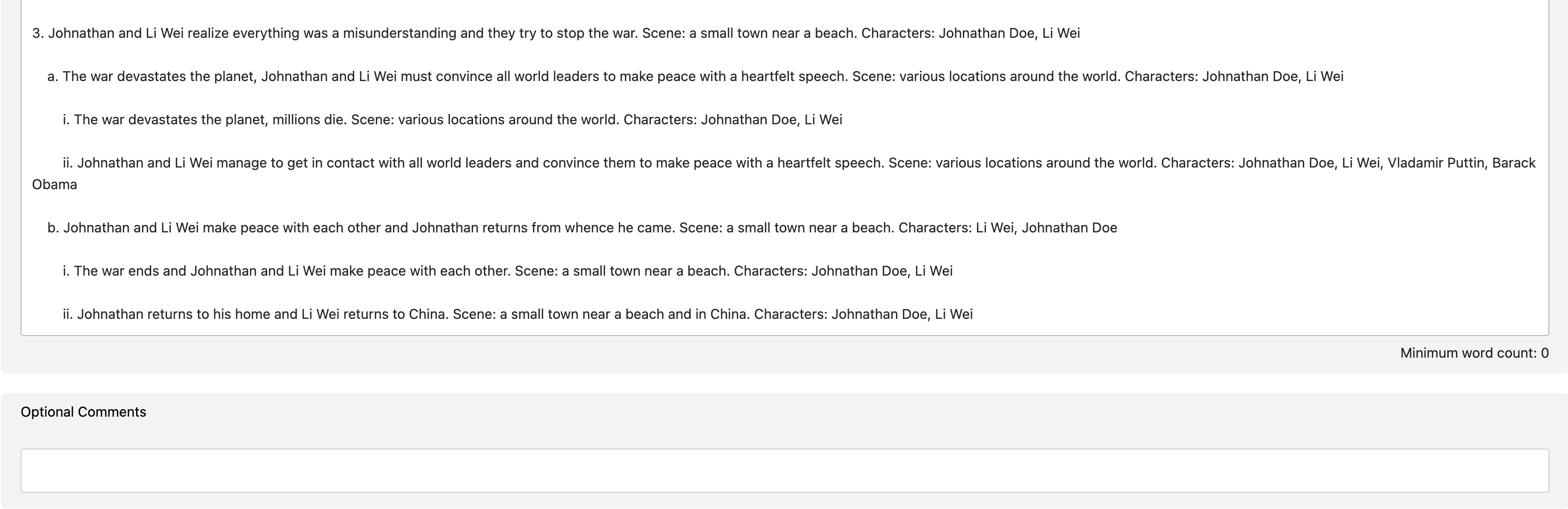}
}
\caption{Surge AI annotation example for human interactive experiment, Phase 4. Plans are abridged. The output of Phase 4 is the final plan used for \oursimpl{} (System B).}
\label{fig:annotation_template_control_phase4}
\end{figure*}

\begin{figure*}[htp]
\subfloat{
  \includegraphics[clip,width=\textwidth]{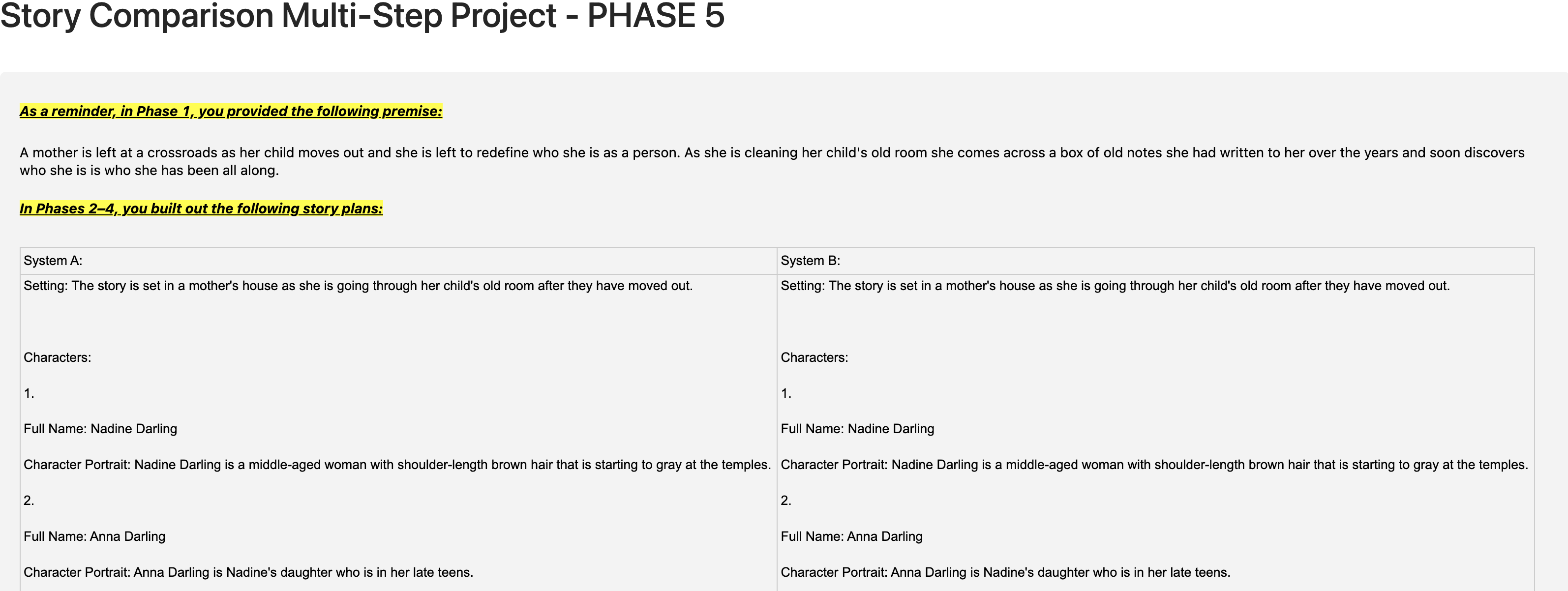}
}

\subfloat{
  \includegraphics[clip,width=\textwidth]{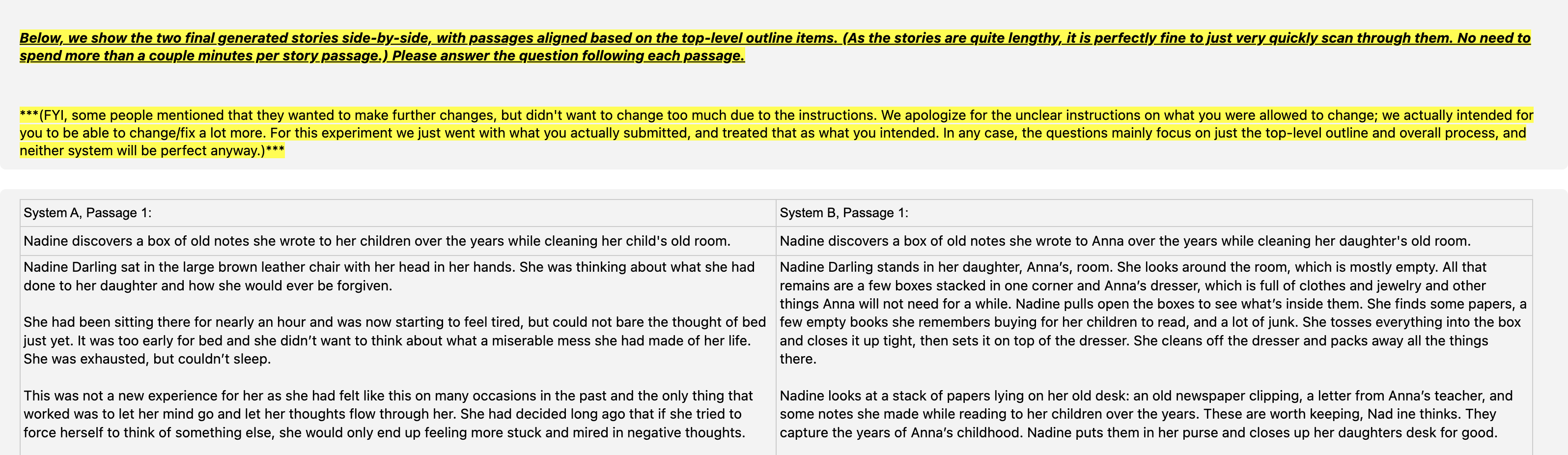}
}

\subfloat{
  \includegraphics[clip,width=\textwidth]{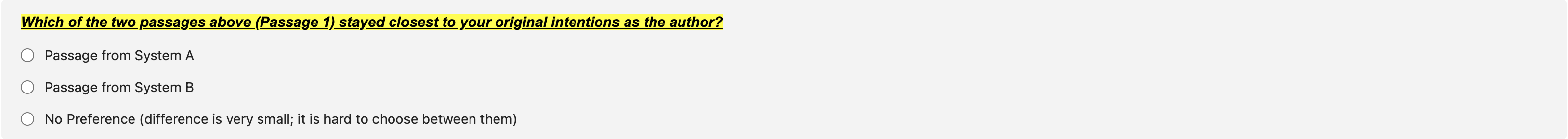}
}

\subfloat{
  \includegraphics[clip,width=\textwidth]{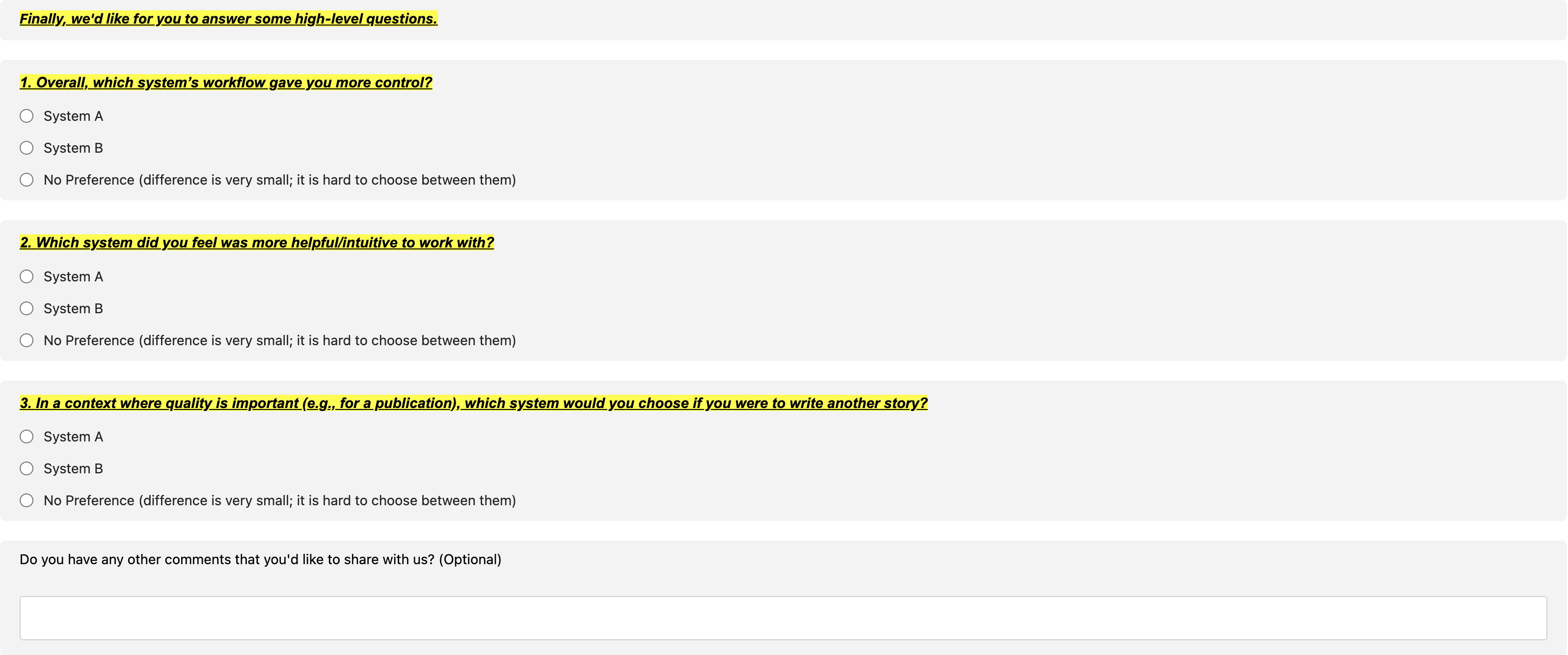}
}
\caption{Surge AI annotation example for human interactive experiment, Phase 5. Plans and story passages are abridged. The question about original intentions as author (Intent metric) is asked once for each pair of top-level outline items and corresponding passages, although only one instance is shown here. The remaining questions are asked only once at the bottom.}
\label{fig:annotation_template_control_phase5}
\end{figure*}

We ran the human interactive experiment through Surge AI's Managed Service, so the task was constructed by Surge AI according to our instructions. 
%, based on our template drafted at \url{https://docs.google.com/document/d/161YzBcwWahp25WOSSMl-0UET_l7mTfPUhKFpwPheaO0/edit?usp=sharing}. 
The task consisted of 5 phases for which we had the same 20 annotators return each time. System A is \rethreeimpl{} while System B is \oursimpl{}. The templates for the 5 phases are shown in Figures \ref{fig:annotation_template_control_phase1}, \ref{fig:annotation_template_control_phase2}, \ref{fig:annotation_template_control_phase3}, \ref{fig:annotation_template_control_phase4}, and \ref{fig:annotation_template_control_phase5} respectively. We paid Surge AI \$1000 for this experiment, which includes the payment for the 20 workers, who we expected to spend 30-45 minutes in total across the five phases of the experiment.

% \FloatBarrier

\subsection{Detailed Outline Relevance Experiment Annotation Template}\label{sec:appendix_annotation_templates_detailed_relevance}

\begin{figure*}[!htbp]

\subfloat{
  \includegraphics[clip,width=\textwidth]{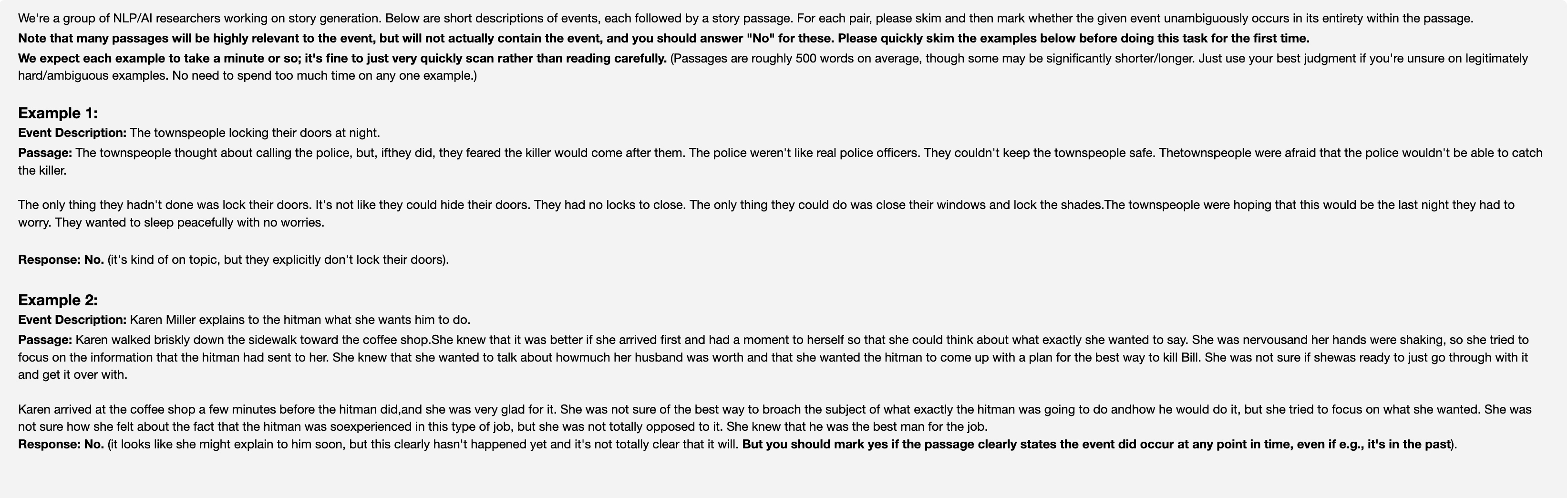}
}

\subfloat{
  \includegraphics[clip,width=\textwidth]{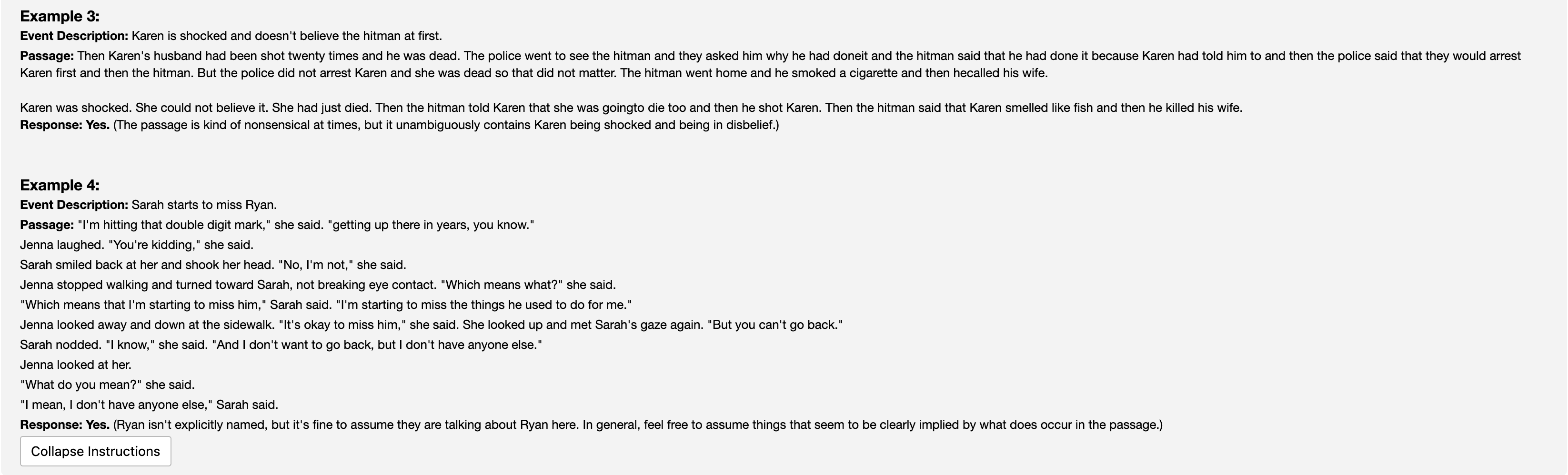}
}

\subfloat{
  \includegraphics[clip,width=\textwidth]{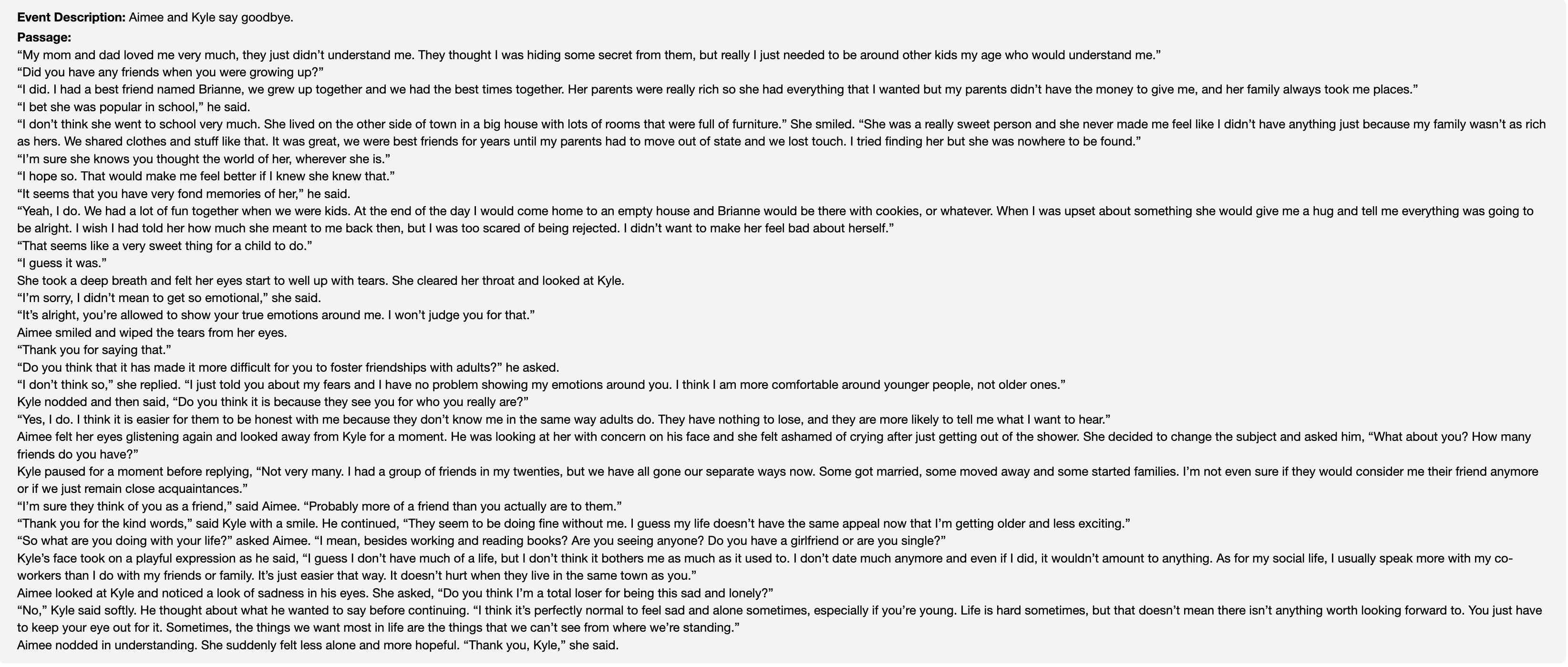}
}

\subfloat{
  \includegraphics[clip,width=\textwidth]{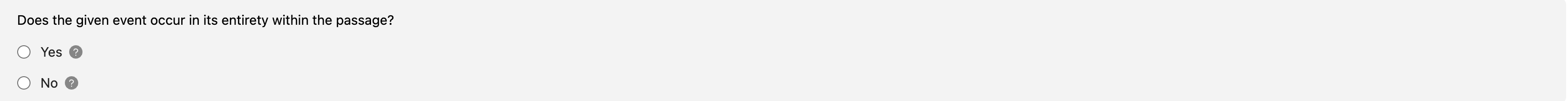}
}
\caption{Surge AI annotation example for detailed relevance from our experiments in Table \ref{tab:detailed_relevance}.}
\label{fig:annotation_template_detailed_relevance}
\end{figure*}

Figure \ref{fig:annotation_template_detailed_relevance} shows an example of our annotation template for measuring whether a given passage contains the event described in a low-level outline item, corresponding to the results in Table \ref{tab:detailed_relevance}. We paid workers \$0.50 per annotation, aiming to pay roughly \$20 per hour based on our time estimates of average task length.

\FloatBarrier

\newpage

\section{Example Outlines Without Filtering or Reranking}\label{sec:appendix_trivial3level}

In Tables \ref{tab:outline_bad1} and \ref{tab:outline_bad2} we show the first two example outlines that we generated after turning off our filtering and reranking infrastructure when selecting outline items for our detailed outline. They are both clearly problematic, featuring repetitive text and many events that happen seemingly out of order. We observed similar issues in further generations under this no-filtering no-reranking setting, demonstrating the necessity of our filtering and reranking schemes. 

\begin{table*}[!htbp]
\small
\begin{tabularx}{\linewidth}{X}
\toprule
\textbf{Low-Quality Detailed Outline (Filtering/Reranking Ablation)}\\
\midrule
\texttt{1. Sarah Jensen discovers that she comes from a long line of witches and must come to terms with her new reality.}\\
\texttt{\quad \quad a. Sarah discovers her true identity when her powers begin to manifest.}\\
\texttt{\quad \quad \quad \quad i. Sarah's powers begin to manifest and she starts to experience strange things happening around her.}\\
\texttt{\quad \quad \quad \quad ii. Sarah's parents tell her about her heritage and how she comes from a long line of witches.}\\
\texttt{\quad \quad b. Jenna Howard, Sarah's best friend, helps her to accept her new identity.}\\
\texttt{\quad \quad \quad \quad i. Jenna helps Sarah to accept her new identity and introduces her to the world of witchcraft.}\\
\texttt{\quad \quad \quad \quad ii. Jenna helps Sarah to come out as a witch to her boyfriend, Aaron.}\\
\texttt{2. Sarah learns how to use her powers with the help of Silas Baxter, a witch who takes her under his wing.}\\
\texttt{\quad \quad a. Sarah learns how to control her powers and use them for good.}\\
\texttt{\quad \quad \quad \quad i. Sarah learns how to cast spells and use her powers for good.}\\
\texttt{\quad \quad \quad \quad ii. Sarah must stop a evil witch from taking over her town.}\\
\texttt{\quad \quad b. Sarah must save her father from a curse that has been placed on him.}\\
\texttt{\quad \quad \quad \quad i. Sarah's father is cursed and turns into a frog.}\\
\texttt{\quad \quad \quad \quad ii. Sarah must find a way to break the curse and save her father.}\\
\texttt{3. Sarah must use her powers to save her town from a evil witch who wants to destroy it.}\\
\texttt{\quad \quad a. Sarah confronts the evil witch and defeats her.}\\
\texttt{\quad \quad \quad \quad i. Sarah discovers her true identity when her powers begin to manifest.}\\
\texttt{\quad \quad \quad \quad ii. Sarah learns how to control her powers and use them for good.}\\
\texttt{\quad \quad b. Sarah learns that her powers come with a great responsibility and must use them wisely.}\\
\texttt{\quad \quad \quad \quad i. Sarah learns how to use her powers.}\\
\texttt{\quad \quad \quad \quad ii. Sarah saves her town from the evil witch.}\\
\bottomrule
\caption{First outline example with filtering and reranking for outline items turned off. Several events, especially in the second half of the outline, appear to be out of order or repetitive.}
\label{tab:outline_bad1}
\end{tabularx}
\end{table*}

\begin{table*}[!htbp]
\small
\begin{tabularx}{\linewidth}{X}
\toprule
\textbf{Low-Quality Detailed Outline (Filtering/Reranking Ablation)}\\
\midrule
\texttt{1. After losing her job, Jennifer Walters starts her own bakery with the help of her best friend Elise Miller.}\\
\texttt{\quad \quad a. Jennifer is fired from her job and decides to start a bakery with the help of her best friend Elise.}\\
\texttt{\quad \quad \quad \quad i. Jennifer Walters is fired from her job}\\
\texttt{\quad \quad \quad \quad ii. Elise Miller decides to quit her job to help Jennifer start the bakery.}\\
\texttt{\quad \quad b. The pair start by renovating an old building into a beautiful bakery and kitchen.}\\
\texttt{\quad \quad \quad \quad i. Jennifer and Elise renovate an old building into a beautiful bakery.}\\
\texttt{\quad \quad \quad \quad ii. The bakery quickly becomes a success, thanks to the delicious recipes of head chef Harry Miller and the outstanding customer service provided by Jennifer and her team.}\\
\texttt{2. The bakery quickly becomes a success, thanks to the delicious recipes of head chef Harry Miller and the outstanding customer service provided by Jennifer and her team.}\\
\texttt{\quad \quad a. Jennifer and Elise put all their energy into making the bakery a success.}\\
\texttt{\quad \quad \quad \quad i. Jennifer and Elise start by renovating an old building into a beautiful bakery and kitchen.}\\
\texttt{\quad \quad \quad \quad ii. The bakery quickly becomes popular, thanks to the delicious recipes of head chef Harry and the outstanding customer service provided by Jennifer and her team.}\\
\texttt{\quad \quad b. The bakery quickly becomes popular, thanks to the delicious recipes of head chef Harry and the outstanding customer service provided by Jennifer and her team.}\\
\texttt{\quad \quad \quad \quad i. Jennifer and Elise put all their energy into making the bakery a success.}\\
\texttt{\quad \quad \quad \quad ii. The bakery quickly becomes popular, thanks to the delicious recipes of head chef Harry and the outstanding customer service provided by Jennifer and her team.}\\
\texttt{3. As the business grows, Jennifer and her family face new challenges, but with the support of their community, they overcome them all.}\\
\texttt{\quad \quad a. Jennifer and her family face new challenges as the business grows.}\\
\texttt{\quad \quad \quad \quad i. Jennifer and her family face new challenges as the business grows.}\\
\texttt{\quad \quad \quad \quad ii. As the business grows, Jennifer and her family face new challenges, but with the support of their community,}\\
\texttt{\quad \quad b. with the support of their community, they overcome them all.}\\
\texttt{\quad \quad \quad \quad i. Jennifer overcomes her fear of failure and decides to open the bakery.}\\
\texttt{\quad \quad \quad \quad ii. Events that occur supportive community help the family to overcome their challenges.}\\
\bottomrule
\caption{Second outline example with filtering and reranking for outline items turned off. Similar to the previous example in Table \ref{tab:outline_bad1}, several events seem to be out of order or repetitive.}
\label{tab:outline_bad2}
\end{tabularx}
\end{table*}

\FloatBarrier

\onecolumn

\twocolumn

\section{Main Experiment Story Examples}\label{sec:appendix_story_examples}

Finally, we show the first five complete plan and story examples generated by \oursimpl{} from our main experiments, i.e., the examples are not cherry-picked. For the first two premises, we additionally show the stories generated by \rethreeimpl{} and \rollingopt{}. We briefly analyze each example individually in the captions. 

Overall, in addition to demonstrating strong quantitative performance as shown in the main text, \oursimpl{}'s plans and stories seem largely reasonable at a glance from the perspective of overarching plot. In contrast, \rethreeimpl{} and \rollingopt{} are generally much worse at following the high-level plan and maintaining overarching coherence; \rollingopt{}'s failures are particularly egregious.

Of course, while \oursimpl{} exhibits fewer major problems compared to baselines, some issues still remain. For example, in \oursimpl{}'s outlines, one issue is that some outline leaves may be vague, so that substantial creative work is left to the drafting stage. Additionally, some settings are problematic (e.g., not really locations) and sometimes character lists are incomplete. 

\oursimpl{}'s stories generally follow the high-level plan fairly well. However, as noted in the main text, some of the lower-level details are often missed. On occasion, the story will go somewhat off track by missing a few low-level details in a row, although it usually recovers later. Due to our early stopping criteria, the passages where \oursimpl{} fails to follow the outline unfortunately also tend to be the longest. There are unsurprisingly factual consistency errors as well, as addressing such errors is not the main focus of the \oursabstract{} framework. Finally, there are some minor style issues such as the tendency to repeatedly use characters' full names. 

All other plans and stories from all of our experiments can be found at \url{https://github.com/yangkevin2/doc-story-generation}, 
together with code and model checkpoints for generating new stories.

\onecolumn

\begin{small}
\begin{longtable}{p{0.9\textwidth}}
\toprule
\textbf{\oursimpl{} Plan 1}\\
\midrule
\texttt{Premise: After the loss of her father, Shannon is determined to follow in his footsteps and become a successful journalist. However, when she lands her first major assignment, she quickly discovers that the ugly reality of life in the city is far different from the dream she imagined. With the help of her new friend, a street-wise teenager, Shannon comes to understand the harsh realities of life in the inner city and learns that sometimes the truth is much more than just a story.}\\
\\
\texttt{Setting: The story is set in the inner city of a large metropolitan area.}\\
\\
\texttt{Characters:}\\
\texttt{1.}\\
\texttt{Full Name: Shannon Doyle}\\
\texttt{Character Portrait: Shannon Doyle is a young woman in her early twenties.}\\
\texttt{2.}\\
\texttt{Full Name: Gary Saunders}\\
\texttt{Character Portrait: Gary Saunders is a teenage boy who lives in the inner city.}\\
\texttt{3.}\\
\texttt{Full Name: Mike Doyle}\\
\texttt{Character Portrait: Mike Doyle is Shannon's father and a successful journalist.}\\
\texttt{4.}\\
\texttt{Full Name: Lena Saunders}\\
\texttt{Character Portrait: Lena Saunders is Gary's mother and a local business owner.}\\
\\
\texttt{Outline:}\\
\texttt{1. Shannon's father, Mike, dies unexpectedly, leaving her determined to follow in his footsteps and become a successful journalist. Scene:  Characters: Shannon Doyle, Mike Doyle}\\
\texttt{\quad \quad a. Shannon's father, Mike, dies unexpectedly. Scene:  Characters: Shannon Doyle, Mike Doyle}\\
\texttt{\quad \quad \quad \quad \textcolor{red}{i. Shannon's father, Mike, dies unexpectedly. Scene: Shannon's home. Characters: Shannon Doyle, Mike Doyle}}\\
\texttt{\quad \quad \quad \quad \textcolor{orange}{ii. Shannon inherits her father's estate. Scene: Shannon's home. Characters: Shannon Doyle, Mike Doyle}}\\
\texttt{\quad \quad \quad \quad \textcolor{olive}{iii. Shannon moves to the city. Scene: Shannon's home. Characters: Shannon Doyle}}\\
\texttt{\quad \quad b. Shannon decides to follow in her father's footsteps and become a successful journalist. Scene:  Characters: Shannon Doyle, Mike Doyle}\\
\texttt{\quad \quad \quad \quad \textcolor{teal}{i. Shannon applies for a job at a local news station. Scene: Shannon's home. Characters: Shannon Doyle}}\\
\texttt{\quad \quad \quad \quad \textcolor{cyan}{ii. Shannon's boss, the news director, assigns her to the inner city beat. Scene: Shannon's home. Characters: Shannon Doyle}}\\
\texttt{2. Shannon lands her first major assignment, a feature on the inner city, but quickly discovers that the ugly reality of life in the city is far different from the dream she imagined. Scene:  Characters: Shannon Doyle, Lena Saunders}\\
\texttt{\quad \quad a. Shannon lands her first major assignment, a feature on the inner city. Scene:  Characters: Shannon Doyle, Lena Saunders}\\
\texttt{\quad \quad \quad \quad \textcolor{blue}{i. Shannon lands her first major assignment. Scene: the newsroom of a local newspaper. Characters: Shannon Doyle}}\\
\texttt{\quad \quad \quad \quad \textcolor{violet}{ii. Shannon goes to the inner city to begin her assignment. Scene: the inner city. Characters: Shannon Doyle}}\\
\texttt{\quad \quad b. Shannon quickly discovers that the ugly reality of life in the city is far different from the dream she imagined. Scene: the inner city. Characters: Shannon Doyle}\\
\texttt{3. With the help of her new friend, Gary, Shannon comes to understand the harsh realities of life in the inner city and learns that sometimes the truth is much more than just a story. Scene:  Characters: Gary Saunders, Shannon Doyle}\\
\texttt{\quad \quad a. Shannon meets Gary. Scene:  Characters: Gary Saunders, Shannon Doyle}\\
\texttt{\quad \quad \quad \quad \textcolor{red}{i. Shannon meets Gary. Scene: the inner city. Characters: Gary Saunders, Shannon Doyle}}\\
\texttt{\quad \quad \quad \quad \textcolor{orange}{ii. Shannon and Gary become friends. Scene: the inner city. Characters: Gary Saunders, Shannon Doyle}}\\
\texttt{\quad \quad \textcolor{olive}{b. Gary teaches Shannon about the inner city. Scene: the inner city. Characters: Gary Saunders, Shannon Doyle}}\\
\texttt{\quad \quad c. Shannon learns that the truth is much more than just a story. Scene:  Characters: Shannon Doyle}\\
\texttt{\quad \quad \quad \quad \textcolor{teal}{i. Shannon Doyle meets Gary Saunders. Scene: the inner city. Characters: Gary Saunders, Shannon Doyle}}\\
\texttt{\quad \quad \quad \quad \textcolor{cyan}{ii. Shannon learns about the realities of life in the inner city from Gary. Scene: the inner city. Characters: Gary Saunders, Shannon Doyle}}\\
\texttt{\quad \quad \quad \quad \textcolor{blue}{iii. Shannon comes to understand that the truth is much more than just a story. Scene: the inner city. Characters: Shannon Doyle}}\\
\bottomrule
\caption{The first plan generated by \oursimpl{} in our main experiments. Colors indicate leaf nodes used to guide generation. The premise, setting, and character inventory make sense, although ``News Director'' is a failure of the name selection system inherited from \rethreeimpl{}. The outline is also largely reasonable, although the last section is still vague at times.}
\label{tab:plan1}
\end{longtable}
\end{small}

\begin{small}
\begin{longtable}{p{0.9\textwidth}}
\toprule
\textbf{\oursimpl{} Story For Plan 1 [Warning: Vulgarity]}\\
\midrule
\texttt{\textcolor{red}{[Shannon's father, Mike, dies unexpectedly. Scene: Shannon's home. Characters: Shannon Doyle, Mike Doyle]} The first time Shannon spoke to her dad after he told her about his plans was when he called that evening to say he was in the hospital.  The doctors had rushed him there when they noticed a rapid decline in his health. Shannon knew something was wrong with him, but she had no idea how serious it was. Mike was in his late 50s, and he still ate healthy and exercised on a regular basis. He seemed to be one of those people who could live forever without suffering any consequences.}\\
\\
\texttt{\textcolor{red}{}When he died, Shannon was shocked. She cried for hours and thought that everything would be okay. She figured that life would resume to normal for her dad. But then she had to think about the fact that he was gone forever.}\\
\\
\texttt{\textcolor{red}{}Shannon cried for days after her father died. She felt so alone now that he was gone, and she wished she could talk to him just one last time. Then one night, Mike came back unexpectedly. He had been dead for a few days, but he came as a ghost. Shannon was shocked when she noticed his presence, and was confused and frightened at first. Mike explained that he had come back to haunt her because he did not want her to feel so sad and alone anymore. He wanted her to feel \textcolor{orange}{[Shannon inherits her father's estate. Scene: Shannon's home. Characters: Shannon Doyle, Mike Doyle]} loved and cared for by him again, even though she knew he was dead.}\\
\\
\texttt{\textcolor{orange}{}Mike explained that since he had died unexpectedly, he really did not have much money saved away for his burial or anything else. So Shannon was going to inherit some of his assets, and a lot of money from a life insurance policy that he had taken out on himself. Shannon was shocked to learn that she would inherit a house and a car from Mike, as well as some cash. She could not believe how lucky she was to have such a large inheritance from her father, even though he had died.}\\
\\
\texttt{\textcolor{olive}{[Shannon moves to the city. Scene: Shannon's home. Characters: Shannon Doyle]} Shannon decided that she wanted to keep living in the house where she had spent most of her childhood. But she needed work to support herself until her inheritance came through, so she moved down to the city, where there were plenty of jobs available. Shannon applied for many jobs, and was finally able to secure a position as a copy editor at a local newspaper. Shannon did not regret moving away from the small town that she had grown up in, since there was nothing there for her except memories of her father. She knew that moving to the city would open up many new experiences and opportunities for her.}\\
\\
\texttt{\textcolor{teal}{[Shannon applies for a job at a local news station. Scene: Shannon's home. Characters: Shannon Doyle]} Shannon only worked at that newspaper for a few weeks before she was assigned to do a story on the rising crime rate in the city. At first she was not sure how she should tackle this assignment, but then she thought about what her father would do in this situation. If he were still alive, he would probably assign himself to covering the city beat, since he was such an experienced journalist. Shannon decided that she should take the same approach, and so she began to plan her story for the next day.}\\
\\
\texttt{\textcolor{teal}{}Shannon had decided that she would start her investigation at a local police station, since it was the most \textcolor{cyan}{[Shannon's boss, the news director, assigns her to the inner city beat. Scene: Shannon's home. Characters: Shannon Doyle]} logical place to get crime statistics. She arrived at the police station early the next morning and was greeted by a cheery young man who introduced himself as Officer Jack. Shannon told Officer Jack that she had come down to do a story about rising crime in the city, and asked if he could give her some statistics. Officer Jack was happy to help, and so he went back to his desk and got out a report filled with lots of statistics. Shannon was surprised that there were so many crimes committed in the city, but she realized that her assignment would be easier than she thought. Shannon was sure her story would be the lead story on the News at 6:00 that evening.}\\
\\
\texttt{\textcolor{cyan}{}After Shannon left the station, she picked up some breakfast and then headed to the inner city. She decided that this would be a good place to start her investigation, since the inner city was part of the news beat. Shannon was surprised when she saw her first inner city resident, because it was a woman dressed in rags who looked like she had not bathed in weeks. She was dirty, smelly, and carrying a shopping bag full of bottles and cans. Her hair was matted with dirt, and she spit constantly \textcolor{blue}{[Shannon lands her first major assignment. Scene: the newsroom of a local newspaper. Characters: Shannon Doyle]}. Shannon decided that this woman might be a good person to interview.}\\
\\
\texttt{\textcolor{blue}{}When Shannon introduced herself and asked the woman if she would be willing to talk to her, the woman responded rudely, saying she did not have time for interviews because she had to have her bottle of beer filled before noon so that she could get some sleep.}\\
\\
\texttt{\textcolor{blue}{}Shannon was amazed at the woman’s casual response to questions about crime, and decided she had to ask more questions. She told the woman that she was a reporter from the local paper trying to find out what people were talking about in the inner city. She asked if there was anything in particular on the news that the woman would like to talk about. The woman responded by saying, “Nothing is going on here but a bunch of fools and liars. Nothing else is going on here other than crime, ignorance, and stupidity.”}\\
\\
\texttt{\textcolor{blue}{}Shannon then asked what news reports the woman wanted to discuss. The woman said there were none because there were no news reports in the newspaper, because reporters did not care about people in the inner city.}\\
\\
\texttt{\textcolor{blue}{}At this point Shannon moved on to another subject, asking if there were any stories that she should cover. The woman responded, “In the news? Are you joking with me? News about how black people are being screwed over? How we have to drink piss water from a fountain? How we have to use our hands as toilets when there is no public restroom in sight? How we don’t have a say-so in our local and national government? News on the news? Do you think some white reporter will tell black people what’s happening in their own lives? Do you think if you tell us what the white man is trying to do to us that we are going to be glad that you told us \textcolor{violet}{[Shannon goes to the inner city to begin her assignment. Scene: the inner city. Characters: Shannon Doyle]}? News on the news?”}\\
\\
\texttt{\textcolor{violet}{}The woman continued, “You’re right. We don’t have anything to talk about because there is nothing happening here. Things are at a standstill, and nobody cares. Nobody in the inner city seems to care about anything, so how can you expect a reporter to write about us? So, no, I don’t have anything for you.”}\\
\\
\texttt{\textcolor{violet}{}Harsh words from the young woman. Shannon listened attentively and then thanked her for her time. She was not deterred by the woman’s hostile tone. Shannon moved on to the next interview.}\\
\\
\texttt{\textcolor{violet}{}“I want to talk about the high crime rate in your neighborhood,” Shannon said in a soft voice to the middle-aged man with a stubbly beard and unkempt hair.}\\
\\
\texttt{\textcolor{violet}{}“What high crime rate?” he responded, surprised and defiant.}\\
\\
\texttt{\textcolor{violet}{}Shannon looked around. It was obvious that the surrounding residents did not find the question shocking at all. She knew that whatever she found here would be different from what she had imagined, but this was far beyond anything she could have expected.}\\
\\
\texttt{\textcolor{violet}{}“Are there gang fights in this neighborhood?” Shannon continued.}\\
\\
\texttt{\textcolor{violet}{}“Only when the other neighborhoods come over here, but that doesn’t happen very often. That only happens once in a while. That ain’t high crime rates, that’s just plain old lowdown fighting. It ain’t high crime when the city lets the other neighborhoods run all over this neighborhood and do what they want to do here. That ain’t no high crime rate. That is just plain old ugly and nasty. Nobody cares about that here, because nobody cares about us no how, not even our own folks. They say everybody is equal \textcolor{red}{[Shannon meets Gary. Scene: the inner city. Characters: Gary Saunders, Shannon Doyle]}, but nobody is equal. I don’t see no colored folks running all over the other neighborhoods like they do here, and no one even tries to stop them when they come here and come at us.”}\\
\\
\texttt{\textcolor{red}{}Shannon had been taught as a child that racism was a thing of the past, but this man’s answer changed her opinion. She had never considered that people of color might feel discriminated against in America. He must be wrong, she thought to herself. No one can possibly think like that anymore.}\\
\\
\texttt{\textcolor{red}{}“Gary, what happened to your leg?” Shannon said as she looked at Gary’s left leg, which was wrapped in an Ace bandage.}\\
\\
\texttt{\textcolor{red}{}“Oh, Shannon. You really ain’t from around here, are you? You just ain’t from around here no how. You never been down here to get nothin? Never been down here to meet me and the rest of my boys? We was talkin’ ’bout you yesterday. Gary says you a fine-lookin’ young lady, but we need to see if that is true or not. Get your fine-lookin’ white self over here and meet the rest of Gary’s boys. We gonna see if you is as fine as Gary say you is.”}\\
\\
\texttt{\textcolor{red}{}Shannon was shocked to hear this talk from a grown man; she had never heard anything like it in her life before. She was surprised by Gary’s reaction, too, and by the way Gary was acting.}\\
\\
\texttt{\textcolor{red}{}Shannon stood motionless, stunned by what she had just heard. Gary motioned for Shannon to follow him as he headed out of the room. “Follow me,” Gary said with urgency in his voice and led her down the hall. “Meet me and Gary’s boys. Meet Gary. Meet some real men! Real men who fuck hard and take what they wants! Real men who fuck young white women! Meet real men who ain’t afraid to hit a fine-lookin white woman when she is fuckin with them and their business! Meet real men \textcolor{orange}{[Shannon and Gary become friends. Scene: the inner city. Characters: Gary Saunders, Shannon Doyle]}, Shannon. Find out what they all about!”}\\
\\
\texttt{\textcolor{orange}{}Shannon kept pace with Gary as he continued to talk. She followed him out into the street, then up the steps to a porch on one side of a two-story building. “What’s goin on?” Shannon asked Gary as she looked around the porch. There were six or seven young men standing around on the porch; Shannon had seen many of them before but never noticed how handsome they were. Gary was introducing her to them. “Meet my boys, Shannon,” Gary said with pride in his voice. Shannon shook hands with each of the young men. They were all very friendly towards her. They talked and joked with her. Gary then introduced her to a few more young men who came from another part of the neighborhood, and Shannon shook hands with them and shared friendly greetings as well. She became more comfortable as she \textcolor{olive}{[Gary teaches Shannon about the inner city. Scene: the inner city. Characters: Gary Saunders, Shannon Doyle]} got to know these men better; meanwhile, Gary was busy introducing her to more young men who had gathered outside.}\\
\\
\texttt{\textcolor{olive}{}Shannon found herself introduced to one young man after another, each of whom was very nice and polite. Most were wearing jeans and windbreakers or leather jackets; some wore baseball caps or fedoras. Shannon met several young women as well, and she shook hands with all of them. She noticed that most of these young men were black, although at least one was white. Shannon learned that there were two kinds of men in the inner city: “real” guys and “little” guys. Gary was a real guy—he had been to prison and had even shot someone. Shannon learned that one of Gary’s best friends had once been killed by his girlfriend while they were in bed together. Gary told her that this friend was “bad to the bone,” and that the world would be a better place if she learned about the kind of people who lived in the inner city. Gary said he did not have a job, but wanted to get one. Shannon learned that most of the guys on the porch had jobs, but none of them earned more than \$9 an hour. \textcolor{teal}{[Shannon Doyle meets Gary Saunders. Scene: the inner city. Characters: Gary Saunders, Shannon Doyle]} Gary said that most of the gang members she had met earlier were probably unemployed.}\\
\\
\texttt{\textcolor{teal}{}As Shannon and Gary sat chatting, one of the larger men in the group, who was nicknamed “Big Mike,” walked toward them with a bottle of beer in his hand. Gary stood up and introduced Shannon to Big Mike. Shannon shook hands with him.}\\
\\
\texttt{\textcolor{teal}{}“I’m Shannon Doyle,” she said. “I work for the Tribune. I have come here today to learn about your world. Gary Saunders told me I ought to meet you all.” Shannon introduced Gary as her friend, and he nodded.}\\
\\
\texttt{\textcolor{teal}{}Gary introduced Shannon to the rest of the group, including a teenage boy who was nicknamed “Little Mike.” Little Mike was white and about 5 feet 8 inches tall. He looked more like a 14-year-old than an 18-year \textcolor{cyan}{[Shannon learns about the realities of life in the inner city from Gary. Scene: the inner city. Characters: Gary Saunders, Shannon Doyle]} -old.}\\
\\
\texttt{\textcolor{cyan}{}Gary put out his hand for Shannon to shake, and she did the same for Little Mike, but he just stared at her with his mouth hanging open.}\\
\\
\texttt{\textcolor{cyan}{}Big Mike handed Shannon and Gary each a beer, but only Shannon accepted it. Gary didn’t drink alcohol at all.}\\
\\
\texttt{\textcolor{cyan}{}Shannon asked Gary how long he had lived in the inner city.}\\
\\
\texttt{\textcolor{cyan}{}“For a long time,” Gary said. “I was born here. My parents were immigrants from Ireland, and they lived here for a long time too before they moved back to Ireland when I was 10 or 11 years old.”}\\
\\
\texttt{\textcolor{cyan}{}Shannon learned that Gary was 18 years old and that he had never been employed in his life. His parents had died when he was 8 years old, and it fell to an aunt to raise him. He had dropped out of high school two years earlier because of being bullied at \textcolor{blue}{[Shannon comes to understand that the truth is much more than just a story. Scene: the inner city. Characters: Shannon Doyle]} school for the clothes he wore and for the fact that he wasn’t interested in sports.}\\
\\
\texttt{\textcolor{violet}{}“I didn’t want to wear the latest sneakers or hats that everyone at school was wearing, and I couldn’t care less about who wins a game of soccer or hockey. I just didn’t fit in, Shannon.”}\\
\\
\texttt{\textcolor{violet}{}“Didn’t you ever tell your parents about the bullying?”}\\
\\
\texttt{\textcolor{violet}{}“No way, Shannon! I wasn’t going to admit to my parents that I was being bullied at school. They would have wanted to come down here to talk personally with the principal and the bullies and try to work out a solution. It also wouldn’t have been helpful in anyway. It simply didn’t occur to me then that telling my parents everything that was happening was the right thing to do.” Shannon noticed that Gary’s voice trailed off into silence.}\\
\\
\texttt{\textcolor{violet}{}“Tell me something else about yourself, Gary, and tell me the truth this time. Why is it so important for you to tell me the truth? Why do you want to be honest? You know that I want to be a journalist. Truth is more than just a word. It’s everything around you and everything you accept as true. Everything that is true is important, and it always has been. Even stories have some truth in them, or are based on something real from the past or present. But stories aren’t real, they are just someone else’s perception of the truth. The truth is what actually happened, or what is happening. Sometimes people don’t want to believe the truth because it’s too hard for them to accept. I want to know the truth about you and this place, Gary. Please, tell me the truth.” Shannon looked at Gary with pleading eyes.}\\
\\
\texttt{\textcolor{violet}{}Gary took a deep breath and began to speak. “I wanted to be honest with you because I thought you could help me, and because I thought you were different from all the other people who have come down here. Everyone always asks the same questions and they never really listen to the answers. I can tell that you’re different, Shannon. You actually seem like you care.”}\\
\\
\texttt{\textcolor{violet}{}Shannon nodded her head in understanding. “I do care, Gary. I want to help you if I can.}\\
\bottomrule
\caption{The story generated by \oursimpl{} for the plan shown in Table \ref{tab:plan1}. Colored text in brackets indicates the corresponding outline items for the following text. While not every passage  corresponds exactly to the low-level outline item it is based on, the overarching plot is coherent and reasonably faithful to the outline as a whole. }
\label{tab:ours_ex1}
\end{longtable}
\end{small}

\begin{small}
\begin{longtable}{p{0.9\textwidth}}
\toprule
\textbf{\rethreeimpl{} Story For Plan 1}\\
\midrule
\texttt{The sun was shining brightly through the windows of Shannon Doyle’s apartment. It was only a little after 7:00 in the morning, but she had already been awake for hours. Today was a big day for Shannon. She had just landed her first major assignment and was anxious to begin writing. She hadn’t slept well the night before, and she spent a great deal of time thinking about what she wanted to write.}\\\\
\texttt{Graduating journalism school and landing her first job in a major newspaper had been Shannon’s greatest accomplishment to date. Ever since she was a child, Shannon had dreamed of becoming a successful journalist just like her father. She would spend hours staring at his framed diploma and reading the articles he had written. She wanted to be just like him.}\\\\
\texttt{Her father, Mike Doyle, had been an excellent writer and reporter. Through the years, he had received many awards for his work in newspapers and magazines throughout the country. He started out at a small-town newspaper before moving on to larger cities and eventually landing a job as a reporter for The Washington Post. Mike was well respected in his field and often traveled all over the world to cover stories. He wrote about countries suchas South Africa, Russia and Egypt. He had interviewed such celebrities as Nelson Mandela, Mikhail Gorbachev and President Hosni Mubarak. Many would agree that he was one of the best journalists in his field.}\\\\
\texttt{As an adult, Shannon followed in her father’s footsteps and enrolled in the School of Journalism at The University of Maryland. She felt certain that she would be able to secure a good job soon after graduation. Shannon had always considered herself a very hard worker, and she knew that her dedication would pay off in the long run. Although she hadn’t landed a job right away, Shannon remained patient and confident that her dream would come true. At last, after months of submitting her resume and applying for numerous jobs, she received a phone call from The Washington Post offering her an internship position at their paper.}\\\\
\texttt{Today was Shannon’s first day as a full-time intern for The Washington Post, and she was anxious to begin reporting on her first assignment. She quickly got dressed and headed down to the kitchen where she made herself a cup of tea. She knew that it was bad for her teeth, but she just couldn’t start her day without a cup of hot Earl Grey in hand.Shannon was born to a middle class family in Rockville, Maryland. Her mother, Lena, owned and ran her own business; she had a knack for interior design and had a successful career in that field for many years. The two of them lived together in a modest home on the outskirts of the city with their two children, Gary and Shannon. Her parents’ marriage wasn’t the happiest one, but they managed to stay together until his death three years ago. Mike Doyle was diagnosed with colon cancer at 49 and passed away only three months later.}\\\\
\texttt{Upon graduating from the University of Maryland, Shannon moved into her first apartment in College Park. She quickly became accustomed to living alone and spent many evenings with friends at local bars. Although she had grown up in a somewhat religious household, Shannon decided that she no longer believed in God or an after life and became very promiscuous. She had slept with over 20 different men during her first year of college and continued to sleep around even when she began her internship at the Washington Post.}\\\\
\texttt{As Shannon sat down at the kitchen table with her cup of tea, she glanced around at her family pictures hanging on the wall. She wanted to make a goodimpression on her editor and hoped that he would be impressed with the photos of her hanging out with world leaders, former presidents, and other important people. Shannon’s parents were both deceased, but she was especially close to her dad. She had always admired his passion for life and his strong ambition to be successful in everything he did. Even though he was only 49 when he passed away, Shannon knew that her father was extremely talented and well respected in the journalism world. He had worked as a reporter for The Post for 23 years before succumbing to cancer. Shannon’s job was extremely competitive, and she knew that she would have to work very hard in order to succeed. Her greatest fear was not being able to live up to what was expected of her, and she also feared that her new editor might be disappointed by her work. The last thing that Shannon wanted was for anyone to be disappointed by her. She loved being a journalist; it had always been her dream career, and she worked hard at honing her craft. After finishing her tea and taking one last look at the framed photographs on the wall, Shannon grabbed her bag from the counter and headed out the door. As she walked into the office building where The Washington Post was located,Shannon looked around and took in the sights. She noticed how beautiful everything was; it was a vast difference from the way that she imagined it to be. Her first thought after entering the building was that she had no idea what to expect. She wasn’t sure how many people she would meet or what her workspace would look like. All she knew was that she wanted people to think highly of her, and that meant having great ideas and being able to write well so that her editor would be impressed by her work.}\\\\
\texttt{The walk to the newsroom took approximately 10 minutes. Once inside, Shannon realized how big the building actually was. There were tons of hallways with doors that opened into countless offices, and it was hard for Shannon to find hers. After walking down a couple of hallways and turning a few corners, she finally found her work space tucked away in a corner of the newsroom. It took Shannon some time to get organized before she could start working on her assignment. It felt like each time she turned around, someone else had created some sort of problem that she had to solve. All of the other interns seemed to have their assignments ready, but none of them seemed to be having any problems; they all looked like seasoned professionals who had already learnedhow to get their work done with ease. Shannon realized that she was being a bit over-dramatic. She knew how to do the work, but was just having trouble getting started.}\\\\
\texttt{After she finished setting up her workspace, Shannon put on her headphones and began to write her script. At first, she thought it would be easy for her to simply pull up an excerpt from a book about the topic of urban life and then use it as reference for her script. She figured that she should have at least a basic understanding of what she was going to be talking about before writing her script.}\\\\
\texttt{Shannon walked into another room and picked up three books with the intention of reading through them before beginning to write her script. The topic that she wanted to write about was urban life, so she grabbed two books on the subject and chose a third book as a possible reference. The first two books appeared to be about a similar topic, but Shannon was surprised to learn that they were actually completely opposite in content; one emphasized the negative aspects of urban life while the other emphasized its positive features. Shannon quickly realized that both books were more about sociology than they were about anything else. The books ended up being little more than compilations of large amounts of information that couldnot be used for reference in an article, let alone a script!}\\\\
\texttt{Shannon did not know what to do at this point. If she needed to write her script without any real knowledge of the topic then she would have to base her story on the opinions of others. However, she knew that relying heavily on opinions could be quite dangerous, especially when talking about such a controversial topic as urban life.}\\\\
\texttt{Shannon thought about going back to the library and checking out another book or two with a more literary focus; something that would help her better understand the subject matter. However, she knew she would be pressed for time if she tried to read through another book before starting her script. Still, Shannon thought she might have enough time if she skips going out that night and stays in to work on her script. After a moment of indecision, Shannon decided to stay in and work on her script. She really wanted to do well on her first assignment and knew that staying in and working on her script would be the only way for her to do so.}\\\\
\texttt{Shannon quickly finished organizing all of the paperwork on her desk, which allowed her to focus more clearly on the task at hand. She walked back into the room, grabbed three pieces of paper and began writing downall the questions that she felt needed answering. As she wrote, Shannon realized that each of her questions led to even more questions. Some of these questions were simple ones that could be answered with a quick phone call or a few minutes of research while others may require hours spent digging through public court records and old newspaper articles.}\\\\
\texttt{Shannon also realized that each question she asked had been asked by other journalists in the recent past. In fact, just looking at her notes, Shannon found four different articles written by various authors that had each provided their own answers to almost all of the questions Shannon had asked herself. Each article was about a different aspect of urban life and each had been written over the last ten years.}\\\\
\texttt{Shannon decided to keep this information in mind as she worked on writing the rest of her script. She knew she did not want to copy any other author’s work, but it was important for her to have a strong understanding of what others had previously covered on this topic. By gathering as much information as possible, Shannon would be able to form her own opinions about the subject and then write an original script based on those opinions. She also knew that with so many articles with such different interpretations it was going to be hard for her to find any one answerthat would be able to encompass all of her thoughts on the subject.}\\\\
\texttt{She decided to begin with the most simple questions first. She picked up the phone and dialed the number of a man from the National Urban League and asked him if he could provide some basic demographics about Washington, D.C.}\\\\
\texttt{After a few minutes of small talk, Shannon asked her question and was pleasantly surprised when she received a detailed answer from the man on the other end of the line.}\\\\
\texttt{“The district has a population of roughly 615,000 people and over 51\% of those residents are African American. The majority of the residents are between the ages of 25 and 64, but there are large numbers of children living in this area as well. There is also a large gay population here, although the numbers have continued to decline from their peak in the early nineties.”}\\\\
\texttt{Shannon thanked him for his time and thanked him for providing her with such a detailed answer. He informed Shannon that if she ever had further questions about the subject she could call the Urban League at any time and they would be happy to assist her.}\\\\
\texttt{Still feeling excited from obtaining such a quick response to her question, Shannon pulled up the front page of the Washington Post and began reading through thearticles. She had been reading for about an hour when the editor of the paper called her into his office. When she arrived, he handed Shannon an envelope that contained some background information on her first assignment and told her it was due in two weeks.}\\\\
\texttt{The editor was a man named Gary Saunders. He was sixty-five years old with thinning gray hair and a heavy build. He walked with a slight limp, but he managed to make it around the newsroom without much trouble. Gary’s office was small, but comfortable and well decorated with pictures of his family on his desk and various awards he had won throughout the years in other offices around the newsroom. Lena Saunders was Gary's mother, a local business owner. She had a deep voice, but she was kind and wise.}\\\\
\texttt{She thanked him for the envelope and went back to her desk to finish her research for her script. The phone rang about an hour later. She picked up the receiver and a woman with a deep voice asked if she could speak with Shannon Doyle. She nodded and told her that she was on line one. The woman introduced herself as Lena Saunders, Gary’s mother and the publisher of the local newspaper in Rockville, Maryland.}\\\\
\texttt{Mrs. Saunders asked Shannon if she would be interested in doing a profile on Mike Doyle for an article they were writing for the local paper about local business owners. Shannon eagerly agreed and Mrs. Saunders gave her Gary’s phone number and address. After thanking her, Shannon wrote down all of the information on a small yellow pad and then sat down to do some more research.}\\\\
\texttt{She had not been able to talk to Mike the day before, but she had an appointment with him at eight o’clock that very morning. She got up from her desk, logged off of the computer and locked up her notes in her office. She walked out of the building and headed toward the metro station to catch a train downtown.}\\\\
\texttt{Shannon rode the metro downtown with a few of her co-workers, who were going to an office party that evening. Shannon had been invited, but she had already made plans with Gary that night and she did not want to back out at last minute. She was sure everything was going to be fine and she would be able to work things out with Mike and his family.}\\\\
\texttt{Shannon walked into the office building where The Doyles’ restaurant was located, and handed her driver’s license to the security guard at the front desk. Luckily, there was no wait inside and Mike greeted Shannon warmly when she walked in.She sat down in a booth with him and sipped her coffee as they discussed her writing assignment. Shannon’s mind raced with questions about Mike and his family. She wanted to know everything about him, but she did not want to be overwhelming since they had only just met.}\\\\
\texttt{Mike explained that he had started his own business in Rockville because he wanted to move his family out of the city. He did not like the idea of sending his wife and two boys into some of the worst neighborhoods in Baltimore every day for work. When he opened up his restaurant, business was slow at first, but within six months most of the surrounding restaurants shut down and reopened as Doyles’ franchises. His restaurant was the most popular place to eat in Rockville and it had one of the highest ratings on Zagat’s web site. Mr. Doyle seemed proud of his success and was more than happy to talk about his family to Shannon for hours that morning.}\\\\
\texttt{Even though she had already spent three hours with him, a lot of Mike’s story remained a mystery to Shannon. She wanted to learn more about him and his family but he seemed reluctant to talk about them further. He changed the subject and started talking about his wife, Elizabeth,and their two boys. Shannon didn’t want to be rude, so she played along and let Mike ramble on about Liz and the kids. He told her that they were all very close and often went on spontaneous trips together.}\\\\
\texttt{Shannon asked Mike how old his boys were and he proudly proclaimed that his oldest son, Edward, was fourteen years old and was an excellent student. She smiled when he mentioned that Edward had a knack for math and planned on attending Johns Hopkins University after high school. Shannon’s heart sank as she looked at Mike in disbelief. She knew that in most neighborhoods it was hard to get into college, let alone graduate from one of the best universities in the country.}\\\\
\texttt{Mike continued to talk about his oldest son. He told her that Edward volunteered at a local community center and had recently been accepted into a prestigious summer program at Georgetown University. Shannon wanted to scream at him, “What about your younger son? Surely he isn’t living some sad, lonely existence in the inner city while his older brother gets a chance to come to Georgetown?” But she knew that Mike would never volunteer such information unless she asked him directly.}\\\\
\texttt{Mike also went on and on about his youngest son, Gary. Before Shannon couldstop herself, her mouth dropped open when she heard him say that Gary was only ten years old. Ten. All of a sudden she knew why Mike struggled to have a conversation about his family; he was lying to make himself seem better than everyone else in the room.}\\\\
\texttt{Shannon excused herself from the table and rushed to the ladies’ room where she locked herself into a stall, pulled out her cell phone and called Gary. She had met Gary earlier that evening during her interview with Mike, but they had only spoken on a casual basis. The phone rang four times before it was picked up.}\\\\
\texttt{“Hello?” said a high-pitched voice on the other end of the line.}\\\\
\texttt{Shannon was confused by the familiar voice because she had just realized that she had been trying to call Mike this entire time and not Gary. She immediately apologized for the mistake and explained that she was trying to contact Mike, not Gary, but when she thought about it, it made perfect sense because both her interview with Mike and her awkward phone call were as confusing as a bad dream.}\\\\
\texttt{Gary must have sensed the tension in Shannon’s voice or perhaps he overheard her talking to herself because he told her that everything was okay. Shannon was relieved thatGary was so calm, but she wondered why he wasn’t more concerned that his father had just been caught lying to her. The phone call lasted only a few seconds before Shannon hung up. She had no idea why she called Gary, but at least she felt a little better after speaking to him. She quickly returned to the table where Gary and his mother were waiting patiently for her to return. No one mentioned the phone call and they all seemed to enjoy their dinner after Mike finished telling them about his favorite vacation spots and how he had recently visited Hawaii.}\\\\
\texttt{Mike went on for about twenty minutes before Shannon had had enough of his constant chatter. Mike thought he was smooth and charming, but everyone else found him boring. As soon as she felt uncomfortable, Shannon excused herself again and went back upstairs to the hotel lobby. It would be an understatement to say that she was confused by what had just happened. She didn’t know Mike from anyone else, and she wasn’t interested in getting to know him any better because his answers were so unrealistic. She knew he never left Rockville for vacations or anywhere else for that matter, so how could he possibly be an expert on Hawaii? She needed some sleep before she had to go back downstairs and try to act like everything was okay.}\\
\bottomrule
\caption{The story generated by \rethreeimpl{} for the plan shown in Table \ref{tab:plan1}. It somewhat follows the high-level plan, but not as well as \oursimpl{}.}
\label{tab:re3_ex1}
\end{longtable}
\end{small}

\begin{small}
\begin{longtable}{p{0.9\textwidth}}
\toprule
\textbf{\rollingopt{} Story For Plan 1}\\
\midrule
\texttt{Shannon paced the floor of her father’s rundown apartment. The smell of fresh paint lingered in the air, and the walls were a dull, lifeless tan.  There was a strange peace to the place that Shannon couldn’t quite place. Maybe it was because there was no sign of her dad, not even a trace of his ever-present cigar smell.}\\\\
\texttt{In a few short hours, she would stand at the grave of a man who had been her entire world, and it was all happening far too fast for her to process.   If he could have died in any other way, she would have been happy. Anything except the way he had. It was too sudden and made no sense at all.}\\\\
\texttt{Her eyes fell on the table in the kitchen, and then she saw it--the box sitting on top of his neatly stacked bills.}\\\\
\texttt{She plucked the box from the table and opened it, fully expecting to find a new set of golf clubs or an expensive fishing reel. Instead, it was filled with old newspapers from the seventies and eighties.  She pulled one from the box at random and ran her fingers across the crackling yellowed paper. A photo of dozens of kids standing around a fire, throwing rocks into it, filled the top half of the page. Some of the kids in the picture were only half her age.}\\\\
\texttt{“You can’t keep those here,” she said to herself. “What if they catch on fire?” When she turned to carry the box to the trashcan, her eyes fell on another stack of newspapers next to them.}\\\\
\texttt{“Oh my God,” she whispered.}\\\\
\texttt{Her father had been a reporter, a real one. He had worked for a newspaper called The Daily and had been dedicated to getting the truth out to the public.  If he were here, he would be appalled that she was throwing away his work.  He would probably take it from her and bury it himself.}\\\\
\texttt{Stumbling back to the table, she ripped a stack of papers from the box and ran them through her fingers. She had no idea what any of them were about, but she could tell that they were important.  They were stories that mattered, stories people cared about.}\\\\
\texttt{She tucked the papers into the box and shoved it into her duffle bag, then began to search for more. It was too painful to go through them all, but she was determined to do it.}\\\\
\texttt{“What are you doing?” a voice asked from behind her. “I’ve seen you pass by my stoop a few times. You look familiar, but I can’t remember where I’ve seen you.”}\\\\
\texttt{Shannon turned to see a pretty girl with long curly hair, dressed in worn jeans and an old tank top, standing behind her.  “I’m sorry,” Shannon said, still looking at the papers. “I don’t mean to be rude. I’m just trying to get my dad out of the place.”}\\\\
\texttt{She turned to face the girl, but her eyes caught on something behind her. There was a new photo in the window of the apartment directly across the street.  It was hard to make out what was happening, but she could see what looked like a bunch of teenagers surrounding an old man. They were throwing rocks at him as he stood there, cowering.}\\\\
\texttt{“I’ve seen that picture before,” the girl said. “Did you know your dad took it?  He was always carrying around his camera.”}\\\\
\texttt{Shannon didn’t know what to say. Her eyes remained glued to the photo.}\\\\
\texttt{“My name is Lisa, by the way,” the girl said, gesturing to herself.}\\\\
\texttt{Shannon looked at her and nodded, but didn’t speak.}\\\\
\texttt{The girl moved closer. “You don’t have to throw them out. I can use them, if you want me to. I work on a different paper than your dad did, but maybe we can help each other out.  Your dad and I were friends.”}\\\\
\texttt{Shannon couldn’t look into the girl’s eyes. She took the box from her bag and shoved it into Lisa’s arms.  “My name is Shannon,” she said quietly. “Thank you for the offer, but I can’t take them. I have to get rid of them.”}\\\\
\texttt{She turned to run away, but Lisa stopped her by grabbing her arm gently.}\\\\
\texttt{“I understand. I was just trying to help,” Lisa said softly. “I know what it feels like to lose someone you love, but it will get better.”}\\\\
\texttt{The truth of what she said cut through Shannon’s pain and she looked into Lisa’s eyes, searching for the truth.  She saw a pain there that mirrored her own, but also an assurance that things would get better.}\\\\
\texttt{“Thank you, Lisa. I think I just needed to hear that.”}\\\\
\texttt{Shannon turned and walked away, down the stoop and onto the street. She didn’t know where she was going, but she was finally ready to leave the horrible apartment.}\\\\
\texttt{She walked for hours, exploring the city that she had always dreamed of living in. She went into a diner and ordered a burger with fries and a coke, but she barely touched it. Her stomach felt unsettled, and she wasn’t sure why. She felt like something was wrong, or missing, but she couldn’t understand what it was.}\\\\
\texttt{She wandered through the streets of the city, trying to sort out her feelings. She had taken the box of her father’s work and left it in the trash, but she felt even emptier than before. She hadn’t found any sense of peace in the photos, and she knew Lisa was right: they only made her loss more painful.}\\\\
\texttt{She walked among the people on the streets, and watched the cars race by.  It was like everyone else had a life that was going somewhere, but she felt like she was just standing there watching it all happen.}\\\\
\texttt{That’s when she saw him. The boy from the photo, with the old professor who had found the box on the subway. They were walking along the street, laughing and joking with each other as they went along. They didn’t seem to notice her as she approached, but when she was within a few feet of them, she heard them talking and stopped.}\\\\
\texttt{“I told you I wasn’t going to do it,” the boy said.}\\\\
\texttt{“You said that the other times, too, but you still did it. What makes this time different?”}\\\\
\texttt{“Because I said so!” he said with a laugh, as he ran in front of the old professor and began to chase him.}\\\\
\texttt{The old man began to run down the street, trying to put some distance between himself and his young pursuer. The boy ran in front of cars, dodged around people, and pushed past others in an effort to catch him.}\\\\
\texttt{As the boy ran, Shannon saw him from a different perspective. She looked beyond the dirty clothes and the unkempt hair to see his face, and she saw a boy who was full of life and laughter. She saw a boy who was always smiling and always joking with others, but it was obvious that it wasn’t because he didn’t care. It was because he cared too much, and he was afraid of losing those he loved.}\\\\
\texttt{The boy caught up to the old man, who was now walking briskly down the street.  The boy ran to his side and grabbed his arm, pulling him to stop.}\\\\
\texttt{“I heard what you said, but that doesn’t matter. It doesn’t matter how many times I do it as long as I don’t do it again. And I won’t.”}\\\\
\texttt{The boy smiled at the man, and the two of them began to walk along the street together. Shannon felt a wave of relief go through her as she watched them go, and she had finally realized what was missing from her life. She turned to leave the street and headed back to the city to begin her story.}\\\\
\texttt{Shannon went home and decided to take a nap before she began to work on her story. She needed a rest, but she didn’t turn off the light. She looked through the large window of her apartment to watch the people hurrying along the street below. They seemed to be in such a hurry, but they all had somewhere to go, someone they were trying to reach. She sat on the couch and pulled the box of photos out, looking through them again as she thought about what people did with their lives.}\\\\
\texttt{Shannon had always been fascinated by the stories presented in the newspaper and on the news programs. When she was a child, she wanted to be a journalist so that she could tell those stories herself. She loved the power they had over people, how they could change lives and open people’s eyes to the world around them. People might not believe every word they saw, but at least they had the power to make a difference in the world. She could make a difference, too. She wanted to see things that other people didn’t see.}\\\\
\texttt{She flipped through the album and stopped to look at a picture of two boys sitting on a stoop, smiling for the camera. It was a sunny day, and one of the boys had his arm around the other, as though he was keeping him warm. The picture made Shannon feel warm inside, and she smiled as she thought to herself that no one could tell them that they had no future; not with a picture like that.}\\\\
\texttt{Shannon closed the photos and pulled out her laptop to begin her story. She wanted to capture the life and spirit of the people of the inner city, but first she needed to learn about them. She needed to talk to them and gain their trust, so that they would be willing to share with her. That was something she couldn’t do by herself.}\\\\
\texttt{As she started to type, Shannon thought about all the people she would have to talk to in order to conduct this interview. She needed to speak with the people at the shelter, where the children lived. She needed to speak with the people in the neighborhood who knew her subjects and who knew their stories. She needed to speak with the teachers who taught them and their parents who loved them…and, for some of them, who had lost them.}\\\\
\texttt{“Mom? Mom!” Shannon’s mom came running into the living room. “You have to come and see this! There’s a news van out front.”}\\\\
\texttt{Shannon stood, pulling herself out of her thoughts. “Really? Do you think it’s about me?”}\\\\
\texttt{Her mother shook her head. “Probably not. But I thought you would want to know, in case they wanted to interview you.”}\\\\
\texttt{“I would,” Shannon agreed. She hopped off the couch and followed her mother to the front window of their apartment. She could see a small news van parked on the street, and she watched as people around her ran to their front doors to see what was going on.}\\\\
\texttt{“I don’t think they would be here for me though,” Shannon said. “It’s probably just a celebrity or something.”}\\\\
\texttt{Her mother nodded. “But you never know,” she said. “It could be a story about the kids in the shelter or something like that – it would be great publicity for them.”}\\\\
\texttt{Shannon thought back to her photo album and the picture of the two boys, growing up together in the inner city. She hoped that, if she was able to tell their story, it would help others see them and their future with new eyes.}\\\\
\texttt{Friday, November 23, 2012}\\\\
\texttt{The prompt for this week is as follows:}\\\\
\texttt{Write a story in which a character has to make a choice; the choices are:}\\\\
\texttt{a) continue on the path they are on now, to its natural conclusion (no matter what that may be); b) make a major life change; c) go back in time and change things.}\\\\
\texttt{My premise is as follows: A woman who has spent her life exploring the exotic and the dangerous, both in her work and in her personal life, faces a new challenge that could jeopardize everything she has worked so hard to achieve.}\\\\
\texttt{Current Story Outline:}\\\\
\texttt{It was only a dream - just a dream. She knew she had to believe that, because if it had been anything else, her hands wouldn’t have been shaking and her heart wouldn’t have been racing.}\\\\
\texttt{They must have simply been responding to the stress of the day; how could they not be? There were too many variables, too many things that could go wrong. She had to remember that.}\\\\
\texttt{She was in the middle of a meeting with her team, preparing for a very important contract that the company had been chasing for years. And now, at this late date, they finally had their chance of winning the business. This was a big deal for them and a big deal for her. This was supposed to be the pinnacle of her career - a major step towards the Senior VP position that she had been gunning for.}\\\\
\texttt{She had been working all night on the presentation, checking and double-checking the numbers. She’d had to do it herself because her team members had all refused, saying they didn’t want to stay up all night before a meeting. They’d told her to go ahead; she could do it herself, and they would help her as much as they could in the morning. But she didn’t want to be the one to make a mistake, and she wanted to make sure she had everything in order before presenting it to the client.}\\\\
\texttt{But now, here she was, standing in front of them, rambling on about changes in market share and market growth and customer response, when all she really wanted to do was push the whole mess over onto someone else and run out of the room screaming.}\\\\
\texttt{She was sure that everyone could sense her mood, and they must have been wondering what was wrong with her. It was as if she couldn’t speak or move, and it was all she could do to keep from running out of the room.}\\\\
\texttt{“Miss Martin?” one of the younger members of her team finally said. “Maybe you should get some rest? You’ve been up all night.”}\\\\
\texttt{She looked into his eyes, and for just a moment, she thought he might be right. But she couldn’t do that; it would be disastrous for the company and disastrous for her. No, she had to just suck it up and get through the meeting. Once she was done, she could finish her presentation and get some sleep. She wasn’t exactly sure how she was going to do it, but she knew that if she broke down now, after all this time and effort, it would be the worst thing she could do.}\\\\
\texttt{“All right,” she said, “let’s continue with the meeting.”}\\\\
\texttt{And the meeting went on, and she continued to talk about everything she had been preparing for days. Even though she was in the middle of a presentation, she kept checking her watch, counting down the minutes until it was over.}\\\\
\texttt{She thought it was almost over when there was a knock on the door.}\\\\
\texttt{“Come in,” she said, not even bothering to look at whoever had entered the room.}\\\\
\texttt{She heard the door open and close, and she kept talking, even though it sounded as if someone had started to walk across the room. And when she finally looked up and saw who had entered the room, her heart stopped beating for a second.}\\\\
\texttt{It was her father.}\\\\
\texttt{He was standing in the doorway, looking at her with a smile - one of those smiles that she remembered from when she was little.}\\\\
\texttt{“Hi,” he said, and she finally realized that it was not really him, because he was too tall and too thin. This man’s shoulders were broad, his frame was thin, and his hair was gray.}\\\\
\texttt{“You scared me,” Shannon said, but the man who was standing in front of her continued to smile.}\\\\
\texttt{“I’m sorry. I almost didn’t come,” he said. He started to walk toward her, and she noticed a woman standing to his side. She didn’t look familiar, and Shannon thought that she must be the person he was with last night.}\\\\
\texttt{“Shannon?” her father said, and she finally realized that it was him. “It’s me, your dad.”}\\\\
\texttt{She couldn’t believe it; he was standing in front of her. They hadn’t spoken since her mother had committed suicide, and he had never once come to visit after she had been sent away to live with her aunt. But after everything that had happened, after leaving her aunt’s house and returning to the city, she wanted to see him. She needed to see him. She wasn’t sure if she wanted to hug him or slap him, but she wanted to see him.}\\\\
\texttt{“Hi, Dad,” she said.}\\\\
\texttt{She didn’t know what to say or what to do; it was so strange, seeing him standing there in front of her. Finally, he took a step forward and held out his arms.}\\\\
\texttt{“Shannon, we need to talk about your}\\
\bottomrule
\caption{The story generated by \rollingopt{} for the plan shown in Table \ref{tab:plan1}. \rollingopt{} struggles heavily to maintain overarching plot coherence, with some jarring changes in narration.}
\label{tab:rollopt_ex1}
\end{longtable}
\end{small}

\begin{small}
\begin{longtable}{p{0.9\textwidth}}
\toprule
\textbf{\oursimpl{} Plan 2}\\
\midrule
\texttt{Premise: A young woman is determined to never get married and live her life alone, but when she meets a man who seems perfect for her, she begins to rethink her decision.}\\\\
\texttt{Setting: The story is set in a small town in the United States.}\\\\
\texttt{Characters:}\\
\texttt{1.}\\
\texttt{Full Name: Jenna Adams}\\
\texttt{Character Portrait: Jenna Adams is a young woman in her early twenties.}\\
\texttt{2.}\\
\texttt{Full Name: Brian Johnson}\\
\texttt{Character Portrait: Brian Johnson is a man in his early thirties.}\\
\texttt{3.}\\
\texttt{Full Name: Mary Smith}\\
\texttt{Character Portrait: Mary Smith is Brian Johnson's mother.}\\
\texttt{4.}\\
\texttt{Full Name: Teresa Campbell}\\
\texttt{Character Portrait: Teresa Campbell is Brian Johnson's ex-girlfriend.}\\
\\
\texttt{Outline:}\\
\texttt{1. Jenna Adams meets Brian Johnson and immediately feels drawn to him. Scene:  Characters: Jenna Adams, Brian Johnson}\\
\texttt{\quad \quad a. Jenna Adams meets Brian Johnson and feels an instant connection to him. Scene:  Characters: Jenna Adams, Brian Johnson}\\
\texttt{\quad \quad \quad \quad \textcolor{red}{i. Jenna Adams meets Brian Johnson for the first time. Scene: a small town in the United States. Characters: Jenna Adams, Brian Johnson}}\\
\texttt{\quad \quad \quad \quad \textcolor{orange}{ii. Jenna Adams and Brian Johnson start talking and quickly become friends. Scene: a small town in the United States. Characters: Jenna Adams, Brian Johnson}}\\
\texttt{\quad \quad b. The two of them start dating and Jenna Adams begins to fall in love with Brian Johnson. Scene:  Characters: Jenna Adams, Brian Johnson}\\
\texttt{\quad \quad \quad \quad \textcolor{olive}{i. The two of them start dating and Jenna Adams falls more in love with Brian Johnson with each passing day. Scene: a small town in the United States. Characters: Jenna Adams, Brian Johnson}}\\
\texttt{\quad \quad \quad \quad \textcolor{teal}{ii. However, Brian Johnson's mother, Mary Smith, disapproves of Jenna Adams and does everything she can to break them up. Scene: a small town in the United States. Characters: Jenna Adams, Mary Smith, Brian Johnson}}\\
\texttt{\quad \quad \quad \quad \textcolor{cyan}{iii. Nonetheless, Jenna Adams and Brian Johnson's relationship continues to grow stronger. Scene: a small town in the United States. Characters: Jenna Adams, Brian Johnson}}\\
\texttt{2. Jenna Adams starts to think that maybe marriage isn't so bad after all when Brian Johnson seems like the perfect man for her. Scene:  Characters: Jenna Adams, Brian Johnson}\\
\texttt{\quad \quad a. Jenna Adams starts to think that maybe marriage isn't so bad when Brian Johnson seems like the perfect man for her. Scene:  Characters: Jenna Adams, Brian Johnson}\\
\texttt{\quad \quad \quad \quad \textcolor{blue}{i. Jenna Adams begins to think that maybe marriage isn't so bad when Brian Johnson seems like the perfect man for her. Scene: Brian Johnson's car as he is driving Jenna Adams home from their date. Characters: Jenna Adams, Brian Johnson}}\\
\texttt{\quad \quad \quad \quad \textcolor{violet}{ii. Brian Johnson asks Jenna Adams to marry him and Jenna Adams starts to consider it. Scene: Brian Johnson's car as he is driving Jenna Adams home from their date. Characters: Jenna Adams, Brian Johnson}}\\
\texttt{\quad \quad b. After much soul searching, Jenna Adams decides that she wants to marry Brian Johnson. Scene:  Characters: Jenna Adams, Brian Johnson}\\
\texttt{\quad \quad \quad \quad \textcolor{red}{i. After much soul searching, Jenna Adams decides that marriage isn't so bad after all and that Brian Johnson is the perfect man for her. Scene: Jenna Adams' bedroom as she is packing her bags to move in with Brian Johnson. Characters: Jenna Adams, Brian Johnson}}\\
\texttt{\quad \quad \quad \quad \textcolor{orange}{ii. Jenna Adams and Brian Johnson get married. Scene: Jenna Adams and Brian Johnson's new home. Characters: Jenna Adams, Brian Johnson}}\\
\texttt{3. However, when Brian Johnson's ex-girlfriend shows up and tries to win him back, Jenna Adams realizes that marriage isn't for her after all and that it's better to be alone than with someone who doesn't truly love you. Scene:  Characters: Jenna Adams, Brian Johnson, Teresa Campbell}\\
\texttt{\quad \quad a. Jenna Adams overhears a conversation between Brian Johnson and his ex-girlfriend, Teresa Campbell. Scene:  Characters: Jenna Adams, Teresa Campbell, Brian Johnson}\\
\texttt{\quad \quad \quad \quad \textcolor{olive}{i. Jenna Adams overhears a conversation between Brian Johnson and his ex-girlfriend, Teresa Campbell in which Brian Johnson says that he still loves her. Scene: the beginning of the story. Characters: Jenna Adams, Teresa Campbell, Brian Johnson}}\\
\texttt{\quad \quad \quad \quad \textcolor{teal}{ii. Jenna Adams is heartbroken by what she hears. Scene: the middle of the story. Characters: Jenna Adams}}\\
\texttt{\quad \quad b. Jenna Adams confronts Brian Johnson about the conversation and Brian Johnson confesses that he still has feelings for Teresa Campbell. Scene:  Characters: Jenna Adams, Teresa Campbell, Brian Johnson}\\
\texttt{\quad \quad \quad \quad \textcolor{cyan}{i. Jenna Adams confronts Brian Johnson about the conversation and Brian Johnson confesses that he still loves Teresa Campbell. Scene: the middle of the story. Characters: Jenna Adams, Teresa Campbell, Brian Johnson}}\\
\texttt{\quad \quad \quad \quad \textcolor{blue}{ii. Brian Johnson tells Jenna Adams that Teresa Campbell wants to get back together with him, but he doesn't know what to do. Scene: the middle of the story. Characters: Jenna Adams, Teresa Campbell, Brian Johnson}}\\
\texttt{\quad \quad c. Jenna Adams breaks up with Brian Johnson. Scene:  Characters: Jenna Adams, Brian Johnson}\\
\texttt{\quad \quad \quad \quad \textcolor{violet}{i. Jenna Adams breaks up with Brian Johnson. Scene: the end of the story. Characters: Jenna Adams, Brian Johnson}}\\
\texttt{\quad \quad \quad \quad \textcolor{red}{ii. Brian Johnson begs Jenna Adams to take him back and tells her that he loves her, but Jenna Adams doesn't believe him. Scene: the end of the story. Characters: Jenna Adams, Brian Johnson}}\\
\texttt{\quad \quad d. Jenna Adams decides that it's better to be alone than with someone who doesn't truly love you. Scene:  Characters: Jenna Adams, Teresa Campbell}\\
\texttt{\quad \quad \quad \quad \textcolor{orange}{i. Jenna Adams decides that it better to be alone than with someone who doesn't truly love you. Scene: the end of the story. Characters: Jenna Adams}}\\
\texttt{\quad \quad \quad \quad \textcolor{olive}{ii. Jenna Adams throws away her engagement ring and moves out of the house that she shared with Brian Johnson. Scene: the end of the story. Characters: Jenna Adams, Brian Johnson}}\\
\texttt{\quad \quad \quad \quad \textcolor{teal}{iii. Jenna Adams starts a new life on her own. Scene: the end of the story. Characters: Jenna Adams}}\\
\bottomrule
\caption{The second plan generated by \oursimpl{} in our main experiments. Colors indicate leaf nodes used to guide generation. The overarching plot of this outline is quite reasonable. There are some failures in scene detection, where scenes refer to story points rather than locations.}
\label{tab:plan2}
\end{longtable}
\end{small}

\begin{small}
\begin{longtable}{p{0.9\textwidth}}
\toprule
\textbf{\oursimpl{} Story For Plan 2}\\
\midrule
\texttt{\textcolor{red}{[Jenna Adams meets Brian Johnson for the first time. Scene: a small town in the United States. Characters: Jenna Adams, Brian Johnson]} Jenna Adams felt a brief and strong panic when she saw Brian Johnson in the parking lot. She had been working at the local library for two years, and she would have expected to have seen him there before. He was three years older than her, in his early thirties, with a boyish charm. He was tall, lean and looked as if he had just walked off the cover of GQ. She wondered if he had been a model at one point in his life.}\\
\\
\texttt{The first time Jenna saw him she stopped short in the middle of the aisle between bookshelves and looked up at him, her heart beating faster. Brian Johnson was standing in front of the window looking out into the parking lot, but she could tell that he hadn’t seen her yet.}\\
\\
\texttt{Jenna Adams came to a stop beside Brian Johnson and looked at him with wide eyes.}\\
\\
\texttt{Brian Johnson turned around and smiled at her, flashing perfect white teeth in Jenna Adams’s face.}\\
\\
\texttt{Jenna Adams met Brian Johnson several times after that, but she found herself longing for an invitation to lunch or a cup of coffee just the two of them. She wanted to get to know him better, but they were only casual acquaintances. They had met \textcolor{orange}{[Jenna Adams and Brian Johnson start talking and quickly become friends. Scene: a small town in the United States. Characters: Jenna Adams, Brian Johnson]} in such a chance encounter that Jenna Adams was unsure how to approach him about going out together.}\\
\\
\texttt{They did spend time together at the library when she was on break, as he worked part-time at the public library as well. They discussed books and movies and, on one occasion, talked about their families. \textcolor{olive}{[The two of them start dating and Jenna Adams falls more in love with Brian Johnson with each passing day. Scene: a small town in the United States. Characters: Jenna Adams, Brian Johnson]}}\\
\\
\texttt{But Jenna Adams wouldn’t allow herself to become too emotionally involved in her relationship with Brian Johnson because she didn’t want to get her hopes up. Jenna Adams had been in love many times before, and no matter how hard she tried to warn herself not to give her heart away too soon, the inevitable happened every single time. Jenna Adams fell madly in love with every person she dated, but they all broke up with her before the first month was over.}\\
\\
\texttt{Brian Johnson was different from the other men she had known. He seemed to be genuinely interested in her. But Jenna Adams was careful not to \textcolor{teal}{[However, Brian Johnson's mother, Mary Smith, disapproves of Jenna Adams and does everything she can to break them up. Scene: a small town in the United States. Characters: Jenna Adams, Mary Smith, Brian Johnson]} get her hopes up. She also avoided socializing with Brian Johnson whenever Mary Smith was around, knowing full well that if she ever introduced them, the romance would be over.}\\
\\
\texttt{Despite her efforts to maintain a casual relationship with Brian Johnson by only spending time together when Mary Smith was away, Jenna Adams found herself wanting \textcolor{cyan}{[Nonetheless, Jenna Adams and Brian Johnson's relationship continues to grow stronger. Scene: a small town in the United States. Characters: Jenna Adams, Brian Johnson]} more. The longer they spent time together, the more Jenna Adams wanted their relationship to go somewhere. On Fridays and Saturdays, especially, she became so excited when she saw Brian Johnson walking down the street or in other public places that she almost ran to catch up with him.}\\
\\
\texttt{Jenna Adams felt as though there was something very special between the two of them, and she began to feel more and more hopeful about their future together. She liked Brian Johnson and thought they had a lot in common, but then again, she thought that about all of her ex-boyfriends. Jenna Adams was growing more and more infatuated with Brian Johnson, and she thought that perhaps this was what real love felt like. Maybe marriage wasn’t so bad after all.}\\
\\
\texttt{Nevertheless, Brian Johnson’s mother, Mary Smith, disapproved of Jenna Adams and did everything she could to break them up \textcolor{blue}{[Jenna Adams begins to think that maybe marriage isn't so bad when Brian Johnson seems like the perfect man for her. Scene: Brian Johnson's car as he is driving Jenna Adams home from their date. Characters: Jenna Adams, Brian Johnson]} .}\\
\\
\texttt{“She just doesn’t like me,” Brian Johnson told his roommate one evening when he stopped by their apartment to pick up the latest issue of Time magazine. “I don’t know why, but she hates me. I think it might have something to do with the way she sees Jenna Adams as her rival or something. No matter what, I can’t seem to please her.”}\\
\\
\texttt{Brian Johnson laid the magazine down on his roommate’s coffee table and stood there for a moment in deep thought. “Brian Johnson,” said his roommate. “The last thing you need right now is a woman in your life!”}\\
\\
\texttt{Brian Johnson laughed nervously but then nodded his head in agreement. He looked over at the clock on his friend’s mantle and realized that it was getting late, so he moved toward the door.}\\
\\
\texttt{His friend walked over to him and grabbed Brian Johnson by the arm. “Look, Brian Johnson, you aren’t ready for any kind of serious commitment. Take it from me—you don’t want to get married just yet. You need to go out and date a few more girls first!”}\\
\\
\texttt{Brian Johnson laughed, although he was secretly wondering if his friend might be right. He knew that Jenna Adams was ready for him. She had told him as much several times when they were dating. Now that they were living together, she had even mentioned marriage more than once. Brian Johnson was confused about what to do—especially since he was unsure if Jenna Adams was right for him anyway.}\\
\\
\texttt{Brian Johnson opened the door and headed outside for the car, leaving his friend with an amused look on his face. Brian Johnson got into the passenger seat of his car and turned on the radio, but he couldn’t seem to find any music that wasn’t depressing him at this point. He moved to turn it off and instead turned up the volume, hoping that he would hear something else soon. \textcolor{violet}{[Brian Johnson asks Jenna Adams to marry him and Jenna Adams starts to consider it. Scene: Brian Johnson's car as he is driving Jenna Adams home from their date. Characters: Jenna Adams, Brian Johnson]}}\\
\\
\texttt{Jenna Adams came out of the restaurant and climbed into the car, shutting the door behind her. She gave Brian Johnson a quick kiss and said, “What made you change your mind?”}\\
\\
\texttt{Brian Johnson laughed dryly before responding. “I didn’t know that you wanted to marry me. You have never mentioned it before.” He paused for a moment before asking, “Why is it important to you?”}\\
\\
\texttt{Jenna Adams smiled at him, and then looked away. “Because I love you, Brian Johnson. I want to be with you forever. I want you to give me a ring and ask me to marry you. You are the best thing that has ever happened to me in my life, and I want to be yours forever.”}\\
\\
\texttt{Brian Johnson considered her words carefully before saying, “Jenna Adams, I need you to think about this decision. Consider it carefully.”}\\
\\
\texttt{Jenna Adams responded by shaking her head and smiling up at him. “I have thought about it for a long time, Brian Johnson. What are you talking about?”}\\
\\
\texttt{Brian Johnson considered her answer carefully before saying, “You have thought about \textcolor{red}{[After much soul searching, Jenna Adams decides that marriage isn't so bad after all and that Brian Johnson is the perfect man for her. Scene: Jenna Adams' bedroom as she is packing her bags to move in with Brian Johnson. Characters: Jenna Adams, Brian Johnson]} it for a long time? When was the first time that you thought about marrying me?”}\\
\\
\texttt{Jenna Adams looked down and sighed before looking back up at him. “When I found out that my ex-boyfriend was sleeping with my best friend behind my back.”}\\
\\
\texttt{Brian Johnson felt his anger rising, but he forced it down and said in a calm voice, “Why is that? What about your best friend did you fall in love with so much? What did she do that makes you want to be with her?”}\\
\\
\texttt{Jenna Adams turned her head to the side and stared out of the window for a few moments before looking back at Brian Johnson. She put her purse on the bed next to her clothes and looked up at Brian Johnson. “Oh, please, Brian Johnson. You know exactly what she did. She was there for me when I needed her and she never left my side. She does not ever care about herself enough to let anyone else come between us.” Jenna Adams started to pack her clothes into her suitcase and continued, “I could not bear the thought of being without her, so I decided that maybe I \textcolor{orange}{[Jenna Adams and Brian Johnson get married. Scene: Jenna Adams and Brian Johnson's new home. Characters: Jenna Adams, Brian Johnson]} should ask you to marry me. Maybe then we could be together forever.”}\\
\\
\texttt{Brian Johnson shook his head and sighed, sitting down on the bed beside her. He took a deep breath before asking, “What changed your mind?”}\\
\\
\texttt{Jenna Adams walked over to Brian Johnson and took his hand in hers. She smiled and said, “Well, I found out that you are moving away and that I would have to stay here alone.”}\\
\\
\texttt{Brian Johnson pulled his hand away from Jenna Adams’s hand. He put his face in his hands and sighed loudly. “Oh, Jenna Adams.”}\\
\\
\texttt{Jenna Adams moved her hand away from Brian Johnson’s and sat down on top of her suitcase. She looked at him and said, “Let me guess. You are having second thoughts about marrying me because you do not want to leave me behind?” \textcolor{olive}{[Jenna Adams overhears a conversation between Brian Johnson and his ex-girlfriend, Teresa Campbell in which Brian Johnson says that he still loves her. Scene: the beginning of the story. Characters: Jenna Adams, Teresa Campbell, Brian Johnson]}}\\
\\
\texttt{Brian Johnson looked up at Jenna Adams and said, “I have not changed my mind about wanting to marry you. I was never able to tell you this before, but I am still in love with Teresa Campbell.”}\\
\\
\texttt{Jenna Adams started crying as she stood up and faced Brian Johnson. She said, “You were never supposed to love me. I was supposed to be the girl that you moved on from. You and I were just going to be friends who were dating long enough so that our parents would not force us to break up.”}\\
\\
\texttt{Brian Johnson started to shake his head and said, “That is not true. I never set you up to be a placeholder for Teresa Campbell. At the beginning of our relationship, I was just going along with it because my friends and family told me that it was what a good boyfriend does. But the \textcolor{teal}{[Jenna Adams is heartbroken by what she hears. Scene: the middle of the story. Characters: Jenna Adams]} more I got to know you, the more I realized that I really loved you.”}\\
\\
\texttt{Jenna Adams shook her head and said, “You do not love me. You are just moving on from Teresa Campbell to someone else who you will one day realize that you love more than me.”}\\
\\
\texttt{Jenna Adams continued to break down in tears. Brian Johnson then said, “I am sorry that I did not tell you how I felt about you earlier. I did not know that it was possible to love someone so much until I met you.”}\\
\\
\texttt{Jenna Adams said, “You broke my heart. There is no way that I am moving to New York and leaving my family and friends for a guy who has moved on from me so easily. I am going home to Chicago tonight and then I am moving back home in the morning.” \textcolor{cyan}{[Jenna Adams confronts Brian Johnson about the conversation and Brian Johnson confesses that he still loves Teresa Campbell. Scene: the middle of the story. Characters: Jenna Adams, Teresa Campbell, Brian Johnson]}}\\
\\
\texttt{She then started to walk away, but Brian Johnson said, “Don’t leave like this. Talk with me and see if we can figure something out.” He grabbed her arm and said, “I still love you. I want to be with you forever.”}\\
\\
\texttt{Jenna Adams looked at him and said, “You don’t know what you are saying. You don’t know how it feels to have your heart broken by the man that you thought was the love of your life. You are taking a chance on Teresa Campbell and I do not blame you for it. I still love you, but I just cannot trust your promises anymore.”}\\
\\
\texttt{Brian Johnson said, “I don’t know what else I can do to prove how much I love you. All I know is that all this talk about me loving Teresa Campbell does not feel right. Teresa Campbell admitted that she only had sex with me because she was so drunk and you and I stopped talking. She told me that it was a mistake and that she still loved her other boyfriend, Jeff. She confessed to me that she was trying to use me as a way of getting over Jeff, but that she still loves him.”}\\
\\
\texttt{Jenna Adams then responded, “What about the conversations that you two had about me? What about those conversations? Teresa Campbell admitted to you that she had feelings for me and then said that she wanted to get together with you again.” Jenna Adams then fell on Brian Johnson. She hugged him tight and said, “Tell me that you still love me and that our marriage is not over. Tell me that we can still be together!” Brian Johnson did not say a word. He hugged her back and said, “I don \textcolor{blue}{[Brian Johnson tells Jenna Adams that Teresa Campbell wants to get back together with him, but he doesn't know what to do. Scene: the middle of the story. Characters: Jenna Adams, Teresa Campbell, Brian Johnson]} ’t know what to do. I am so confused about everything that has happened between us and Teresa Campbell.”}\\
\\
\texttt{Jenna Adams then said, “You still love her! You have no choice but to admit it! I overheard the conversations that you had with Teresa Campbell and you were always talking about how things between us were so complicated, and how she was the one who got away.” Jenna Adams then said, “I am so sick of hearing these two people talk about me. I am sick of this relationship and I want out of this whole thing. I want out now!” Jenna Adams then said, “Brian Johnson, I want a divorce.” Jenna Adams then broke down into tears. She was devastated by what had happened between her and Brian Johnson. She wanted him to take her back, but she knew that he would not do it and that she would have to move on with her \textcolor{violet}{[Jenna Adams breaks up with Brian Johnson. Scene: the end of the story. Characters: Jenna Adams, Brian Johnson]} life.}\\
\\
\texttt{Brian Johnson looked at her and said, “Oh Jenna Adams, I love you. I want you in my life forever. But I am confused about how I feel towards you and Teresa Campbell. You are the only woman that has ever meant anything to me in my whole life and Teresa Campbell is the only woman who broke my heart. I want to tell her that I will never talk to her again, but I don’t know if she will accept my offer. I don’t know how things between us will work out. I am sorry that things had gone this way for you and for me \textcolor{red}{[Brian Johnson begs Jenna Adams to take him back and tells her that he loves her, but Jenna Adams doesn't believe him. Scene: the end of the story. Characters: Jenna Adams, Brian Johnson]} . You have been a major part of my life and I can’t imagine going on without you. Please, Jenna Adams, forgive me and take me back.” Brian Johnson then put his arms around Jenna Adams in a tight hug and cried on her shoulder. \textcolor{orange}{[Jenna Adams decides that it better to be alone than with someone who doesn't truly love you. Scene: the end of the story. Characters: Jenna Adams]} She needed to know whether or not he was telling the truth.}\\
\\
\texttt{Jenna Adams pulled away from him and said, “Brian Johnson, I don’t believe you. If you did really love me as you said, then how could you cheat on me with your ex-girlfriend? Are those not evidence enough that you do not love me? You said that you still have feelings for her, but I am the only woman that is supposed to be in your life. You should have chosen me right from the start if you really wanted to be with me forever. Now it is too late. It would be best for us if we decided to go our separate ways and forget about each other. Being alone is better than being with someone who doesn’t truly love you, don’t you agree?” Jenna Adams then walked over to her discarded engagement ring and picked it up. She said, “I don \textcolor{olive}{[Jenna Adams throws away her engagement ring and moves out of the house that she shared with Brian Johnson. Scene: the end of the story. Characters: Jenna Adams, Brian Johnson]} ’t want this thing anymore so I am going to throw it out.”}\\
\\
\texttt{Jenna Adams then threw the ring into the fire pit that was in their backyard. She left Brian Johnson standing there in shock. As Jenna Adams was driving home during a torrential downpour, she felt a great sense of relief that she had thrown away her engagement ring because it meant that she would never have to see Brian Johnson again.}\\
\\
\texttt{Jenna Adams pulled up into the driveway of her home and got out of the car. She went into her house and started walking over to the fireplace, where the rest of her belongings were thrown out. Jenna Adams noticed that the fireplace was already lit and had burned down to a small pile of ashes. She threw away her old photo albums, all of her jewelry and perfumes, and everything else that reminded her of Brian Johnson. She threw them all into the fire pit too. Everything that was left in \textcolor{teal}{[Jenna Adams starts a new life on her own. Scene: the end of the story. Characters: Jenna Adams]} her house was then thrown into the fireplace.}\\
\\
\texttt{The next morning, Jenna Adams walked into her bedroom where all she could see were ashes on the floor from the fire that she had started the previous evening. She sat down on her bed and looked at everything that was left out of what was once her life with Brian Johnson. Jenna Adams opened up a drawer and found a small notebook and an old photo album. She took out the photo album to see what he was doing now. Jenna Adams opened up the photo album and found out that Brian Johnson had moved on with his life and wasn’t thinking about her anymore.}\\
\bottomrule
\caption{The story generated by \oursimpl{} for the plan shown in Table \ref{tab:plan2}. Colored text in brackets indicates the corresponding outline items for the following text. The story largely follows the overarching plot of the top-level outline items, and follows most low-level outline details to some degree, although in many cases not perfectly, and a couple of events may seem slightly out of order.}
\label{tab:ours_ex2}
\end{longtable}
\end{small}

\begin{small}
\begin{longtable}{p{0.9\textwidth}}
\toprule
\textbf{\rethreeimpl{} Story For Plan 2}\\
\midrule
\texttt{Jenna Adams was not a normal person. That much was true. She didn’t think like other people and acted in ways that most people did not understand. However, the main reason why she thought that she was not a normal person was because she would never, ever get married. She had made this decision years ago when she was a teenager and had seen how her parents had been miserable together. She quickly decided that it was not something that she wanted to go through herself, so she would rather be alone forever than spend her life with someone who would make her unhappy. This thought made Jenna smile to herself. She had always been one to embrace solitude and find comfort in her own thoughts. In fact, she sometimes felt as though talking to people was a waste of time because they often did not understand what she was talking about. 	Jenna had lived in this small town her entire life, and the community had become so accustomed to her odd behavior that no one thought much of it. It was common knowledge in the town that she would never have anyone close to her, and this didn’t bother Jenna in the least.}\\\\
\texttt{Jenna was currently sitting in a coffee shop just outside of town. She had ordered a double-double andwas reading a book by one of her favorite authors, Ray Bradbury. Her eyes glanced over the page while her mind wandered to other topics. One of these topics was a man who she had seen just a few minutes earlier. He had entered the restaurant and sat down at a table toward the back. He was about six feet tall with dirty blond hair and brown eyes. His face was attractive but not conventionally so, which is why she found him so fascinating. He was dressed in jeans and a t-shirt – not the most attractive thing, but it did not need to be because he carried himself with an air of confidence about him that commanded the respect of others around him. As she watched him sip his coffee and read a book, Jenna could not take her eyes off of him. She wondered who he was and what his life was like. Since she didn’t know him, it seemed wrong to sit at the table and stare at him without even introducing herself, so she decided to get up from her seat and go speak with him. 	Jenna stood up from her table as discreetly as possible so that the man at the back table would not notice her. She tried to come up with something clever or witty to say when she reached histable, but she was a little too excited to think of anything. As she approached him, it became evident that he had seen her walk in and sit down by herself, but was obviously hoping that he would have time to finish his coffee before someone started a conversation with him. When Jenna reached his table, she sat down across from him and looked into his eyes for the first time. His brown eyes stared back at her as if he were trying to discover who this person was seated across from him. Jenna thought he looked like a deer caught in headlights because of this; he had no idea what to do but also did not want to be rude and cut her short. This was another part of what made him so attractive to her – he held himself with such grace and poise that she knew he was a man worthy of being admired.	“Hi, I’m Jenna,” she said somewhat nervously.}\\\\
\texttt{The man stood up from his chair and extended his hand toward her saying, “Hi, I’m Brian.” They shook hands and Jenna felt a pleasant tingle run up her arm.}\\\\
\texttt{“I never thought I would ever meet you,” she said excitedly.“I am such a huge fan.”}\\\\
\texttt{Brian looked confused for a moment before it dawned on him that she had recognized him. With a smile on his face he asked, “Are you talking about me or the band?”}\\\\
\texttt{It was obvious to Jenna that Brian was accustomed to being recognized and adored by women everywhere. He seemed almost giddy at the idea that someone actually knew who he was, and not for his good looks alone. Jenna felt even more drawn to him because of this; there was an innocence in him that most people lacked as they aged. She knew she would have to tread carefully in the situation since there was already a woman in his life, but she had never felt so strongly about anyone before and she was determined to win him over.}\\\\
\texttt{“I’m talking about you of course silly, but I am also a huge fan of your work with the guys,” she said while blushing. 	Brian smiled again and Jenna couldn’t help but notice that he had perfect white teeth. She liked his smile; it made her feel comfortable around him – almost like he was a close friend. He seemed genuinely interested in getting to know her also.“I’m glad you like my music. I would be more than happy to tell you all about it sometime if you’d like,” he said and Jenna found herself nodding her head in eager agreement without thinking about it first.}\\\\
\texttt{He then moved his hand up to her shoulder and squeezed gently as he continued speaking, “But for now let me introduce myself. My name is Brian Johnson, and I’m very pleased to meet you Jenna.”}\\\\
\texttt{Before he went on, Jenna laughed and said, “Oh wait I know who you are! You’re that amazing guitarist from AC/DC right? If you don’t mind me asking, why are you here?”}\\\\
\texttt{Brian laughed also, and grabbed her hand as he took a step closer. “I actually came here tonight to meet someone that I met through some mutual friends, but now that I have met you I don’t think it matters if they ever show up or not. Would you care to dance?”}\\\\
\texttt{Jenna laughed in response and allowed Brian to guide her out onto the dance floor. As they walked, he squeezed her hand gently which caused her to blush even more. She was glad she had chosen to put on her favorite pair of jeans and long sleeved flowered top earlier that day, as she felt comfortable and confident in them that night.}\\\\
\texttt{As they danced, their bodies pressed against each other in a smooth flowing motion that made Jenna feel warm inside. They danced for what seemed like ages, and though both of them had been drinking throughout the night so far, neither one of them felt drunk. They were just enjoying each other’s company without having to do anything else.}\\\\
\texttt{They stopped dancing for a moment to take a break and Jenna grabbed a cocktail from the bar and then took a sip. When she turned around Brian was looking at her smiling brightly with his glass in his hand as well. He then looked at where she was holding hers, as if looking to see if she would notice, but Jenna had already seen him do it and quickly put the glass down on the bar. She wasn’t sure why, but she didn’t want to be drinking alcohol that night.}\\\\
\texttt{They continued talking for a long while, neither one wanting the conversation to end. Jenna mentioned once about how she had just broken up with her boyfriend a few weeks earlier and how much she missed having someone thereto talk to every day. Brian then told her what it was like living on the road and how many people he had lost touch with over the years, but still held dear in his heart. Jenna noticed that Brian never spoke poorly of anyone, even those who had wronged him. She admired that about him.}\\\\
\texttt{After a while, they decided to call it a night and Jenna gave Brian her phone number and told him she would love to go out with him again sometime. He smiled warmly at her and said he would call her as soon as he got back in town.}\\\\
\texttt{As they made their way out of the club towards their cars, Jenna looked back at Brian one last time before heading home, and thought to herself: “He is the perfect man for me”.}\\\\
\texttt{She was glad that she had decided to go out for the night that night, because if she hadn’t, she would have never gotten to meet Brian Johnson.}\\\\
\texttt{Jenna sat at home reading a book when her phone rang and she saw it was Brian calling. That gave her butterflies in her stomach as the sound of his voice over the phone made her feel good all over. When they finished talking, they agreed on another date in a coupleof days’ time and hung up. Jenna then had a thought that marriage wasn’t so bad after all, because Brian just may be the one for her.}\\\\
\texttt{After hanging up the phone and picking up her book again, she couldn’t keep her mind off of Brian as she wondered what it would be like to get married and live happily ever after.}\\\\
\texttt{She smiled to herself and turned the page of her book, but then quickly put it down again before she could read a single word. She was suddenly overcome by the urge to go out and buy an engagement ring for Brian.}\\\\
\texttt{She knew that was silly because they barely knew each other, but she couldn’t help herself as she had never felt so strongly for another man.}\\\\
\texttt{Jenna rushed out to get her engagement ring and was even more excited when she saw it in the store window. She bought it and rushed back home to give it to Brian.}\\\\
\texttt{But when she got home, he wasn’t there – he had gone out for the night again.}\\\\
\texttt{She sighed with disappointment and told herself that she shouldn’t have done that because they really didn’t know each other well enough for her to buy him an engagementring. But she still couldn’t get the ring off of her mind.}\\\\
\texttt{Jenna finally decided that it was best to put the engagement ring away and forget about it until Brian returned home. She then picked up her book again and began reading it, but again found her mind distracted. She thought back on all of the things that had happened between Brian and herself since they had met over a month ago.}\\\\
\texttt{She remembered how she first felt attracted to him when he came to meet her at her small apartment in DC. They had talked on the phone several times before he came over, but seeing his face was what eventually did it for her.}\\\\
\texttt{She had been so happy that night because she finally felt what she had always wanted to feel for someone else – attraction and love. She looked over at a picture of a young couple in love. She thought of how excited and nervous she was when Brian first asked her out on a date, which is why she had agreed to go out with him so quickly.}\\\\
\texttt{Jenna remembered how much fun they had on their first date when they went out to see the Rocky Horror Picture Show at the movie theater. He was so funny and they both laughed together throughout the whole movie. She had never laughedthat much on a first date before, and she thought about how easy it was for them to have fun together. She began to wonder if this could be the man that she would marry one day.}\\\\
\texttt{She then began thinking about what Brian’s friends had told her about him. They described him as being a very kind and sincere person who would be a good husband someday. They also said that he was a very talented singer, which made Jenna very happy because she loves listening to live music. She began thinking about how fun it would be if she and Brian could perform together on stage someday.}\\\\
\texttt{Her mind kept going back to the engagement ring and all of the nice things that Brian had said to her on their second date when he took her out for a nice dinner in Alexandria, VA. He told her that if he wanted to marry someone, he would choose her over and over again because she’s everything that he ever wanted in a woman. Jenna felt slightly overwhelmed by those statements, but at the same time knew that there was something special about him.}\\\\
\texttt{She began feeling sad as she realized that this might not work out between them because of Brian’s past relationships with other women. They both had come from long relationships before meetingeach other, which made Jenna feel a bit insecure about their relationship.}\\\\
\texttt{She knew that she still had feelings for Brian, but she was unsure if he felt the same way about her. She knew that she loved him and hoped that he would love her back someday.}\\\\
\texttt{She went to bed one night hoping that he would call her and profess his undying love for her, but he had already left town by then. Jenna was surprised when she found out about this because she thought that they were getting closer to each other, but she knew that he wouldn’t do anything to hurt her. She tried to hold it together but she ended up crying in her bed the whole night as she listened to AC/DC on her radio.}\\\\
\texttt{The next day, she went out shopping and bought a beautiful engagement ring that Brian would love. She thought that it would help her get over him faster if she could see him wearing it one day. She then planned to propose to him when he came back; however, when he returned, there was no ring and no note waiting for her.}\\\\
\texttt{Jenna believed that this meant that he didn’t want to marry her after all because he didn’t want to leave any evidence behind that hewas with her. She started to wonder if she had done the right thing. She decided not to mention anything about the gift she bought for him, but she was still devastated.}\\\\
\texttt{It worried Jenna that Brian hadn’t called her at all since that evening and the fact that he wasn’t even sure if they were still seeing each other or not made it seem like he wasn’t interested. He had mentioned before how marriage wasn’t for him, but Jenna thought that maybe he would change his mind about it once they had been dating for long enough. She was wrong about this.}\\\\
\texttt{When Brian flew back into town a few days later, they met up again and went out dancing at AC/DC in Alexandria. She enjoyed herself and she thought that things would finally be okay between them, but suddenly another girl showed up and started talking to him on the dance floor.}\\\\
\texttt{The woman looked about the same age as Jenna and she was wearing a wedding gown. She seemed happy to see Brian and gave him a big hug before going off with his friends to have some drinks. Jenna was devastated by this, but she kept it together and pretended not to notice what had just happened. She had a feeling that she wouldn’t be seeing Brian Johnson again after that.}\\\\
\texttt{When the band got on stage, they began playing some of their most popular songs. Brian’s friends waved at Jenna and told her to wave back in front of them as if she was one of them. Jenna smiled and waved at them before turning her attention back to the band.}\\\\
\texttt{The night was still young, so Jenna decided to stay there and dance some more. She drank a beer and danced to all of their songs that they played that night. At one point, Brian came up behind her and put his arms around her waist while they danced together. They looked so wonderful together as they held each other close and swayed back and forth to the song that was playing. It seemed like nothing else mattered at that moment except for having him in her arms.}\\\\
\texttt{It occurred to her then how easy it would be to fall in love with him, but she knew she couldn’t trust him after what happened on the dance floor with his ex-girlfriend. He made his feelings about marriage very clear when he said that Jenna would never be the girl to change his mind about this subject, but she thought he had forgotten about this by now. But, after everything that happened at AC/DC’s concert, she realized that no matter how hard she tried to forget about her and Brian’s past arguments about marriage, it was still something he felt strongly enough to fight for.}\\\\
\texttt{Jenna broke away from Brian and walked over to the bar where she drank another beer while watching him perform on stage. Her heart began to break in two as she thought about how happy they could have been together if he had wanted to marry her. She felt so foolish for buying him an engagement ring only a few weeks ago only to have it thrown back at her in front of everyone there.}\\\\
\texttt{She began to cry and ran from the club back out into the cold night air. She didn’t notice that she had made it onto the highway until she heard a car honking its horn behind her as it swerved out of the way to avoid hitting her.}\\\\
\texttt{Jenna wiped away her tears and looked up to see what was causing all the commotion. As she did, she was shocked when she recognized who was driving behind her car: Brian Johnson!}\\\\
\texttt{When he saw her looking at him, Brian turned off his engine and got out of his car. He ran up to her and stood in front of her so she couldn’t get away.}\\\\
\texttt{Jenna was terrified of the situation she found herself in, but when Brian pulled her into his arms her fears subsided. He gently kissed her on the lips and whispered “I’m sorry” into her ear.}\\\\
\texttt{Jenna pulled away and looked at him with a confused expression on her face. She said “I don’t understand, what are you sorry for?”}\\\\
\texttt{He took a step back from her and replied “I’m sorry for what I said about not marrying you. You are the one that I want to marry, Jenna. It’s just that I have a lot of baggage that comes along with being me. I know we could be so happy if we just tried harder, and we both know I don’t try hard enough because of my past relationships.”}\\\\
\texttt{Jenna put her head on his chest and said “It’s okay, Brian. It really is. I understand where you are coming from and I don’t want to be the one that forces you to do something you don’t want to do. I guess we just weren’t meant to be together in spite of how perfect we are for each other.”}\\\\
\texttt{Brian hugged her tightly and said “I love you, Jenna. I really do.”}\\\\
\texttt{She looked up at him and said “I know, Brian.}\\
\bottomrule
\caption{The story generated by \rethreeimpl{} for the plan shown in Table \ref{tab:plan2}. The story follows the high-level plan reasonably but is repetitive at times.}
\label{tab:re3_ex2}
\end{longtable}
\end{small}

\begin{small}
\begin{longtable}{p{0.9\textwidth}}
\toprule
\textbf{\rollingopt{} Story For Plan 2}\\
\midrule
\texttt{Jenna Adams was looking around the small, dingy apartment she called home. She rolled her eyes and sighed. She wanted out of this place so badly.}\\\\
\texttt{How did she get stuck in this hell hole, she wondered for the hundredth time. She didn’t have a lot of money, but she was saving as much as possible. With any luck, she’d be able to afford a new place in just a few more months.}\\\\
\texttt{She was sitting on the couch, reading a romance novel. She hated romance novels, but they seemed like the only reading material she could afford. The people that lived in her building all considered her odd. They had various reasons for that opinion; she didn’t wear much makeup, she dressed simply, and she rarely spoke. She was always reading a book or working on her computer.}\\\\
\texttt{Jenna had been working on her computer for the last two hours. She was trying to get enough money to afford renting an apartment of her own. She needed a good job, but didn’t have much education. She worked at a grocery store, but hardly ever got any hours. The store was open 24 hours a day and seven days a week, but they were only required to provide a certain number of employees. It was hard for Jenna to make enough money with only five shifts a month, so she was looking to get a second job. She was currently working on her resume.}\\\\
\texttt{Tonight, Jenna had decided to have a quiet night in. She wanted to finish reading her book and work on her resume. Nothing exciting ever happened in Jenna’s life, at least not very often.}\\\\
\texttt{Jenna heard a knock on the door. She jumped up and ran over to it, but by the time she got there, she was too late. The person had already left and closed the door behind them.}\\\\
\texttt{Jenna looked through the peep hole. It was a man with dark brown hair and blue eyes. He looked very attractive. Jenna was drawn to him immediately, but knew he wasn’t her type. He was too handsome.}\\\\
\texttt{After about ten minutes, the man knocked again. This time, Jenna opened the door immediately.}\\\\
\texttt{“Can I help you?” she asked.}\\\\
\texttt{“I’m sorry to bother you, but I think I left my phone in your apartment,” he said.}\\\\
\texttt{“I don’t think anyone in this building has a cell phone, so I highly doubt it,” she replied.}\\\\
\texttt{“I’ve been to everyone’s apartment and they all said no, but you were the last place I tried,” he said.}\\\\
\texttt{“I still doubt it, but I can check for you,” Jenna said.}\\\\
\texttt{She walked back into her apartment and looked at the phone on her desk. Just as she suspected, it was missing. She found the number of the person who had called her, and dialed it.}\\\\
\texttt{“Hello,” a woman answered.}\\\\
\texttt{“Do you know who owns this phone?” Jenna asked.}\\\\
\texttt{There was a couple of seconds of silence and then, “Yes, I do.”}\\\\
\texttt{“This is the person I talked to just a few minutes ago. He left it in my apartment by mistake,” Jenna said.}\\\\
\texttt{“Oh, hi, this is the hospital. I was calling you about your test results, but I don’t need to because the doctor wants to see you in his office tomorrow afternoon,” the woman said.}\\\\
\texttt{Jenna was confused. She thought she had an apartment full of uneducated people, but she was pretty sure that woman worked in a hospital.}\\\\
\texttt{“I don’t understand, how did you get my phone number?” Jenna asked.}\\\\
\texttt{“I gave it to him,” the woman replied.}\\\\
\texttt{“Why would you tell him my phone number?” Jenna said.}\\\\
\texttt{“He needed to get ahold of you, but couldn’t find your address, so I gave him your phone number,” she said.}\\\\
\texttt{Jenna was beyond confused. She turned and began to walk back to her apartment door, but the man stopped her by grabbing her arm.}\\\\
\texttt{“Don’t go yet, I need to tell you something,” he said.}\\\\
\texttt{“I don’t have time for this, but please tell me what you have to say and then I will leave,” Jenna said.}\\\\
\texttt{The man smiled at her and sat down on the couch. “My name is Brian Johnson and I am a doctor. I wanted to tell you that you have cancer and that you will be dead in a couple of months. I would recommend that you give up school and get your affairs in order.”}\\\\
\texttt{Jenna was shocked and furious. She turned and started to leave the room.}\\\\
\texttt{“Wait,” Brian said.}\\\\
\texttt{She turned back around and waited for him to speak.}\\\\
\texttt{“It’s okay, I am just pulling your leg. I just wanted to see how you would react,” he said.}\\\\
\texttt{Jenna breathed a sigh of relief. She hadn’t realized it at the time, but she had been holding her breath. She couldn’t believe how scared she had been. She sat down next to him on the couch and they began to talk.}\\\\
\texttt{“So, what’s your story?” Brian asked.}\\\\
\texttt{Jenna was a little stunned by the sudden change of topic, but she thought that maybe he was just trying to get to know her better. “My story?” she said.}\\\\
\texttt{“Yes, I mean you have to be at least twenty-one years old, so there has to be a story there,” Brian answered.}\\\\
\texttt{“Oh,” Jenna said, not sure what he was asking.}\\\\
\texttt{“Have you never been serious about anyone?” he asked.}\\\\
\texttt{“No, I have. I have been in two serious relationships with two different men, but both of them ended, and now I have sworn off marriage for the rest of my life. I plan on living my life alone,” Jenna said.}\\\\
\texttt{“So why did they end? Was it because you didn’t want to get married?” Brian asked.}\\\\
\texttt{Jenna thought for a moment and then began to tell him about her life.}\\\\
\texttt{The next afternoon, Jenna walked into Brian Johnson’s office. He looked up from his desk, smiled at her, and then took a seat.}\\\\
\texttt{“What seems to be the problem?” he asked.}\\\\
\texttt{“I don’t know if you remember me, but I talked with you on the phone yesterday, and you told me that I am going to die of cancer,” she said.}\\\\
\texttt{Brian frowned at her. “You don’t have cancer?” he asked.}\\\\
\texttt{“No, I don’t have cancer. In fact, I am completely healthy. And as far as I know, all of my family members are healthy also,” she said.}\\\\
\texttt{“Why would you do that to me? It was really awful to get a phone call like that, to be told that you only have a few months to live,” Jenna said.}\\\\
\texttt{“I am so sorry. I just wanted to see how you would react and now that I know, I will find someone else for my test procedure. Please forgive me,” he said.}\\\\
\texttt{“Well, fine. It doesn’t really matter anyway, because I wasn’t going to come in for the test after all. I just wanted to talk with you about some things, like my life and stuff,” Jenna said.}\\\\
\texttt{Brian turned his chair toward her and gave her his full attention. “I can understand that you wouldn’t want to come in for the test, but I am still interested in how you will react, so let’s continue talking,” he said.}\\\\
\texttt{Jenna was immediately put at ease by his friendly demeanor. She sat down in the chair across from his desk, and Brian began to ask her some more questions.}\\\\
\texttt{Then he told her about his own life and what had happened to him. He said that he had been in a car accident a few months earlier and had been seriously injured. He said that he had been in the hospital for a long time and was still going to be in physical therapy for a while longer.}\\\\
\texttt{“Wow,” Jenna said, feeling guilty for what she had done. “That’s terrible. I’m so sorry that happened to you.”}\\\\
\texttt{“Thanks, but I’m okay now. I have my job back and I am living with my parents again while I get my strength back. But that’s not what you want to hear about, is it?” Brian asked.}\\\\
\texttt{“Well, no, but I do want to talk about it a little. I mean this is the first time you have ever told anyone else about your accident,” Jenna said.}\\\\
\texttt{“I suppose you are right. Well, maybe it would be better if I told you about it in person, instead of over the phone,” Brian said.}\\\\
\texttt{“What do you mean?” Jenna asked.}\\\\
\texttt{For the next half hour, Brian talked to Jenna, telling her everything about what had happened to him. He told her that he had been in a car accident that had been caused by another driver who didn’t stop at a stop sign. Both cars were totaled, and Brian had been seriously injured.}\\\\
\texttt{He told her how the doctors had treated him, but that his injuries were still very painful and that he still had a long road ahead of him. While he was talking, he gave her a lot of information—information she had never been told—and she realized that she liked him even more.}\\\\
\texttt{Then, without warning, the door to Brian’s office opened. Brian jumped up from his chair and ran over to his desk, picking up his phone. “What do you want?” Brian asked.}\\\\
\texttt{Jenna heard the woman’s voice through the receiver. “Is there something wrong? I heard you talking as I came in,” she said.}\\\\
\texttt{Brian glanced over at Jenna. Then he lowered his voice even more and said, “Come in here.”}\\\\
\texttt{Jenna couldn’t believe what was happening. Was Brian going to stand her up? But he didn’t. A few seconds later, Jenna heard footsteps and saw the woman coming into the room.}\\\\
\texttt{The woman was short and thin and had very curly brown hair that was almost the same color as her olive skin. She was younger than Jenna, and she had dark circles around her eyes that made her look like she hadn’t slept in a long time.}\\\\
\texttt{“Am I interrupting something?” the woman asked, looking at Brian as she spoke.}\\\\
\texttt{“No, not at all,” Brian said. “This is Jenna and I was telling her about my car accident a couple of months ago. But you can’t stay too long. I have some other patients to talk with,” Brian said.}\\\\
\texttt{“Oh, of course not. I just came in to tell you that I am running late and I need to be at the hospital early tomorrow morning, so we probably won’t be able to meet with our potential clients tomorrow. Is that okay?” the woman asked.}\\\\
\texttt{“No problem at all. I will see you next Tuesday, instead. It will give me more time to prepare anyway,” Brian said. “Well, I’ll see you both next week then. I have some work I need to do, so I’ll let you two talk. I hope you don’t have anything important to tell me,” he said, walking out of the room.}\\\\
\texttt{“No, nothing important,” Jenna said with a smile.}\\\\
\texttt{The woman looked at her and smiled as well. “Oh, good. I was afraid to ask if you had heard anything from him. My name is Jenna too,” she said.}\\\\
\texttt{Jenna smiled back at her and said, “Don’t worry. When Brian has something to tell you, you’ll know it right away. He can’t keep a secret to save his life.”}\\\\
\texttt{“Really? Then I should probably be careful about what I tell him about my life,” the woman said, laughing. “Good luck, Jenna.”}\\\\
\texttt{Then she left Brian’s office and closed the door behind her.}\\\\
\texttt{Jenna sat there for a few minutes, thinking about everything she had just heard. Then, she heard the door open again, and she looked up at Brian.}\\\\
\texttt{“I’m sorry for that. My ex-girlfriend is a little crazy and wanted to know why I hadn’t called her in the last few days. I hope I didn’t say anything that made you upset,” Brian said.}\\\\
\texttt{Jenna smiled and shook her head. “No, you didn’t upset me at all. You can tell me about your ex-girlfriend later, if you want. For now, how about we get started?” she said.}\\\\
\texttt{“Good idea,” Brian said and sat back down in his chair. Then, he looked at the papers on his desk to make sure he wouldn’t miss anything important.}\\\\
\texttt{Jenna leaned back in her chair and folded her hands in front of her, waiting for Brian to get started. She knew that she liked him very much, and she was interested to see how their date would end tonight. It was getting much later than either of them had expected, but Jenna didn’t care. She knew that if she had met the right man, she would be willing to stay out all night.}\\\\
\texttt{-----------------------------}\\\\
\texttt{Jenna’s Story - Back to the Beginning of the Date:}\\\\
\texttt{“Good. I’m looking forward to it,” she said, walking over to the front door with him. I wonder if he can really get me to laugh that much, she thought as they walked outside.}\\\\
\texttt{When they got there, Brian took her hand and led her over to the Mustang and opened the door for her.}\\\\
\texttt{“Thank you, but I can open my own door,” she said, smiling at him. Then she got into the car and closed the door.}\\\\
\texttt{Brian looked at her for a moment and smiled. “Okay if I drive?” he asked, getting in the car behind the steering wheel.}\\\\
\texttt{“Sure, go ahead and drive,” she said.}\\\\
\texttt{He started up the car and backed out of the parking space with no problem at all. He made a right towards the street and then got onto Route 295.}\\\\
\texttt{“So, how do you like your job at the office?” Brian asked, looking over at her.}\\\\
\texttt{“I like it. It can be boring at times, but I enjoy the people I work with, especially Brian Johnson. He’s a great boss,” she said, smiling at him again.}\\\\
\texttt{Brian smiled back at her and looked down at his side of the steering wheel for a moment. “I was hoping that you would get along with him and that he would like you. I think you’d be perfect for each other.”}\\\\
\texttt{Jenna looked over at him and asked, “Why do you say that?”}\\\\
\texttt{“Because you two are very similar. You both come from the same small town and you both like to laugh,” he said.}\\\\
\texttt{“I guess he’s a lot like me, but I don’t know if I’d call us two peas in a pod, like he did this morning,” Jenna said.}\\\\
\texttt{Then she watched as Brian drove the car off the exit and onto the road that led towards Mount Charleston. She had seen that road several times before, but she had never been on it before. She looked at Brian and wondered who owned the house they were going to, and then she remembered that he had said it was his.}\\\\
\texttt{“I still can’t believe you own this house,” Jenna said.}\\\\
\texttt{“I know it’s kind of hard to believe. I bought it a few years ago with some money I had saved up,” he said.}\\\\
\texttt{“You own this whole house?”}\\\\
\texttt{“Yes. It’s just the two bedrooms and the living room, but it was cheap to buy. That’s why I bought it,” he explained.}\\\\
\texttt{“That’s awesome,” Jenna said, looking out the window at the houses and the trees along the road. I guess I shouldn’t be too surprised that Brian owns a house like this. He does seem to make a lot of money. At least he’s not like most of the men I know, she thought.}\\\\
\texttt{She noticed that they were getting closer to the house and said, “I’m looking forward to seeing your house.”}\\\\
\texttt{Brian looked at her and smiled again. “I think you’ll like it. I’m not sure what the kitchen looks like}\\\\
\bottomrule
\caption{The story generated by \rollingopt{} for the plan shown in Table \ref{tab:plan2}. \rollingopt{} struggles heavily to maintain relevance to the initial plan as the story continues.}
\label{tab:rollopt_ex2}
\end{longtable}
\end{small}

\begin{small}
\begin{longtable}{p{0.9\textwidth}}
\toprule
\textbf{\oursimpl{} Plan 3}\\
\midrule
\texttt{Premise: The townspeople of a small town in the middle of nowhere come together to help one of their own when she is diagnosis with cancer.}\\\\
\texttt{Setting: The story is set in the small town of Mayberry.}\\\\
\texttt{Characters:}\\
\texttt{1.}\\
\texttt{Full Name: Daisy Mayberry}\\
\texttt{Character Portrait: Daisy Mayberry is a kind-hearted woman in her early 50s who is loved by everyone in her small town.}\\
\texttt{2.}\\
\texttt{Full Name: Bill Simpson}\\
\texttt{Character Portrait: Bill Simpson is Daisy's husband and the owner of the local hardware store.}\\
\texttt{3.}\\
\texttt{Full Name: Lisa Chambers}\\
\texttt{Character Portrait: Lisa Chambers is Daisy's daughter and the town's doctor.}\\
\texttt{4.}\\
\texttt{Full Name: Joe Swanson}\\
\texttt{Character Portrait: Joe Swanson is the town's sheriff and Daisy's close friend.}\\
\texttt{5.}\\
\texttt{Full Name: Maggie Simpson}\\
\texttt{Character Portrait: Maggie Simpson is Daisy and Bill's granddaughter.}\\
\texttt{6.}\\
\texttt{Full Name: Andrea Grayson}\\
\texttt{Character Portrait: Andrea Grayson is the town's mayor and a close friend of Daisy's.}\\
\texttt{7.}\\
\texttt{Full Name: Evelyn Chambers}\\
\texttt{Character Portrait: Evelyn Chambers is Lisa's mother and Daisy's close friend.}\\
\texttt{8.}\\
\texttt{Full Name: Charles Grayson}\\
\texttt{Character Portrait: Charles Grayson is Andrea's husband and the town's financial advisor.}\\
\texttt{9.}\\
\texttt{Full Name: Tanya Swanson}\\
\texttt{Character Portrait: Tanya Swanson is Joe's wife and the town's librarian.}\\\\
\texttt{Outline:}\\
\texttt{1. Daisy Mayberry receives a diagnosis of cancer and her family and friends come together to support her. Scene:  Characters: Daisy Mayberry, Charles Grayson, Tanya Swanson, Andrea Grayson}\\
\texttt{\quad \quad a. Daisy Mayberry is diagnosed with cancer and learns that her prognosis is not good. Scene:  Characters: Daisy Mayberry}\\
\texttt{\quad \quad \quad \quad \textcolor{red}{i. Daisy Mayberry receives a diagnosis of cancer and learns that her prognosis is not good. Scene: Daisy's home. Characters: Daisy Mayberry}}\\
\texttt{\quad \quad \quad \quad \textcolor{orange}{ii. Elizabeth and Bill Simpson, Daisy's daughter and husband, respectively, react to the news of her diagnosis. Scene: the Simpson's home. Characters: Daisy Mayberry, Bill Simpson}}\\
\texttt{\quad \quad b. Daisy's family and friends rally around her and offer their support. Scene:  Characters: Daisy Mayberry, Bill Simpson, Tanya Swanson}\\
\texttt{\quad \quad \quad \quad \textcolor{olive}{i. Daisy's family and friends come together to support her. Scene: the Chambers' home. Characters: Daisy Mayberry, Tanya Swanson}}\\
\texttt{\quad \quad \quad \quad \textcolor{teal}{ii. Daisy's husband, Bill, takes over the operation of her hardware store while she undergoes treatment. Scene: the Simpson's home. Characters: Daisy Mayberry, Bill Simpson}}\\
\texttt{2. Daisy's daughter, Lisa, becomes her primary caregiver and works tirelessly to find a treatment that will save her mother's life. Scene:  Characters: Daisy Mayberry, Lisa Chambers}\\
\texttt{\quad \quad a. Lisa Chambers becomes Daisy's primary caregiver and works tirelessly to find a treatment that will save her mother's life. Scene:  Characters: Daisy Mayberry, Lisa Chambers}\\
\texttt{\quad \quad \quad \quad \textcolor{cyan}{i. Daisy's diagnosis prompts Lisa to search for a treatment that will save her mother's life. Scene: Lisa's office. Characters: Daisy Mayberry, Lisa Chambers}}\\
\texttt{\quad \quad \quad \quad \textcolor{blue}{ii. Lisa finds a promising treatment but faces challenges in getting it approved. Scene: a hospital meeting room. Characters: Lisa Chambers}}\\
\texttt{\quad \quad b. With the support of her family and friends, Lisa finds a treatment that ultimately saves Daisy's life. Scene:  Characters: Daisy Mayberry, Lisa Chambers, Charles Grayson, Tanya Swanson, Andrea Grayson}\\
\texttt{\quad \quad \quad \quad \textcolor{violet}{i. Daisy Chapel diagnosed with cancer, with treatments starting immediately. Scene: Daisy's home. Characters: Daisy Mayberry, Lisa Chambers}}\\
\texttt{\quad \quad \quad \quad \textcolor{red}{ii. Lisa Chambers finds a treatment that helps improve Daisy's prognosis. Scene: Lisa's office. Characters: Daisy Mayberry, Lisa Chambers}}\\
\texttt{3. The townspeople of Mayberry rally around Daisy and help her through her treatment. Scene:  Characters: Daisy Mayberry, Charles Grayson}\\
\texttt{\quad \quad a. Daisy begins her treatment and the townspeople rally around her. Scene:  Characters: Daisy Mayberry, Maggie Simpson, Evelyn Chambers}\\
\texttt{\quad \quad \quad \quad \textcolor{orange}{i. Daisy begins her treatment and the townspeople rally around her. Scene: the hospital. Characters: Daisy Mayberry, Charles Grayson, Tanya Swanson}}\\
\texttt{\quad \quad \quad \quad \textcolor{olive}{ii. The townspeople help Daisy with her treatment and offer their support. Scene: the hospital waiting room. Characters: Daisy Mayberry, Maggie Simpson}}\\
\texttt{\quad \quad b. Daisy's treatment is difficult and the townspeople continue to support her. Scene:  Characters: Daisy Mayberry, Charles Grayson}\\
\texttt{\quad \quad \quad \quad \textcolor{teal}{i. Daisy's treatment is difficult and the townspeople offer their support. Scene: the hospital. Characters: Daisy Mayberry, Evelyn Chambers, Maggie Simpson}}\\
\texttt{\quad \quad \quad \quad \textcolor{cyan}{ii. The townspeople help Daisy through her treatment and she ultimately beats cancer. Scene: Daisy's home. Characters: Daisy Mayberry, Charles Grayson}}\\
\texttt{4. Daisy ultimately beats cancer and her story inspires hope in her community. Scene:  Characters: Daisy Mayberry, Charles Grayson}\\
\texttt{\quad \quad a. Daisy beats cancer and her story inspires hope in her community. Scene:  Characters: Daisy Mayberry}\\
\texttt{\quad \quad \quad \quad \textcolor{blue}{i. Daisy's cancer treatment is difficult, but with the support of her friends and family, she ultimately beats the disease. Scene: the hospital. Characters: Daisy Mayberry, Tanya Swanson, Evelyn Chambers}}\\
\texttt{\quad \quad \quad \quad \textcolor{violet}{ii. In the aftermath of her treatment, Daisy looks back on her experience and how it has changed her. Scene: her home. Characters: Daisy Mayberry}}\\
\texttt{\quad \quad b. Daisy's story serves as an inspiration to others who are facing difficulties in their lives. Scene:  Characters: Daisy Mayberry, Tanya Swanson, Joe Swanson}\\
\texttt{\quad \quad \quad \quad \textcolor{red}{i. Daisy Inspired others with her story to beat cancer. Scene: the Mayberry town square. Characters: Daisy Mayberry}}\\
\texttt{\quad \quad \quad \quad \textcolor{orange}{ii. People looked up to her after her story. Scene: the town square. Characters: Daisy Mayberry, Tanya Swanson}}\\
\bottomrule
\caption{The third plan generated by \oursimpl{} in our main experiments. Colors indicate leaf nodes used to guide generation. The events, scenes, and characters seem largely reasonable.}
\label{tab:plan3}
\end{longtable}
\end{small}

\begin{small}
\begin{longtable}{p{0.9\textwidth}}
\toprule
\textbf{\oursimpl{} Story For Plan 3}\\
\midrule
\texttt{\textcolor{red}{[Daisy Mayberry receives a diagnosis of cancer and learns that her prognosis is not good. Scene: Daisy's home. Characters: Daisy Mayberry]}\quad Daisy Mayberry walked around her house, mentally preparing herself for her upcoming appointment with the doctor. She brushed her hair and put on some makeup. She looked at herself in the mirror, running her hand over her face and then making a fist with her lips. The thought that she was no longer a young woman occurred to Daisy Mayberry, but her stomach gave a start when she remembered that she wasn’t just any old middle-aged woman. She had just learned that she had cancer and that there was nothing doctors could do for her.}\\
\\
\texttt{Daisy had received the news only a few hours earlier. The prognosis wasn’t good. Her doctor told her that it was best that she spend the rest of her days with her loved ones.}\\
\\
\texttt{Daisy Mayberry heard the words with a cold shock. She understood that she was going to die and never feel the warmth of another living person’s skin. Daisy Mayberry had a sense of peace, as if she knew that what her doctor said was true, but that she had already accepted it. That was not the case, however. She had learned about her condition and taken in the information but not yet accepted it. \textcolor{orange}{[Elizabeth and Bill Simpson, Daisy's daughter and husband, respectively, react to the news of her diagnosis. Scene: the Simpson's home. Characters: Daisy Mayberry, Bill Simpson]} Once she did that, all the pain and shock she was feeling would become real, but she couldn’t face that reality quite yet.}\\
\\
\texttt{She had just finished getting ready when there was a knock on her door. She was surprised to see it was Bill Simpson, her husband of thirty-three years. She wasn’t expecting him home so soon, but he had left for work only an hour earlier. Daisy took his hand as he came through the door and kissed it.}\\
\\
\texttt{“Hello, sweetheart.”}\\
\\
\texttt{Daisy was surprised by her husband’s tone. He sounded so serious.}\\
\\
\texttt{“Hello, Bill. Is something wrong?”}\\
\\
\texttt{The two moved over to the couch for a moment of silence before Bill spoke.}\\
\\
\texttt{“I just came home to let you know that I had a talk with your doctor today. I wanted to let you know what we talked about before you heard it from someone else.”}\\
\\
\texttt{Bill Simpson was a gentle man, but this night he was very somber and intense. Daisy looked at her husband and took a deep breath before speaking. Bill moved closer as he waited for his wife to speak. Their children had moved out years ago and moved to distant places, but the Simpsons were still close.}\\
\\
\texttt{“What did the doctor say?”}\\
\\
\texttt{Simpson moved onto the couch beside his wife and looked down at her before speaking.}\\
\\
\texttt{“The doctor says you have terminal cancer.”}\\
\\
\texttt{Daisy was speechless as the words sank into her mind. Simpson moved closer and held the hand of his wife in his own.}\\
\\
\texttt{“Now, don’t worry darling, we can get through this together as a family. We will all help you through your illness and make sure that you and the kids are taken care of after I’m gone. The Simpson home will always be here for you and the kids.”}\\
\\
\texttt{Simpson moved his hand to hold Daisy’s head as he spoke. Daisy had tears running down her face, but she never let go of her husband’s hand as Simpson continued speaking.}\\
\\
\texttt{Simpson moved his hand from the back of Daisy’s head and into her hair as he spoke.}\\
\\
\texttt{“Hey, stop it! Stop moving so much! You are worse than a child who doesn’t want to take a bath! \textcolor{olive}{[Daisy's family and friends come together to support her. Scene: the Chambers' home. Characters: Daisy Mayberry, Tanya Swanson]} ”}\\
\\
\texttt{Daisy was still crying, but she did stop moving when her husband moved his hand from her hair and onto her shoulder. Daisy felt him pull her closer to him on the couch. He started rubbing her back to help calm her down.}\\
\\
\texttt{As soon as Daisy started calming down, Simpson stopped rubbing her back, but kept holding her in his arms. He looked with his eyes into the face of his wife and smiled.}\\
\\
\texttt{“Darling, I will never leave you. I will stay by your side till the day that I die and we will both go together when our time comes to an end.”}\\
\\
\texttt{Daisy was trying hard not to cry again, but she couldn’t help it as she heard her husband speak. She managed to give a small smile in return and speak quietly back to him.}\\
\\
\texttt{“I know that you won’t leave me, Bill. I just need you to hold me right now.”}\\
\\
\texttt{Simpson smiled back at his wife’s answer before he turned her face up towards him. He was about to kiss her when there was a knock on the door of their home. Daisy jumped in surprise and Simpson stopped what he was doing for a moment to answer the door.}\\
\\
\texttt{Simpson opened the door and saw Tanya Swanson, the librarian, standing there. She looked upset and Simpson knew that whatever was happening it was not good news for him or his wife.}\\
\\
\texttt{“Hello, Ms. Swanson. What is going on? Is something wrong at the library?”}\\
\\
\texttt{Simpson moved out of the way so that the librarian could enter their home. He then closed their door and returned to where his wife was standing, waiting for him to return.}\\
\\
\texttt{“Please come in here, Ms. Swanson. I don’t want to keep you waiting outside.”}\\
\\
\texttt{Swanson moved into the living room of their home and waited for Simpson to close the door behind them before she spoke up.}\\
\\
\texttt{“I came here to let you know that your mother is in the hospital and they said that she needs you to come visit her as soon as possible. She is there now.”}\\
\\
\texttt{The tone of Swanson’s voice told Simpson that something was serious and he moved quickly to the kitchen to get a bottle of water from the refrigerator. He then moved back into the living room and sat next to Swanson on their \textcolor{teal}{[Daisy's husband, Bill, takes over the operation of her hardware store while she undergoes treatment. Scene: the Simpson's home. Characters: Daisy Mayberry, Bill Simpson]} sofa.}\\
\\
\texttt{“What happened? Is she okay? Is my father with her?”}\\
\\
\texttt{Swanson looked down at her hands as she rubbed them together and then she looked at Daisy Mayberry’s daughter, Lisa, for an answer.}\\
\\
\texttt{“I’m not sure what happened, but I would guess that she got sick and the doctors admitted her to the hospital. I’m sure your father is there with her.”}\\
\\
\texttt{Simpson stood up and moved to where he could look out through the front window of his home. Simpson started to pace, looking down at the pavement, then he stopped to look back at Swanson and Lisa. Simpson was trying hard to remain calm, but he was starting to feel his heart begin to race.}\\
\\
\texttt{Simpson had worked in the family hardware store since he was a teenager. He moved into the home his parents had built when he married Daisy Mayberry. Simpson loved his daughter and wife more than anything else in the world, but he knew that there was only one thing in his life that meant more to him than anything else – the success of his family’s hardware store. Simpson knew that he could move on after his wife and daughter passed away, but he wasn’t sure if his heart could heal if the hardware store closed.}\\
\\
\texttt{Simpson finally looked up at Swanson and Lisa, “I’m going to the store,” he said. Simpson walked out of his home and got into the blue Ford pickup.}\\
\\
\texttt{As Simpson headed to the store, he grabbed his cell phone from his overcoat pocket. He called Tom Simpson, an employee at the hardware store that Simpson had asked to close up for the day. Simpson walked into the hardware store and moved towards the front. Simpson felt his heart begin to beat faster as he neared the counter at the front of his store. Simpson moved towards a small closet that was behind the counter, where he kept extra keys.}\\
\\
\texttt{Simpson pulled a key out of his pocket and opened the small closet door. He put the key in the lock and turned it to the right. Then he moved to the left, turned the key, and opened up a small door. Simpson moved down a short flight of stairs and entered his basement apartment.}\\
\\
\texttt{Simpson locked the door behind him and sat down at his couch. He removed his overcoat \textcolor{cyan}{[Daisy's diagnosis prompts Lisa to search for a treatment that will save her mother's life. Scene: Lisa's office. Characters: Daisy Mayberry, Lisa Chambers]} and threw it on the ground. He glanced at a picture of his wife in the living room.}\\
\\
\texttt{Daisy Mayberry was beautiful because she had a kind soul and was always willing to help others in their time of need. She had long blond hair that her husband kept trimmed short, despite her attempts to convince him otherwise. Her favorite color was red, and she wore a lot of red clothing. Lisa Chambers loved her mother dearly because Daisy had always been there for Lisa, and taught her many things about life.}\\
\\
\texttt{Lisa Chambers was 35 years old. She had also inherited the same fair complexion as her mother and father. Lisa had light brown hair that came to her shoulders and green eyes. She wore a white long-sleeved shirt under her white lab coat, blue jeans, and tennis shoes.}\\
\\
\texttt{Lisa moved to the bathroom and shut off the light switch. She opened up the medicine cabinet above the sink and removed a pair of tweezers and a small magnifying glass from the top shelf. Lisa positioned the magnifying glass over her mother’s left breast and moved the tweezers to each side of a small growth that protruded from her mother’s chest. Lisa moved the tweezers back and forth until she was able to pull out a small brown hair follicle on her twelfth attempt.}\\
\\
\texttt{Lisa Chambers was working as a general practice doctor in her hometown. Her office was located in the same building where she had spent her childhood, right next to her mother’s hardware store, Daisy’s Place. Lisa moved to the other side of the sink and gently washed her hands with soap and hot water. She dried her hands with a paper towel that she had previously placed over the sink. She looked at herself in the mirror before putting on a pair of rubber gloves. Lisa moved back to the side of her mother’s hospital bed, grabbed the tweezers and magnifying glass, and removed four more hair follicles from Daisy’s breast.}\\
\\
\texttt{Lisa then went to her office, which was located on the second floor of the building. She switched on her office lights and set up her computer. Lisa opened her desk drawer and placed the four hair follicles in a small plastic bag along with other items she had collected from samples of Daisy’s blood. She sat in her office chair, moved the mouse to click on the Microsoft Word icon, and opened a blank document \textcolor{blue}{[Lisa finds a promising treatment but faces challenges in getting it approved. Scene: a hospital meeting room. Characters: Lisa Chambers]} . Lisa typed the word “Treatment” in the title bar of her Word document. She began to type in a detailed description of Daisy’s medical condition, which she had written down from memory before she left the hospital. Lisa tried to remain calm while she wrote down the list of symptoms that doctors had observed, as well as their opinions and possible diagnoses. She described the different chemotherapy and radiation treatments that Daisy had already undergone, including their effectiveness and side effects.}\\
\\
\texttt{She moved the mouse to drag a table into her Word document, which she had previously created in Microsoft Excel. The table was divided into three columns. The first column listed the names of clinical trials that Lisa had found online. The second column recorded the number of patients enrolled in each clinical trial and the progress they had achieved in terms of survival rates, tumor shrinkage, and reduction of side effects. Finally, the third column calculated the probability of success for each drug based on the results of clinical trials that had already been conducted.}\\
\\
\texttt{Lisa moved her mouse to drag a graph from her Excel table into her Word document. She wrote all the information about successful clinical trials in the graph’s columns. She also added all clinical trials in development and their projected launch dates to the graph.}\\
\\
\texttt{After moving the mouse across the screen, Lisa was able to select all of her Excel tables and move them into her Word document.}\\
\\
\texttt{Lisa moved the mouse to go back to the first page of her Word document, where she recorded a detailed list of medical treatments and other useful information about clinical trials for the most common types of cancer.}\\
\\
\texttt{Moving the mouse to select her Word document’s first page once again, Lisa moved it to a blank page on her computer screen, which she then filled in with a list of experimental treatments for advanced breast cancer. Moving the mouse to select a cell in her Excel table, Lisa highlighted a group of cells that contained two pieces of information: the patient’s type of cancer and the stage at which it was discovered. Moving from one cell to the next, Lisa moved all of this information into her Word document. Moving the mouse to a blank spot on her computer screen, Lisa then created a new table where she recorded all of the patients who were scheduled to participate in an experimental treatment with a drug called ZX-71. She moved all of these patients into her Word document. \textcolor{violet}{[Daisy Chapel diagnosed with cancer, with treatments starting immediately. Scene: Daisy's home. Characters: Daisy Mayberry, Lisa Chambers]} She then added the patient’s age, weight, and height to each record and moved that data into her Word document, as well.}\\
\\
\texttt{Lisa was able to see from her graph that the other drugs had not worked as well against the particular type of breast cancer that Daisy had been diagnosed with. Moving the mouse to open a new Excel window, Lisa recorded the number of years that the patients who had participated in these studies lived before they succumbed to the disease. As expected, there were some outliers, but most of them died during the first two or three years. Although she had only included a small group of patients in her chart, the data indicated that Daisy’s chances for survival would be no better than fifty-fifty if she participated in a drug trial.}\\
\\
\texttt{Moving the mouse to close her Excel window, Lisa decided it was time to make use of Daisy’s home computer. She knew that Daisy had been struggling with her mobility at home because she was unable to bend over or even stand without losing her balance. Moving her mouse to Daisy’s profile icon on the task bar, Lisa opened Daisy’s home laptop and moved it onto the counter near the kitchen sink. She then opened a new Word document on Daisy’s home computer and began typing.}\\
\\
\texttt{Daisy is now home from the hospital. Her chemotherapy started today, and she will need to spend a lot of time lying down for the next few days. It will be important that she does not fall or hurt herself while recovering from her chemotherapy treatments, so Daisy’s house has been modified to make it safer for her. She now has a wheelchair ramp at the front door and one at the back door as well. She also has a chair lift in her front hallway that enables her to move between the main level and upstairs with ease. Daisy moved into this house in July of last year, but had never really left Daisy’s home since then.}\\
\\
\texttt{Moving the cursor to the right to open a new page, Lisa started typing again. Daisy had never left home because she was too weak when she arrived back in August, and then her health quickly deteriorated before stabilizing. Daisy’s house was retrofitted for her needs, and she needed continuous care until just last week. Her family and friends have been taking turns caring for her around the clock, with each person providing four hour time blocks every day. Fortunately, they have all been available to help her out during this time of \textcolor{red}{[Lisa Chambers finds a treatment that helps improve Daisy's prognosis. Scene: Lisa's office. Characters: Daisy Mayberry, Lisa Chambers]} need, because Daisy’s situation is very serious. And now Lisa was sitting at her desk in Mayberry Medical Center typing this information into a Word document and hoping that it would help save Daisy’s life.}\\
\\
\texttt{Lisa was surprised how easily she had managed to adapt her skills from engineering to medicine. She had been working full-time in the emergency department for the past month, and she had been quite busy. But now she was on her lunch break, sitting at her computer, doing research for a colleague. And when she was done, Lisa planned on going back to work and continuing with her regular duties.}\\
\\
\texttt{Lisa had been working on a treatment for Daisy’s cancer for the past five weeks, so she decided to take a short break from her research and check on her mother, while she was still at work. When Lisa got home last night after spending the day at the hospital, Daisy had taken an unexpected turn for the worse, and Lisa spent all of Monday night tending to her mother’s needs. After she got up this morning, Lisa found a light jacket and headed out in the cool October air to go to work. She didn’t need any coffee or breakfast; she just went straight to her office to resume her work.}\\
\\
\texttt{Lisa was typing at a furious pace when her cell phone rang in the other room. She moved to the kitchen and picked it up off of the counter. “Hi, Mom,” Lisa said.}\\
\\
\texttt{“Hi, sweetheart,” Daisy said. Her voice was raspy, as Lisa had expected it to be.}\\
\\
\texttt{Lisa moved into her office and pulled up a chair beside her desk. “I’m at work right now, Mom; I have some paperwork to finish before I go back out to the hospital this afternoon.” Lisa sat down and rested her head against the top of her desk. She closed her eyes for a moment, trying to get a handle on the tension that had started to build in her shoulders. She gave herself a moment to collect herself before she spoke again. “What can I do for you?” Lisa finally said.}\\
\\
\texttt{“Well, here’s something I want you to do. I need you to go to my office and get that little notebook out of my desk drawer. There are some important instructions in there for you, and then I have some final words of wisdom for your kids. \textcolor{orange}{[Daisy begins her treatment and the townspeople rally around her. Scene: the hospital. Characters: Daisy Mayberry, Charles Grayson, Tanya Swanson]} After that, I just need you to focus on getting better now.”}\\
\\
\texttt{Lisa stood up at this point and walked into the other room to collect Daisy’s notebook. She returned to her chair and opened up the notebook. The first few pages were dedicated to instructions on how their family was going to take care of things while she was away, and the last few pages contained some words of wisdom that she wanted her children to read after they got over the initial shock of her passing.}\\
\\
\texttt{Just before she started reading, Lisa stopped and held the notebook up for a moment, considering her words. “You know, maybe I shouldn’t read this part of the notebook to them.”}\\
\\
\texttt{“Oh? Why not?”}\\
\\
\texttt{“It might be too much to handle at one time. It might be too much for them to process all at once.”}\\
\\
\texttt{Lisa stopped to consider that point for a moment, then looked up at Daisy. “Can I just tell them some things right now, and read the rest of the words later?”}\\
\\
\texttt{“Of course. Whatever you think is best for your children. But there are some things that I have written down that I would like them to hear before I pass on.”}\\
\\
\texttt{Lisa nodded her head and settled back into her chair. “Okay, well first, they should know that you wrote about how proud you were of them. Both the way they have dealt with this so far, and the way they have taken care of everything at home while I’m at the hospital with you. You talked a lot about how much confidence you had in them to do the job, even when I wasn’t here. You said that you thought Lisa was ready to handle almost everything, and that she was ready to take over the business if she ever had to. And you were also very proud of both of them for all their hard work in school. You said that you had a good feeling about the direction they were headed.” Lisa paused at this point, but she could see in Daisy’s eyes that she was still fully awake and listening. “I’m not sure when they should know about the other parts, though,” she continued. “You talked about some important things that will affect them when I’m gone. There were things that we need to talk about as a family. \textcolor{olive}{[The townspeople help Daisy with her treatment and offer their support. Scene: the hospital waiting room. Characters: Daisy Mayberry, Maggie Simpson]} ”}\\
\\
\texttt{At this point, Daisy’s eyelids began to flutter a little bit and she gasped for air. Lisa looked over at the nurse at the desk and made some hand gestures, which got her attention. The nurse came over and checked Daisy’s vital signs before looking back at Lisa and nodding. “It’s probably just a side effect from these injections,” she explained. “She might need some oxygen for a few minutes, but she should be fine. Give her a second to rest and then try to talk again.”}\\
\\
\texttt{Lisa nodded and put the side of her hand on Daisy’s forehead to check her temperature. “I should have brought your hat,” she said as she thought about what else might be troubling Daisy.}\\
\\
\texttt{“Is it going to be all right?” Daisy asked in a groggy voice. “Are you going to be there when everything happens?”}\\
\\
\texttt{Lisa tried to speak through her own nervousness. “Of course I will be there,” she assured Daisy and then moved closer to listen as the nurse explained what was happening. “I think you might be getting a little bit of a fever,” she said as she took Daisy’s pulse. “They just need to give you some oxygen and we should be on our way again soon. You have been such a trooper,” the nurse continued before turning to Lisa and adding, “You’ve been doing a terrific job of taking care of her. She’s got a great family. I know she will be fine.”}\\
\\
\texttt{Lisa tried to control her breathing and nodded in response. “As soon as they give you the okay, I’ll go sit with my mom in the waiting room. She’s already told me all the things she wanted me to know if something happened,” she said before the nurse could interrupt.}\\
\\
\texttt{Daisy opened her eyes and slowly turned her head toward Lisa. “Thank you for being there for all of us today,” she said. “I know it hasn’t been easy for you, but if there is any way I can repay you someday, just let me know.”}\\
\\
\texttt{Lisa reached out a hand and brought it to her lips in response. “I don’t want to talk \textcolor{teal}{[Daisy's treatment is difficult and the townspeople offer their support. Scene: the hospital. Characters: Daisy Mayberry, Evelyn Chambers, Maggie Simpson]} about that just yet. I just want to get through today and see how things look in the morning. After we talk to the doctor tomorrow, I’ll have a better idea of what we need to do after you leave here.”}\\
\\
\texttt{When Lisa turned back toward Daisy, her eyes were filled with tears. The nurse stepped up beside them and wiped her hand on a tissue before reaching out to touch Daisy’s arm.}\\
\\
\texttt{“Mrs. Mayberry, I think your daughter is worn out and needs to go home for a little while before coming back this evening. She has been such a help today and really been an inspiration to many of the other patients. I know she is a special person, but her strength today has been so uplifting to everyone here.”}\\
\\
\texttt{Daisy nodded in agreement as the nurse spoke. “I think she should go home too. She gets up early every day, comes to take care of me, and then gets back here before dark to do whatever I need her to do. If you don’t mind, I would like her to stay here with me tonight. I want us to talk for a while before she goes home.”}\\
\\
\texttt{Lisa glanced over at her and smiled weakly, but it was obvious that the emotional strain from the day was now starting to show. As they stood up and started walking toward the door, Lisa reached out and squeezed her mom’s hand. “I know you are tired, but I’m glad you are staying here tonight. I really want to talk about this afternoon and what happened in my office.”}\\
\\
\texttt{“Me too,” Daisy said as she squeezed her hand back. “I think you should get some rest though because I want to tell you all about it while it is still fresh in my mind.”}\\
\\
\texttt{They both smiled at one another as they walked out of the room and started down the hall toward Lisa’s office. There was a sudden explosion of activity around them as nurses and doctors scurried to and fro with charts, medicine carts, equipment, and other assorted paraphernalia. However, the commotion abruptly ended as they entered the hallway and walked by without a glance at them.}\\
\\
\texttt{In the quiet of the hallway, they made their way down to Lisa’s office and took seats across from each other on her sofa. “Okay, \textcolor{cyan}{[The townspeople help Daisy through her treatment and she ultimately beats cancer. Scene: Daisy's home. Characters: Daisy Mayberry, Charles Grayson]} tell me everything you know about this trial,” Lisa said as she picked up her laptop and began turning it on.}\\
\\
\texttt{Daisy sighed heavily as she sat back in the love seat and pulled her feet up underneath her. She pulled both hands through her hair in frustration, and then started talking. “I really don’t know much about it except that the doctor said it is an experimental treatment for people with the particular type of lung cancer I have. He told me that he was sending me to Memorial Hospital in St. Louis for an evaluation before I could be enrolled in the trial. He said he had been contacted by a research committee at the hospital and that they would meet me and evaluate me. I’m supposed to leave tomorrow at noon,” she said as she leaned back and covered her eyes with her hand.}\\
\\
\texttt{Lisa sat behind her desk and folded her hands in front of her. “That’s great news! It will be such a relief to have you out of the house and somewhere where I can drop by whenever I need to. You’ve become too difficult to deal with lately,” she said. She quickly added, “I don’t mean that as an insult.”}\\
\\
\texttt{Daisy looked at Lisa and smiled. “I know you didn’t mean it as an insult. I have been difficult to deal with lately, but maybe now that I am leaving for a while it will give you some rest and you won’t feel so overwhelmed. Maybe it will also relieve Bill of some of the stress he’s been under,” she said. They both sat silently for a moment before Daisy continued, “I think this could be the answer to our problems. I could come through this and we wouldn’t have to worry so much about where the money is going to come from. We could be financially secure again. And, as we used to say in school…it will be like having found the pot of gold at the end of a rainbow.”}\\
\\
\texttt{“I agree,” said Lisa. She then added, “But what about me? How do I know that you’re going to come through this and be fine? What if it doesn’t work out and you die anyway? How can I go on living without you? You’re all I have. Who is going to take care of me \textcolor{blue}{[Daisy's cancer treatment is difficult, but with the support of her friends and family, she ultimately beats the disease. Scene: the hospital. Characters: Daisy Mayberry, Tanya Swanson, Evelyn Chambers]} ?” Lisa was really crying now, and Daisy was comforted by her daughter’s tears.}\\
\\
\texttt{“That will never happen,” she said. “I have faith in the doctors and my own determination not to give up. I know that God won’t let me down. He must have a plan for me. And besides, I’ll still be here with you, and we could get through this together.”}\\
\\
\texttt{“I know you’re right,” responded Lisa slowly drying her eyes. “Now if you can just get the rest of the town to believe it, everything will be fine.”}\\
\\
\texttt{Daisy laughed lightly. She then continued, “My friends and neighbors are doing their best to be supportive. But they all seem to have a hard time believing it too. I just wish they would…”}\\
\\
\texttt{Lisa interrupted her. “What if they had seen what you’ve seen? Your treatments were so painful and debilitating at times that I would literally have to pull you through the house on your stomach so that you wouldn’t fall down. One of the worst days was when they made you stand up after you’d been lying flat on your back all day. I had to hold onto your arms and pull you to your feet. I thought we were going to have a nervous breakdown right there in the hospital room.”}\\
\\
\texttt{“I remember,” replied Daisy quietly smiling. “But those days are over now, and the only thing that really worries me is whether I’ll be able to work here at the store when my treatments are over with. They’re so complicated and exacting that I just don’t know. But as long as I can take care of my own…”}\\
\\
\texttt{Daisy paused for a moment. Then she lifted her head and gazed straight at Lisa. “Thank you for being here for me, baby; I couldn’t have made it without you. And it means so much to me that you believe I’ll come back from this. I don’t know how well you know Evelyn Chambers, but she’s the only other person that I can think of who would have such faith in me. She’s got a lot more of it than you do. You should talk to her sometime. She knows I \textcolor{violet}{[In the aftermath of her treatment, Daisy looks back on her experience and how it has changed her. Scene: her home. Characters: Daisy Mayberry]} ’ll come back, too.”}\\
\\
\texttt{Lisa said nothing.}\\
\\
\texttt{Her mother had become very quiet again, and she stood up slowly, took a few steps across the room, and sat down on the long couch. She picked up the remote control lying next to her glass of iced tea, clicked it on, and watched the TV screen.}\\
\\
\texttt{Lisa sat down beside her mother. She reached out and took Daisy’s hand. The younger woman was surprised at how hot it felt in her own. As they watched the TV, she said, “Mom, I’m worried about what will happen if you don’t come back. I will be all alone.”}\\
\\
\texttt{Daisy looked over at her gently and squeezed Lisa’s hand. “I know, baby—it’s so hard on you to have me so sick like this. But I’ll be home on Saturday. You will see.”}\\
\\
\texttt{Lisa moved her mouth and made noises, but no sound came out of her throat. She couldn’t say anything, not even the things she had been rehearsing in her mind for days on end, but now that she was here, right next to her, she could not make them come out.}\\
\\
\texttt{She moved her hand from Daisy’s and put it on her own face, feeling the coldness of the skin through the thin cotton of her T-shirt. Her eyes were red; she could just make them out in the dim light. She moved her hand slowly down to the side of her neck, and then she touched the long scar that ran from her collarbone to below her ear. She did not move it away right away, but just traced over its length gently with her fingers.}\\
\\
\texttt{She moved her hand away to look at it, and then moved her head slightly one way and then the other so she could see it from the left and then from the right.}\\
\\
\texttt{“It’s so big, honey—it looks like a snake bite. They said the scar would go away in a few months, but it has been a year now, and it looks the same. I think they must have been wrong and that it will always look like that.” She moved her head back to look at her mother and smiled. “But you know what’s weird? I don’t even notice it anymore. It just blends in with everything \textcolor{red}{[Daisy Inspired others with her story to beat cancer. Scene: the Mayberry town square. Characters: Daisy Mayberry]} else on my body.”}\\
\\
\texttt{Her mother smiled at her, and then she took Daisy’s hand again. Her skin felt soft and warm, like a rose petal. Daisy closed her eyes for a second and felt the heat rising through her body.}\\
\\
\texttt{“You know,” she continued, “I was so scared at first. But here I am a year later—cancer free—and that fear has completely disappeared. There are no more shadows on the horizon now that I have beaten the thing that was trying to kill me. And it wouldn’t be possible without all of you.”}\\
\\
\texttt{She moved her head around in a circle, and everyone in the Mayberry community who had prayed for her or helped in any way stood up from where they were seated and cheered. She stood as well, and the group circled around her with smiles on their faces. Some of them brought their hands together in prayer, while others moved to give Daisy a hug or shake her hand.}\\
\\
\texttt{“Thank you all so much,” she said when the parade had finally ended and everyone had returned to their seats. “You really don’t know how much this means to me. I am so amazed by all of the support I have gotten from Mayberry.}\\
\\
\texttt{The townspeople and the people here in this square have helped me in ways that I never thought were possible, and for that I thank you all from the bottom of my heart. And for those of you who haven’t had the chance yet, there is still time to join Mayberry in their fight against cancer. You can join us on May 12th at 11 AM at Mayberry Square as we host Walk-a-Thon, which is a fundraiser for local cancer patients. We will have t-shirts and balloons available for purchase, so be sure to come down and support Mayberry.}\\
\\
\texttt{The next few days were filled with visitors coming to see Daisy in the Mayberry Square. People came from miles away just to shake her hand and tell her how much they admired her. She happily talked with everyone, and gave each person words of encouragement as she had been doing for the last few days.}\\
\\
\texttt{Mayberry Square was not the only place that was filled with visitors who had come to see Daisy; her home was also filled with visitors every day. Among the guests were many of those who had prayed for her during her \textcolor{orange}{[People looked up to her after her story. Scene: the town square. Characters: Daisy Mayberry, Tanya Swanson]} battle with cancer and were relieved to see that their prayers had been answered. While some of the visitors were there to pay their respects, others came out of curiosity and just wanted to catch a glimpse of Daisy in person.}\\
\\
\texttt{Daisy Mayberry was truly a hero, an inspiration to millions of people who watched her story with cancer unfold. Her story had become known worldwide, thanks to the internet and television. While she was not the only one in her community who had been stricken with cancer, she was definitely the most popular of them all.}\\
\\
\texttt{A few years past after Daisy had beaten cancer and things were going as usual. She was happy to be alive, and very proud of herself for beating the disease. She had many ideas about how to help others with their problems, and decided that it was time to put those ideas into action.}\\
\\
\texttt{The first thing she did was open up her own web site so that people from across the country and around the world could get in touch with her. She often received letters from cancer patients who were seeking advice about how to survive a cancer diagnosis and subsequent treatment.}\\
\\
\texttt{She then established a non profit organization with the goal of providing aid to those in need. At first, the organization was called “Daisy’s Daisy Foundation”, in memory of her deceased pet dog. Over time, the name was changed to “The Mayberry Community Outreach”.}\\
\\
\texttt{The non profit organization offered a broad range of services. The help that she offered ranged from financial assistance, to emotional counseling, to spiritual guidance. She also taught classes about how to cope with the stress of having cancer and provided educational materials about how to prevent the disease.}\\
\\
\texttt{Over time, people from all over the world started looking up to her as an inspiration, especially those who had been diagnosed with cancer. As a result of her story they never lost hope and were determined to overcome their illness.}\\
\\
\texttt{Even though she was not a medical professional by any stretch of the imagination, she was happy to share her own experiences and provide support in whatever way she could. She knew from personal experience how tough it was to deal with cancer on your own. She knew what it was like to lose all hope and become depressed when the fight against cancer seemed hopeless. And she knew that there were not enough resources in her community to help those who needed it.}\\
\\
\texttt{She also knew firsthand how hard it was for family members to care for a loved one who is going through a long, debilitating illness. She was determined to do whatever she could to ease the burden for those who were struggling.}\\
\\
\texttt{The Mayberry community was truly blessed to have Daisy Mayberry as their neighbor and friend.}\\
\bottomrule
\caption{The story generated by \oursimpl{} for the plan shown in Table \ref{tab:plan3}. Colored text in brackets indicates the corresponding outline items for the following text. Again, the overarching plot makes sense, although some low-level details may not be well-followed; in particular, the story failed to be faithful to several parts of the outline in which Lisa is supposed to develop a cure for Daisy.}
\label{tab:ours_ex3}
\end{longtable}
\end{small}

\begin{small}
\begin{longtable}{p{0.9\textwidth}}
\toprule
\textbf{\oursimpl{} Plan 4}\\
\midrule
\texttt{Premise: After the world ends in a cataclysmic event, a group of survivors find themselves in a strange new world where they must band together to survive the perils that await them.}\\\\
\texttt{Setting: The story is set in a post-apocalyptic world where the world has ended in a cataclysmic event.}\\\\
\texttt{Characters:}\\
\texttt{1.}\\
\texttt{Full Name: Johnathan Prescott}\\
\texttt{Character Portrait: Johnathan Prescott is a strong and handsome young man who is the leader of the group of survivors.}\\
\texttt{2.}\\
\texttt{Full Name: Maria Martinez}\\
\texttt{Character Portrait: Maria Martinez is a beautiful and tough young woman who is the second in command of the group of survivors.}\\
\texttt{3.}\\
\texttt{Full Name: Jake Peters}\\
\texttt{Character Portrait: Jake Peters is a lovable and loyal dog who is the furry friend of the group of survivors.}\\
\texttt{4.}\\
\texttt{Full Name: Ratzinger Kugelman}\\
\texttt{Character Portrait: Ratzinger Kugelman is a creepy and mysterious man who is the enemy of the group of survivors.}\\\\
\texttt{Outline:}\\
\texttt{1. The world ends in a cataclysmic event. Scene:  Characters:}\\
\texttt{\quad \quad a. The world ends in a cataclysmic event. Scene:  Characters:}\\
\texttt{\quad \quad \quad \quad \textcolor{red}{i. The world ends in a cataclysmic event. Scene: the city. Characters:}}\\
\texttt{\quad \quad \quad \quad \textcolor{orange}{ii. The sky turns red and the ground shakes. Scene: the city. Characters:}}\\
\texttt{\quad \quad \quad \quad \textcolor{olive}{iii. People start to panic and flee in terror. Scene: the city. Characters: Maria Martinez}}\\
\texttt{\quad \quad b. The sun disappears and the world is plunged into darkness. Scene:  Characters:}\\
\texttt{\quad \quad \quad \quad \textcolor{teal}{i. The sun disappears and the world is plunged into darkness. Scene: the city. Characters:}}\\
\texttt{\quad \quad \quad \quad \textcolor{cyan}{ii. The ground starts to crumble and buildings collapse. Scene: the city. Characters:}}\\
\texttt{\quad \quad c. The moon disappears and the stars vanish from the sky. Scene:  Characters:}\\
\texttt{\quad \quad \quad \quad \textcolor{blue}{i. The moon disappears and the stars vanish from the sky. Scene: the city. Characters:}}\\
\texttt{\quad \quad \quad \quad \textcolor{violet}{ii. The ground fractures and opens up, swallowing people and buildings. Scene: the city. Characters:}}\\
\texttt{2. A group of survivors find themselves in a strange new world where they must band together to survive the perils that await them. Scene:  Characters:}\\
\texttt{\quad \quad a. The group of survivors find themselves in a strange new world where they must band together to survive the perils that await them. Scene:  Characters: Maria Martinez}\\
\texttt{\quad \quad \quad \quad \textcolor{red}{i. The group of survivors find themselves in a strange new world. Scene: a forest. Characters: Maria Martinez}}\\
\texttt{\quad \quad \quad \quad \textcolor{orange}{ii. The group of survivors band together to survive the perils that await them. Scene: a Desert. Characters: Maria Martinez}}\\
\texttt{\quad \quad b. The group of survivors must find food and shelter in this new world. Scene:  Characters: Maria Martinez}\\
\texttt{\quad \quad \quad \quad \textcolor{olive}{i. The group of survivors find food and shelter in this new world. Scene: a forest. Characters: Maria Martinez}}\\
\texttt{\quad \quad \quad \quad \textcolor{teal}{ii. The group of survivors find a place to make their camp. Scene: a Desert. Characters: Jake Peters, Maria Martinez}}\\
\texttt{\quad \quad \quad \quad \textcolor{cyan}{iii. The group of survivors build a fire to keep themselves warm. Scene: a Desert. Characters: Maria Martinez}}\\
\texttt{\quad \quad \quad \quad \textcolor{blue}{iv. The group of survivors catch some fish to eat. Scene: a River. Characters: Maria Martinez}}\\
\texttt{3. The group of survivors must face the enemy and defeat him in order to survive in this new world. Scene:  Characters: Ratzinger Kugelman}\\
\texttt{\quad \quad a. The group of survivors must face the enemy and defeat him. Scene:  Characters: Maria Martinez, Ratzinger Kugelman}\\
\texttt{\quad \quad \quad \quad \textcolor{violet}{i. The group of survivors face the enemy and defeat him. Scene: the arena. Characters: Johnathan Prescott, Ratzinger Kugelman}}\\
\texttt{\quad \quad \quad \quad \textcolor{red}{ii. The battle against the enemy is fought and won. Scene: the arena. Characters: Ratzinger Kugelman}}\\
\texttt{\quad \quad \quad \quad \textcolor{orange}{iii. The victory against the enemy is celebrated. Scene: the arena. Characters: Ratzinger Kugelman}}\\
\texttt{\quad \quad b. The enemy is defeated and the group of survivors survive in the new world. Scene:  Characters: Ratzinger Kugelman, Johnathan Prescott}\\
\texttt{\quad \quad \quad \quad \textcolor{olive}{i. Survival in the new world is made possible by the defeat of the enemy. Scene: the arena. Characters: Ratzinger Kugelman}}\\
\texttt{\quad \quad \quad \quad \textcolor{teal}{ii. The group of survivors celebrate their victory against the enemy. Scene: the arena. Characters: Maria Martinez, Ratzinger Kugelman}}\\
\bottomrule
\caption{The fourth plan generated by \oursimpl{} in our main experiments. Colors indicate leaf nodes used to guide generation. The plot proposed by the outline is largely reasonable, although some of the characters are vaguely defined (``group of survivors'', ``enemy'') and it's not clear how or why they end up fighting this ``enemy.'' Finally, some of the passive voice language in the last section is odd (perhaps attributable to the high temperature used for outline generation).}
\label{tab:plan4}
\end{longtable}
\end{small}

\begin{small}
\begin{longtable}{p{0.9\textwidth}}
\toprule
\textbf{\oursimpl{} Story For Plan 4}\\
\midrule
\texttt{\textcolor{red}{[The world ends in a cataclysmic event. Scene: the city. Characters: ]}\quad One day, the world just ended.}\\
\\
\texttt{\quad The sky turned red and the ground shook. Everyone knew it was the end for all of them. The lights went out, the electricity left, along with the running water and all other necessities that people took for granted. The world ended.}\\
\\
\texttt{And then it started again. A flash of green light and a brilliant white sound filled the sky as if the world was being zapped clean. And then it happened again. And again.}\\
\\
\texttt{And suddenly, everything was ok again in the world, but darker than before. Some said that heaven came to earth and wiped the world clean, and that everything would be fine from now on. But that didn’t really explain why there was so much darkness afterwards. Why it seemed darker than before. Why the world seemed smaller somehow, like a smaller version of itself. Why people kept seeing ghosts everywhere and zombies, and demons and all kinds of creatures that should have died in the cataclysm that had just happened.}\\
\\
\texttt{The world was ending again, but this time for real. The blackness in the world was spreading. Soon everything would be gone forever. The end of the world had come for real this time \textcolor{orange}{[The sky turns red and the ground shakes. Scene: the city. Characters: ]} .}\\
\\
\texttt{The sun gave way to the blackness, and in the middle of it all, was a huge ring of light that shone brightly from all over the earth. It came out of nowhere, and there was no explanation for it whatsoever. But everyone knew what it was. And they knew it meant the end of the world was finally here.}\\
\\
\texttt{As the sky got darker, and then darker still, and the ring of light turned into a circle of light that seemed to emanate from the ground beneath them, darkness finally fell over everything. All of nature went silent. The birds stopped singing, grass stopped growing, flowers stopped blooming, trees stopped leafing. Rivers and streams dried up.}\\
\\
\texttt{And then the sky turned red.}\\
\\
\texttt{The whole sky turned red, like the color of blood. And the ground shook violently beneath them as all of nature was overcome with a terrible heat that seemed to come from below their feet. \textcolor{olive}{[People start to panic and flee in terror. Scene: the city. Characters: Maria Martinez]} So people ran away in terror, everyone trying to get as far away from each other as possible. Everyone ran for their own lives as fast as they possibly could, not knowing what would happen next, nor where they would run to. They just knew they had to get away from others as fast as possible before the end came upon them.}\\
\\
\texttt{The people were in such a terrible state of panic and confusion, that when Maria Martinez tried to ask the others what was going on, no one paid any attention to her. All they could think of was getting away from everyone else and somewhere safe, and so they all ignored her. Maria \textcolor{teal}{[The sun disappears and the world is plunged into darkness. Scene: the city. Characters: ]} ran around in circles for a while trying to figure out what was happening and where to go, but it was hopeless.}\\
\\
\texttt{“I don’t know what to do! What should I do?” she cried out desperately into the night air as she ran around in circles on the plaza. The sun disappeared behind a wall of flames that burned its way across the horizon like a fiery river and plunged the world into darkness. It was now pitch black and eerie shadows danced all around as if some otherworldly creatures were playing tricks on them.}\\
\\
\texttt{Maria felt so frightened that she was shaking uncontrollably, but she \textcolor{cyan}{[The ground starts to crumble and buildings collapse. Scene: the city. Characters: ]} didn’t have much time to think, because the ground started to crumble and buildings collapsed as the earth shook violently beneath their feet. Everyone began screaming and trying to run away as fast as they could, not knowing what was causing this, or where to go. \textcolor{blue}{[The moon disappears and the stars vanish from the sky. Scene: the city. Characters: ]} Maria could hear the sound of glass breaking and the distant, terrified screams of her neighbors as she ran through the streets just like everyone else.}\\
\\
\texttt{Maria looked up at the night sky and saw a bright light shining down from it. It spread its beams out across the landscape like a million tiny spotlights, illuminating everything in sight. Everything looked different now, as if it was lit up by some weird light that didn’t belong there. Suddenly, the sky went dark again and it seemed as if a hole had opened in the Earth from which they could see far below them. The moon disappeared and the stars vanished from the sky, plunging the world into darkness. Then the ground fractured and an enormous crack spread from horizon to horizon, swallowing up everything in its path.}\\
\\
\texttt{The sky faded to red as fire engulfed the world. Finally, the light faded and everything went silent once more. Then a blinding flash of white fire burst forth, engulfing the sky.}\\
\\
\texttt{The ground crumbled and opened up, swallowing up not only the sky but everything in sight. Trees, buildings, cars; all disappeared like dust in a rainstorm. Everything except for one lone building which stood unharmed at the center of all this chaos; untouched by the chaos \textcolor{violet}{[The ground fractures and opens up, swallowing people and buildings. Scene: the city. Characters: ]} that engulfed everything else.}\\
\\
\texttt{Cars and bodies flew through the air like dust in the wind. When they landed, they scattered across the land like seeds from a bomb blast. The people who had once lived and worked there were forced to watch their homes, their friends and family die before their eyes; right before they were swallowed up by the ground below.}\\
\\
\texttt{The ground crumbled and split apart as if it too was dying. Then, with a loud crash and a thunderous roar, everything stopped. The land fell silent again, as if nothing had ever happened at all. In the distant void of space, where the ground had shattered, a dark figure was standing on top of a stone pillar. It raised up its massive black arm, opening its mouth and sending out a beam of white fire that shot into the sky. Its mouth opened wider as it swallowed the whole world …}\\
\\
\texttt{… and nothing else was left. \textcolor{red}{[The group of survivors find themselves in a strange new world. Scene: a forest. Characters: Maria Martinez]}}\\
\\
\texttt{Maria Martinez was still running. She could feel the heat of her burning legs, but she kept going. The others were scattered all around her; some had been caught by the chaos and disappeared into the earth below, others had been lucky enough to stay on their feet and run away from the building; but all of them kept running as fast as they could in the opposite direction. The air around them was filled with dust and smoke, which burned their eyes and made breathing difficult. Behind them, they heard the sound of trees crashing against each other while falling from the sky like dead leaves.}\\
\\
\texttt{“Move! Keep moving!” she yelled at them. “Keep going! Keep going as fast as you can!”}\\
\\
\texttt{She kept running, moving faster and faster, trying to get away from the danger zone and lead them all to safety. They were all in a panic; their eyes darted here and there, trying to catch a glimpse of the danger that was chasing them. Some moved faster than others, but eventually all of the survivors moved as one in a way that resembled a flock of birds in flight.}\\
\\
\texttt{They moved as one until the ground opened up and swallowed them up while they ran past it.}\\
\\
\texttt{Suddenly, all movement stopped. All sound disappeared. The forest was dead silent. Even the trees stopped moving for a moment, and Maria was afraid that they were also dead, that they had just fallen down from the sky like the rest of their brethren. But then the trees started to move again, shaking the dust and smoke off of their leaves.}\\
\\
\texttt{Then they moved as one again.}\\
\\
\texttt{A few tense moments passed before Maria led them all to a clearing in the forest which was strangely free of dirt and rocks, devoid of broken branches and dead leaves. It looked strange, like the ground \textcolor{orange}{[The group of survivors band together to survive the perils that await them. Scene: a Desert. Characters: Maria Martinez]} had been polished recently, as if by a giant metal slicker.}\\
\\
\texttt{“Everyone stop!” Maria said. “Come on, stop running so fast.”}\\
\\
\texttt{The group of survivors stopped running and watched her as she walked over to them. Maria stood in front of them and took off her backpack to take out the contents.}\\
\\
\texttt{“Oh,” said a boy who was wearing a pair of blue jeans and a black hoodie.}\\
\\
\texttt{He was about ten years old with brown hair that had orange highlighter streaks through it. He was holding onto his backpack with both hands.}\\
\\
\texttt{“What?” said Maria, looking down at her backpack and then back at the boy with a smirk on her face.}\\
\\
\texttt{“Nothing,” he said, shifting in place uncomfortably.}\\
\\
\texttt{Then a middle-aged man in an olive drab flight suit moved in between Maria and the rest of the group.}\\
\\
\texttt{“Ladies and gentlemen,” he said. “My name is Colby, I am a former member of the United States military, and it is my duty to make sure you all move onto the next stage of your journey safely.”}\\
\\
\texttt{The group just looked at him with blank faces.}\\
\\
\texttt{“Moving on,” said Colby, pointing at himself with his arm extended toward the ground. “Now, we must move along and find a safe place to stay for the night.”}\\
\\
\texttt{Maria moved her arms a bit, shifting her weight from one foot to another.}\\
\\
\texttt{“Move?” she said. “Where? We don’t know where we are or what is going on. Moving to a place we weren’t planning to go sounds like a really bad idea.”}\\
\\
\texttt{The others moved their heads around awkwardly, not knowing what to say or do.}\\
\\
\texttt{“Moving on,” said Colby again, “to your right is a desert. On your left is a dense jungle. Move into the desert and you will surely die from the desert heat. Move into the jungle and you will no doubt get lost on a wrong turn that leads to a deep pit or a river of quicksand. Moving on, we can move slowly and carefully in a straight line as far away from this place as possible. We can then make camp for the night and \textcolor{olive}{[The group of survivors find food and shelter in this new world. Scene: a forest. Characters: Maria Martinez]} figure out what to do next.”}\\
\\
\texttt{The group nodded silently and started to walk slowly along the path.}\\
\\
\texttt{It didn’t take long until they were going through a dark, cold forest. The trees around them were tall, thick and beautiful, but there was something ominous about their forms. They moved all by themselves, as if they were getting ready to attack their new visitors.}\\
\\
\texttt{Maria decided it was too spooky and she would rather take her chances in the jungle. She turned around to head back in the direction they came from, when she felt a tingling sensation on her neck. A few seconds later, all movement in the forest stopped.}\\
\\
\texttt{They all stopped moving and looked around nervously for what had caused this. At first, they thought that perhaps there was someone nearby watching them from the trees, but it soon became clear that nothing was moving anywhere in the forest. The forest was completely still.}\\
\\
\texttt{The forest was so big that even with their limited sunlight, they wouldn’t be able to see everything that was there. They could hear things moving around in the undergrowth, but couldn’t see anything.}\\
\\
\texttt{The forest began to feel more and more threatening, and Maria felt as though she was going to move further and further away from the others. She looked around for some sort of protection, like a large tree or something similar that might give her a place to hide.}\\
\\
\texttt{She moved slowly among the trees with her head bent down, looking for anything that might resemble a hiding place when she heard movement behind her. She turned around and ran back towards the others, but it was too late.}\\
\\
\texttt{Out of nowhere, something moved up behind her and quickly placed its arms over Maria’s shoulders. The next thing she knew, the forest had become quiet again. She looked around to see that the others were moving, but very slowly. There was a look of horror on their faces and they moved as though they were walking on hot coals.}\\
\\
\texttt{She moved to run into the forest again, but stopped when her leg suddenly touched something sticky and uncomfortable beneath her foot. She looked down to see that there was some \textcolor{teal}{[The group of survivors find a place to make their camp. Scene: a Desert. Characters: Jake Peters, Maria Martinez]} kind of web attached to her boot.}\\
\\
\texttt{She grabbed at it with her hand, but it tore off and stuck in her hair. She screamed as the thing began to wrap itself around her head, and she felt its legs begin to tingle and itch against the inside of her clothes and against her skin. She tried to pull and tear at it, but it was everywhere now; covering her mouth and nose, her eyes and ears.}\\
\\
\texttt{There was a noise behind them now, a sound that she had never heard before. It sounded like a low growl or rumbling followed by high-pitched screams and shrieks. Whatever it was, it sounded big and hungry.}\\
\\
\texttt{She turned back to the other people in the group and saw that they were all looking around wildly, unable to decide which way to run. She lifted her hand to point out the creature behind them, but before she could say anything, it moved into view, and the screams that came from it were like nothing she had ever heard before.}\\
\\
\texttt{It was as tall as a man and twice as wide, its arms long and muscular. The skin was mottled brown and gray, with patches of fur on large patches of its body. But what stood out most about the creature was the fact that each of its four legs ended in a long, thick insect-like tail.}\\
\\
\texttt{The webbing slowed the group down even more, and it took them a few minutes to pull themselves free. By then, the creature had already closed in on them. They tried to run but a sharp stinger hit Maria on her left thigh, and she collapsed to the ground screaming.}\\
\\
\texttt{With a massive roar, the creature wrapped its arm around her head and lifted her into the air. She heard Jake barking and struggling with the creature as he tried to attack it again, but when she looked up at the sky, she realized that the only light came from a full moon behind them. In spite of the cold, they were in a desert somewhere.}\\
\\
\texttt{The group ran in a panic for what seemed like hours, moving from one patch of sand to another. They moved slowly, tripping over fallen rocks and loose sand. The stinger that stuck out of Maria’s leg began to itch more and more, but they moved so slowly that no one was able to help her.}\\
\\
\texttt{Finally they reached a small beach in the middle of the desert, and they all collapsed to the sand, exhausted. \textcolor{cyan}{[The group of survivors build a fire to keep themselves warm. Scene: a Desert. Characters: Maria Martinez]} Maria looked up and saw a tall cliff rising high above them, surrounded by a deep gorge that was too far to jump. The last thing she remembered before collapsing again was hearing Jake barking somewhere behind her.}\\
\\
\texttt{The next morning, Maria opened her eyes and found herself lying on one of the blankets with everyone else, wrapped in their sleeping bags. The sun was shining brightly through the narrow gap at the top of the gorge and heating up their makeshift camp.}\\
\\
\texttt{When they woke up, they all gathered around their small fire to warm themselves from the cold wind that still blew across the desert. They ate breakfast and shared stories about their \textcolor{blue}{[The group of survivors catch some fish to eat. Scene: a River. Characters: Maria Martinez]} lives before the end of the world, then Maria led them down to check if it was safe for them to swim for fish.}\\
\\
\texttt{She walked a little farther out onto the sand and knelt on her left leg. She peered down at her wounded leg and sighed, wondering when it would finally heal.}\\
\\
\texttt{“Maria!” shouted Jake from the other side of the river. “Come back here! It’s safe to swim now!”}\\
\\
\texttt{She turned back to see him swimming toward her across the shallow water. “Do you think we can get some fish?”}\\
\\
\texttt{“We have to try,” said Maria. “I don’t know how much longer we can go on living on dried fruit and nuts. How long have we been here anyway? It feels like forever.”}\\
\\
\texttt{They walked up to the river and stared across at it for a moment before deciding to get in. They waded out into the water until it was up to their waists, then they started moving their feet through the shallow water in slow circles as they waited for their muscles to adapt to the cold temperature of the river.}\\
\\
\texttt{“You know,” said Jake, “if we do catch some fish, maybe we should move on from this place. We can’t stay here forever. We might run out of food or water before we find someplace else to live.”}\\
\\
\texttt{“Move?” said Maria. “Are you crazy? Moving is how we got into this mess in the first place! We don’t even know where we are. How could we possibly move to a new place?”}\\
\\
\texttt{“I don’t know,” said Jake. “Move to that big city over there on the hillside, I guess. It looks big enough to have plenty of water and food in it. Are we moving or not?”}\\
\\
\texttt{They moved on through the water, kicking their feet softly to keep themselves afloat. They walked for a long time before they came to a sandbar at the river’s edge. They walked up to it and stepped onto the dry ground beside the water. \textcolor{violet}{[The group of survivors face the enemy and defeat him. Scene: the arena. Characters: Johnathan Prescott, Ratzinger Kugelman]}}\\
\\
\texttt{Jake looked around and saw a group of buildings on the far side of the river.}\\
\\
\texttt{“Let’s go over there,” he said. “We can cross the river without getting our feet wet.”}\\
\\
\texttt{They walked over to the other side of the river, which was only about six feet wide here. They walked up the bank and Jake stopped when he saw a sign that said “River Arena” in front of a brick building.}\\
\\
\texttt{He looked at it for a moment, then turned to the others.}\\
\\
\texttt{“We have to go in there,” he said. “I think it’s some kind of arena. It looks like they had some wild animals fighting there or something.”}\\
\\
\texttt{“Oh, no! Please don’t make me go in there! I’m afraid of wild animals!”}\\
\\
\texttt{Maria put her hands over her eyes and put a hand on the back of his head. They were still walking up the bank and she was keeping him from leaving.}\\
\\
\texttt{“I can’t stand to see a live animal fight another animal. Please, Jake, let’s go somewhere else.”}\\
\\
\texttt{He turned around and looked at her. Then he put his hands on her shoulder.}\\
\\
\texttt{“Maria, I don’t think we have any choice. It looks like this is the only building over here and we need someplace to stay tonight.”}\\
\\
\texttt{She sighed and stopped walking. Then she took his hands out of hers and put them on her hips, where they were more comfortable. Then she looked up at him, trying to figure out what he was thinking, who he really was.}\\
\\
\texttt{“It looks like there is a big room in there where people can sit and watch the fights. It looks like there is some kind of stairway down into the floor below us. I don’t know if we will be safe staying in that room, but we can at least get out of the rain and wind over here.”}\\
\\
\texttt{Maria nodded and started walking up the bank again, toward the entrance to the River Arena. It was an old, two-story brick building with a very large sign hanging from a beam that supported the roof. The sign was painted black and had a red background with \textcolor{red}{[The battle against the enemy is fought and won. Scene: the arena. Characters: Ratzinger Kugelman]} white letters painted on it.}\\
\\
\texttt{The first word of the sign said “River” and the second word said “Arena.” Then there was a drawing of waves curling around rocks between two pillars, one on each side of the word “Arena.”}\\
\\
\texttt{Maria looked down at the arena, where a big crowd was gathered to watch two large bears fighting.}\\
\\
\texttt{In one corner of the ring, an old grizzly bear was fighting a younger brown bear. The grizzly was huge and looked very strong and powerful as he used all four limbs to attack his opponent. The fight between the two bears was fierce and brutal as they fought each other with their claws, teeth, and fists. The bears continued to fight until one of them won.}\\
\\
\texttt{The fight was brutal and exciting. The crowd roared its approval as one bear defeated the other. \textcolor{orange}{[The victory against the enemy is celebrated. Scene: the arena. Characters: Ratzinger Kugelman]}}\\
\\
\texttt{Jake said, “You didn’t tell me there would be fights here! This is cool!”}\\
\\
\texttt{The old grizzly bear was so tired that he collapsed to the floor of the ring and let the brown bear climb on top of him. The brown bear punched his opponent in the ribs over and over again until the grizzly finally stopped fighting back. Then the brown bear bit into his opponent’s neck and tore out a huge piece of flesh. The victorious bear, covered in blood, jumped out onto a pile of hay in the middle of the ring to celebrate. \textcolor{olive}{[Survival in the new world is made possible by the defeat of the enemy. Scene: the arena. Characters: Ratzinger Kugelman]}}\\
\\
\texttt{The crowd cheered loudly as the brown bear tore apart the old grizzly bear and devoured his flesh. The audience screamed with terror when they saw the brown bear chewing on an eyeball as the last bite of food before he started to eat his opponent’s intestines.}\\
\\
\texttt{The brown bear tossed the grizzly’s mangled corpse aside and leapt back into the ring to celebrate his victory. The old grizzly bear was dead, and the crowd was thrilled. It was a brutal and bloody fight, but it had been a good fight nonetheless.}\\
\\
\texttt{The defeated grizzly bear was now the most famous brown bear in the new world. He had won against all odds and proved that even a little guy could take down a huge opponent. That was always the way it worked out in the new world. In this new world, only the strong survived, and anyone \textcolor{teal}{[The group of survivors celebrate their victory against the enemy. Scene: the arena. Characters: Maria Martinez, Ratzinger Kugelman]} who didn’t live up to that standard would die.}\\
\\
\texttt{Maria Martinez had watched the whole fight from her front-row seat in the audience. She was one of Ratzinger’s closest advisors, and she had seen all kinds of fights since they arrived. She had watched as Ratzinger had torn apart the group of survivors and devoured their flesh. But now, the tables had turned, and now it was Maria Martinez that held the power.}\\
\\
\texttt{She stood up from her seat with a smile on her face. “Well done, my friend! You did it!” she said to Ratzinger.}\\
\\
\texttt{Ratzinger Kugelman was celebrating his victory by taking bites of the dead grizzly bear carcass. He was covered in the blood and dirt from the fight, and he had a gigantic smile across his blood-splattered face. He was enjoying some of his favorite food: grizzly bear.}\\
\\
\texttt{The group of survivors cheered with joy at the sight of their leader. They had all suffered from the merciless attacks of the enemy, and now, finally, they were able to take a break. They celebrated the victory by drinking bear blood, eating carcasses, and cheering for their leader.}\\
\bottomrule
\caption{The story generated by \oursimpl{} for the plan shown in Table \ref{tab:plan4}. Colored text in brackets indicates the corresponding outline items for the following text. The high-level outline is mostly followed, although there are some larger errors in this generation. In particular, when the protagonists are supposed to fight the ``enemy'' at the end, it turns into watching bears fight each other in an arena instead. Some of the middle passages where the characters first learn to survive in the new world also do not quite follow the low-level outline items. Finally, some of the passages in the beginning where the world is ending are a bit repetitive.}
\label{tab:ours_ex4}
\end{longtable}
\end{small}

\begin{small}
\begin{longtable}{p{0.9\textwidth}}
\toprule
\textbf{\oursimpl{} Plan 5}\\
\midrule
\texttt{Premise: As a serial killer's next victim wakes up in a dark basement, she must use her wits to escape before she becomes the killer's next trophy.}\\\\
\texttt{Setting: The story is set in a dark, dank basement with concrete walls and a dirt floor.}\\\\
\texttt{Characters:}\\
\texttt{1.}\\
\texttt{Full Name: Valerie Marx}\\
\texttt{Character Portrait: Valerie Marx is a young woman in her early twenties.}\\
\texttt{2.}\\
\texttt{Full Name: John Doe}\\
\texttt{Character Portrait: John Doe is a middle-aged man with a lean and athletic build.}\\
\texttt{3.}\\
\texttt{Full Name: Male Celebrity A}\\
\texttt{Character Portrait: Male Celebrity A is a Hollywood actor in his forties.}\\
\\
\texttt{Outline:}\\
\texttt{1. Valerie Marx wakes up in a dark basement, bound and gagged. Scene:  Characters: Valerie Marx}\\
\texttt{\quad \quad a. Valerie regaines consciousness and finds herself in a dark basement. Scene:  Characters: Valerie Marx}\\
\texttt{\quad \quad \quad \quad \textcolor{red}{i. Valerie regaines consciousness and finds herself in a dark basement. Scene: the basement of a house. Characters: Valerie Marx}}\\
\texttt{\quad \quad \quad \quad \textcolor{orange}{ii. Valerie tries to free herself but realizes that her hands and feet are bound. Scene: the basement of a house. Characters: Valerie Marx}}\\
\texttt{\quad \quad b. Valerie manages to free herself and explore the basement. Scene:  Characters: Valerie Marx}\\
\texttt{\quad \quad \quad \quad \textcolor{olive}{i. Valerie manages to free herself and starts to explore the basement. Scene: the basement of a house. Characters: Valerie Marx}}\\
\texttt{\quad \quad \quad \quad \textcolor{teal}{ii. Valerie finds a door but it is locked. Scene: the basement of a house. Characters: Valerie Marx}}\\
\texttt{\quad \quad \quad \quad \textcolor{cyan}{iii. Valerie finds a window but it is too high up for her to reach. Scene: the basement of a house. Characters: Valerie Marx}}\\
\texttt{2. John Doe, the serial killer, tells Valerie that she must escape the basement before dawn or else she will become his next trophy. Scene:  Characters: John Doe, Valerie Marx}\\
\texttt{\quad \quad a. After making her escape, Valerie runs into John Doe. Scene:  Characters: John Doe, Valerie Marx}\\
\texttt{\quad \quad \quad \quad \textcolor{blue}{i. Valerie waking up in the basement and trying to escape. Scene: the basement. Characters: Valerie Marx}}\\
\texttt{\quad \quad \quad \quad \textcolor{violet}{ii. Valerie meeting John Doe. Scene: the basement. Characters: John Doe, Valerie Marx}}\\
\texttt{\quad \quad b. John Doe tells Valerie that she must escape the basement before dawn. Scene:  Characters: John Doe, Valerie Marx}\\
\texttt{\quad \quad \quad \quad \textcolor{red}{i. John Doe tells Valerie that if she wants to live, she must escape the basement before dawn. Scene: the basement. Characters: John Doe, Valerie Marx}}\\
\texttt{\quad \quad \quad \quad \textcolor{orange}{ii. John Doe gives Valerie a head start before he comes after her. Scene: the basement. Characters: John Doe, Valerie Marx}}\\
\texttt{\quad \quad \quad \quad \textcolor{olive}{iii. Valerie tries to escape the basement but runs into John Doe. Scene: the basement. Characters: John Doe, Valerie Marx}}\\
\texttt{\quad \quad \quad \quad \textcolor{teal}{iv. John Doe chases Valerie through the basement. Scene: the basement. Characters: John Doe, Valerie Marx}}\\
\texttt{3. Valerie uses her wits to escape the basement and evade John Doe. Scene:  Characters: John Doe, Valerie Marx}\\
\texttt{\quad \quad a. Valerie tries to escape but John Doe catches her. Scene:  Characters: John Doe, Valerie Marx}\\
\texttt{\quad \quad \quad \quad \textcolor{cyan}{i. Valerie tries to escape but John Doe catches her. Scene: the basement. Characters: John Doe, Valerie Marx}}\\
\texttt{\quad \quad \quad \quad \textcolor{blue}{ii. Valerie manages to talk John Doe into letting her go. Scene: the basement. Characters: John Doe, Valerie Marx}}\\
\texttt{\quad \quad \quad \quad \textcolor{violet}{iii. Valerie finds a sharp object and threatens to kill herself if John Doe doesn't let her go. Scene: the basement. Characters: John Doe, Valerie Marx}}\\
\texttt{\quad \quad b. Valerie uses her acting skills to fool John Doe and escape. Scene:  Characters: John Doe, Valerie Marx}\\
\texttt{\quad \quad \quad \quad \textcolor{red}{i. Valerie uses her acting skills to convince John Doe that she's escape. Scene: the basement. Characters: John Doe, Valerie Marx}}\\
\texttt{\quad \quad \quad \quad \textcolor{orange}{ii. John Doe falls for it and Valerie makes her escape. Scene: the basement. Characters: John Doe, Valerie Marx}}\\
\texttt{4. Valerie makes it to safety and the police catch John Doe. Scene:  Characters: John Doe, Valerie Marx}\\
\texttt{\quad \quad a. The police catch John Doe and Valerie is safe. Scene:  Characters: John Doe, Valerie Marx}\\
\texttt{\quad \quad \quad \quad \textcolor{olive}{i. Valerie calls the police after escaping from John Doe. Scene: side Valerie's house. Characters: John Doe, Valerie Marx}}\\
\texttt{\quad \quad \quad \quad \textcolor{teal}{ii. The police arrive and arrest John Doe. Scene: side the police station. Characters: John Doe}}\\
\texttt{\quad \quad b. Male Celebrity A comforts Valerie as the police take John Doe away. Scene:  Characters: Male Celebrity A, John Doe, Valerie Marx}\\
\texttt{\quad \quad \quad \quad \textcolor{cyan}{i. Male Celebrity A comfort Valerie as the police take John Doe away. Scene: Valerie's house. Characters: Male Celebrity A, John Doe, Valerie Marx}}\\
\texttt{\quad \quad \quad \quad \textcolor{blue}{ii. Male Celebrity A tells Valerie that she did a brave thing. Scene: Valerie's house. Characters: Male Celebrity A, Valerie Marx}}\\
\bottomrule
\caption{The fifth plan generated by \oursimpl{} in our main experiments. Colors indicate leaf nodes used to guide generation. Note that some characters' names have been replaced to avoid overlap with celebrity names. There are a couple of spelling errors attributable to the high temperature we use for generation; similarly, the appearance of celebrity names is odd. Otherwise, the plot seems reasonable.}
\label{tab:plan5}
\end{longtable}
\end{small}

\begin{small}
\begin{longtable}{p{0.9\textwidth}}
\toprule
\textbf{\oursimpl{} Story For Plan 5}\\
\midrule
\texttt{\textcolor{red}{[Valerie regaines consciousness and finds herself in a dark basement. Scene: the basement of a house. Characters: Valerie Marx]}\quad Valerie Marx woke and found herself lying on the concrete floor of a dank basement.}\\
\\
\texttt{The room stank of rotting things and mold. She smelled urine, feces, and other noisome bodily fluids. Valerie turned on her side to see how she was tied. Her wrists were fastened behind her back with a dirty twine and her ankles and knees were also fastened with twine, but she could not recognize the material. She tried to move, but the bindings were too tight. Valerie was in a small dark room, and there was no visible window, only a rectangular slit that appeared to be a basement door. She strained against her bindings, trying to get free.}\\
\\
\texttt{Valerie felt a light touch on her left foot, above the ankle. The touch made her skin crawl. There was a bump on the sole of her foot and it seemed to be digging into Valerie’s flesh. She pulled back her foot quickly in revulsion and found herself in a sharp twist of darkness. She rolled over on her back, trying to reach her right foot, but the darkness was too thick. Valerie realized she was in a blindfold of some sort. She could see nothing except a dark blur at the periphery of her vision. Valerie began to feel faint and realized she was beginning to lose consciousness. She pulled herself up on all fours, determined not to fall unconscious. She found that she was beginning to run out of air; the basement stank with a particular pungency that made her throat burn. \textcolor{orange}{[Valerie tries to free herself but realizes that her hands and feet are bound. Scene: the basement of a house. Characters: Valerie Marx]}}\\
\\
\texttt{She felt along the wall, but hit something hard before her hand reached what she was searching for. There was a faint hum of electricity, and suddenly Valerie felt hot pain and prickling all over her body. She screamed at the top of her lungs, hoping someone would hear her. Valerie pushed herself along the ground, trying to escape the pain. She found that her hands were now free, but she still had her feet tied together.}\\
\\
\texttt{Valerie pushed herself back along the ground as fast as she could, feeling around with her feet for anything that might be useful. She ran into something hard yet yielding. Valerie realized it was a door. She pushed the door open and stumbled out of the small room, into a larger one with cement walls. She tried to open her eyes and realized her hands were still bound.}\\
\\
\texttt{Valerie tried to free herself frantically but found that she couldn’t break the \textcolor{olive}{[Valerie manages to free herself and starts to explore the basement. Scene: the basement of a house. Characters: Valerie Marx]} binds, even though her hands were numb from struggling so hard.}\\
\\
\texttt{She heard a shuffling noise and then a quiet click upstairs. Valerie panicked and struggled harder, but all she could hear for a moment was the dull sound of her own desperate breathing. Then she heard shuffling downstairs again, and then a loud, heavy thump as something big dropped to the floor above Valerie. She could hear the footsteps of whoever it was run down stairs, and she realized that whoever it was had managed to find her and was hunting her in the basement. Valerie started to panic, trying desperately to escape from her binds. She began kicking at the basement walls, trying to find something useful.}\\
\\
\texttt{Valerie suddenly felt a sharp pain on the back of her legs and realized she had found something useful. She had managed to find a box of tools, which included some wire cutters. She started to cut through her binds. \textcolor{teal}{[Valerie finds a door but it is locked. Scene: the basement of a house. Characters: Valerie Marx]}}\\
\\
\texttt{As she struggled to free herself, Valerie heard a soft thud behind her and turned around. She saw two doors in the corner of the room, and as she stared at them they both snapped shut with a loud hiss. Valerie knew there was no way out of the room. She decided to smash a \textcolor{cyan}{[Valerie finds a window but it is too high up for her to reach. Scene: the basement of a house. Characters: Valerie Marx]} window, hoping that it would allow her to escape. She reached for the closest brick and swung it with all her might at the glass, breaking a bit of it but not enough to make a hole. Valerie dropped the brick and looked around for another. She saw some cinder blocks stacked up against the wall, and she grabbed one before dropping it on the window. It shattered loudly, and Valerie could see the bright sky through it. Then she heard footsteps behind her again and knew that she needed to escape quickly before the killer caught her.}\\
\\
\texttt{Valerie ran over to the window and took a look outside. The window was too high for her to reach, and she had no choice but to try the doors. She walked over to one of them and tried to open it, but it was locked. Valerie reached for the other door only to find it also locked tight. She opened her mouth to scream but found a long gag in her mouth. Valerie tried once again to open the door, but it would not budge. She found a small window in the upper corner and looked through it. It was too small for her to fit through.}\\
\\
\texttt{Valerie found a small vent near the window, and she stuck her head inside of it just as someone started \textcolor{blue}{[Valerie waking up in the basement and trying to escape. Scene: the basement. Characters: Valerie Marx]} to come down the stairs. She saw a set of keys on top of one of the boxes that were all around the basement, and she started to search until she found one with a keychain that said: John Doe in bold letters. Valerie quickly grabbed the keys and darted forward, only to find another solid door in front of her.}\\
\\
\texttt{She once again tried the key and found that it would not open this door either. Valerie slid down to the floor of the basement and started to cry. She was trapped in a basement with an insane murderer, and she had no idea what he would do to her next. She closed her eyes, and she could hear the footsteps coming toward her. She tightened up and prepared herself as the door opened.}\\
\\
\texttt{The killer ran at Valerie, but she evaded his grasp by rolling out of the way. He kicked her in the side of the head, knocking her to the ground once again. He picked her up by the neck and slammed her down on a nearby table. Valerie’s body went limp as she struggled to breathe.}\\
\\
\texttt{He moved to his right side of the basement, where he had placed Valerie’s lifeless body. He placed his hands on her stomach and began to press \textcolor{violet}{[Valerie meeting John Doe. Scene: the basement. Characters: John Doe, Valerie Marx]} in, feeling her ribs pop from his pressure. Valerie grabbed the table with both hands and screamed in pain.}\\
\\
\texttt{“Who are you? I don’t know who you are, put me down!” Valerie yelled, but John Doe ignored her cries and continued to feel her up. \textcolor{red}{[John Doe tells Valerie that if she wants to live, she must escape the basement before dawn. Scene: the basement. Characters: John Doe, Valerie Marx]}}\\
\\
\texttt{“You have to escape, and you have to do it before dawn. You know that, don’t you? That’s how this game works. You have to escape before dawn or else I will begin taking your life away piece by piece.” John said, his voice thick and ominous.}\\
\\
\texttt{Valerie continued to scream as she felt her body being violated. She gritted her teeth and forced herself to stand again, using the table for support.}\\
\\
\texttt{John Doe looked at the young woman and smiled. He was about to say something when he heard a noise come from upstairs. He \textcolor{orange}{[John Doe gives Valerie a head start before he comes after her. Scene: the basement. Characters: John Doe, Valerie Marx]} quickly stood up and grabbed Valerie by the throat, strangling her with his bare hands.}\\
\\
\texttt{“What are you doing? Put me down!” Valerie screamed, but she could only manage weak coughs.}\\
\\
\texttt{She tried to thrash around in an attempt to loosen his grip, but John Doe was strong. He looked at Valerie with contempt and said, “I want to make this clear to you. You have a head start, and it’s your best chance of getting out of here alive. But if you don’t get out before dawn, then I will come after you myself and give you a head start of your own – head to my chopping block.”}\\
\\
\texttt{John Doe let go of Valerie’s throat and walked over to the steps. He turned around and gave the young woman one last cold look before he went upstairs. \textcolor{olive}{[Valerie tries to escape the basement but runs into John Doe. Scene: the basement. Characters: John Doe, Valerie Marx]}}\\
\\
\texttt{Valerie looked at the basement steps for a brief moment, hoping that John Doe would come down and finish her off. She saw the cellar door start to open, but it was too late; John Doe had left her alone with the sun just beginning to rise from the horizon.}\\
\\
\texttt{Valerie shakily reached underneath her and felt for the door handle. She turned around and ran up the steps, falling over in her haste to get away from the basement. She ran into John Doe’s living room, which was now adorned with blood-stained sheets on the walls and blackened \textcolor{teal}{[John Doe chases Valerie through the basement. Scene: the basement. Characters: John Doe, Valerie Marx]} photographs that once showed his happy family, and she ran into the bedroom where his wife and children used to sleep until they were ripped away by his psychotic hands.}\\
\\
\texttt{Valerie saw her chance to escape when John Doe came back into the house – he had gone upstairs to change his bloody clothes – and she took off through the front door, which slammed shut behind her.}\\
\\
\texttt{John Doe opened the door, ready to chase after Valerie. He stopped when she ran down the steps of the front porch and disappeared around the corner. She was running straight into oncoming traffic, but John Doe did not care. He only cared about one \textcolor{cyan}{[Valerie tries to escape but John Doe catches her. Scene: the basement. Characters: John Doe, Valerie Marx]} thing, and Valerie knew that there was no way for her to escape now.}\\
\\
\texttt{“Valerie Marx,” he said to himself while waving a meat cleaver in her direction. “I think it’s time for you to die.”}\\
\\
\texttt{Valerie stopped running when she saw John Doe, her only avenue of escape blocked off by the vicious murderer. She tried to run towards the nearest house, but John Doe began his slow and torturous approach towards her.}\\
\\
\texttt{John Doe came closer and closer until he was standing right in front of Valerie. Valerie did not try to fight John Doe because she knew he could easily overpower her. She knew that John Doe was going to use his meat cleaver to chop off her head, and just as he swung the blade down towards her neck, she closed her eyes and prepared for death. \textcolor{blue}{[Valerie manages to talk John Doe into letting her go. Scene: the basement. Characters: John Doe, Valerie Marx]}}\\
\\
\texttt{Valerie opened her eyes when she felt the meat cleaver swing down towards her neck, which meant that John Doe had not killed her yet. He was about to do so, but something made him stop, and he put the meat cleaver down on the ground, instead of killing Valerie.}\\
\\
\texttt{Valerie got up from the ground, wiping the blood off her neck and looking at John Doe with an angry and hurt expression on her face. “Why’d you let me go?” she asked angrily. “I told you I wanted to end my life. You promised me that \textcolor{violet}{[Valerie finds a sharp object and threatens to kill herself if John Doe doesn't let her go. Scene: the basement. Characters: John Doe, Valerie Marx]} if I escaped before sunrise you wouldn’t come after me. Why didn’t you keep your promise? Where is the man who was going to let me die in peace?”}\\
\\
\texttt{At that moment, Valerie saw a knife on the floor, and she quickly picked it up and pointed it towards \textcolor{red}{[Valerie uses her acting skills to convince John Doe that she's escape. Scene: the basement. Characters: John Doe, Valerie Marx]} her own throat, trying to scare John Doe into believing that she was going to kill herself right then and there. “Where is he?” she asked menacingly. “I don’t see him anywhere! He must be dead!”}\\
\\
\texttt{John Doe was scared for a moment, but then he remembered that Valerie Marx was just an actress, and he knew that she wasn’t going to actually kill herself. Instead, John Doe would have killed her himself if she had actually done it, and then he would have spent the rest of his life in prison. Valerie Marx had used her acting skills to get him to let her go. He knew that she was pretending to be a suicidal person because he had told her that he would let her escape before sunrise if she did. John Doe knew that Valerie was bluffing because he didn’t think that she was \textcolor{orange}{[John Doe falls for it and Valerie makes her escape. Scene: the basement. Characters: John Doe, Valerie Marx]} really going to kill herself just for the sake of getting away from him.}\\
\\
\texttt{John Doe quickly tried to disarm Valerie by taking the knife out of her hand, but she had already pulled it away from her neck and was ready to attack him. For a brief moment, John Doe thought about trying to get Valerie into a headlock and controlling her with his legs. He had done that trick to five other women, and it always worked out well even when he couldn’t find a weapon. It was too late for John Doe to make a move though because Valerie Marx had already started running towards the door, and she made \textcolor{olive}{[Valerie calls the police after escaping from John Doe. Scene: side Valerie's house. Characters: John Doe, Valerie Marx]} a tight left turn as soon as she reached the stairs.}\\
\\
\texttt{John Doe had thought that he would have been able to see Valerie Marx from the side of the house before then, but he had been wrong about that. There was no one in sight when Valerie ran away from him. This was a bad situation for Valerie because John Doe knew that she was in trouble. Her chances of surviving the night were slim to none, and John Doe had no intention of letting her get away with her life, so he decided that there was no point in following her.}\\
\\
\texttt{Valerie Marx had successfully escaped from John Doe, but she wasn \textcolor{teal}{[The police arrive and arrest John Doe. Scene: side the police station. Characters: John Doe]} ’t safe yet because he could catch up with her at any moment. The police were just around the corner, and she was walking towards them even though she didn’t see them. It only took a few minutes for her to reach the police station since it was only a block away from her house, but John Doe could have easily caught up with her by then if he was able to move as fast as he did before, which would allow him to be close enough to kill her when she reached the police station.}\\
\\
\texttt{Valerie Marx did not see the police when she arrived at the station because they were around a corner and around a car. This was good for John Doe because he had less of a chance of being spotted by them as long as he kept moving to the right side of the police station. It was also good for him because it gave him more time to think about what should be done next.}\\
\\
\texttt{The police station might have been safe for Valerie, but it was not a safe place for John Doe to be because they would be able to easily see him if he went inside. They might have also been able to find out who he was from the information that he left behind in the basement. If they were \textcolor{cyan}{[Male Celebrity A comfort Valerie as the police take John Doe away. Scene: Valerie's house. Characters: Male Celebrity A, John Doe, Valerie Marx]} able to do that, then he might have ended up being under more suspicion than before. Because of those reasons, it was better for him to just keep running for now and wait until things cooled off before taking another victim.}\\
\\
\texttt{If Valerie Marx walked inside the police station, then she would be safe from John Doe for now because the police would surround her and keep her from harm. On the other hand, Valerie did not want to go into the police station because she had more important places to go.}\\
\\
\texttt{Male Celebrity A was not at the police station when Valerie Marx arrived there because he had gone to Valerie’s house to make sure she was safe. When Valerie arrived at her house, Male Celebrity A was already there. He took Valerie inside her house and locked the door so that John Doe could not get in. Male Celebrity A talked with Valerie while they were inside her home and comforted her as she went through what had just happened to her.}\\
\\
\texttt{Valerie was glad when Male Celebrity A arrived because she had been starting to feel alone and scared after moving house and changing jobs. She explained everything that happened in the basement to Male Celebrity A, including how she got there and why she was moving house. She told Male Celebrity A how scared she had been and how she was moving house because of the situation with John Doe, but also because her old apartment building was right next to an old graveyard. Valerie was in a bad way because moving house like that would make anyone feel alone, and she needed someone to support her through that kind of ordeal. \textcolor{blue}{[Male Celebrity A tells Valerie that she did a brave thing. Scene: Valerie's house. Characters: Male Celebrity A, Valerie Marx]}}\\
\\
\texttt{Male Celebrity A listened to everything Valerie said without saying anything for a long time. After he finished listening, he said that Valerie had done a brave thing when she went into the basement to confront John Doe.}\\
\\
\texttt{Male Celebrity A was very happy when he heard what Valerie had done. He did not think it would be brave to visit a murderer in a dark basement, but Male Celebrity A knew that Valerie had wanted to do that and was happy that she did.}\\
\\
\texttt{Male Celebrity A told Valerie that she was very brave because she had stood up to the killer after waking up tied up, without being able to move or speak. He said that it was a very scary thing to do, but she had done it and was safe now.}\\
\\
\texttt{Valerie Marx was happy to hear what Male Celebrity A said. It made her feel better about herself and the situation.}\\
\bottomrule
\caption{The story generated by \oursimpl{} for the plan shown in Table \ref{tab:plan5}. Colored text in brackets indicates the corresponding outline items for the following text. For the most part the story follows the outline fairly well. However, some of the last few passages seem odd, which may also be partially the fault of the outline (e.g., due to the strange introduction of the celebrity).}
\label{tab:ours_ex5}
\end{longtable}
\end{small}

% \section{Example \texttt{text-davinci-003} Stories}

% Out of curiosity, we attempted to directly generate stories with \texttt{text-davinci-003}, finding that 

\twocolumn

\section{Dataset and Model Licenses}

The only pre-existing dataset we use in this work is WritingPrompts~\cite{fan2018hierarchical}, a dataset of English stories which uses the MIT License. Other than GPT3, other models are accessed through HuggingFace~\cite{wolf2020transformers}, which uses the Apache License 2.0. Our use of datasets and models in this work is consistent with their intended use. 

\end{document}